\documentclass[conference]{IEEEtran}
\pdfoutput=1
\usepackage{float}
\usepackage{authblk}
\usepackage{amsmath}%
\usepackage{amsfonts}%
\usepackage{amssymb}%
\usepackage[english,oldcommands,plain,lined]{algorithm2e}
\usepackage{graphicx} \DeclareGraphicsExtensions{.png,.jpg,.eps}   %%% standard extension for included graphics

\usepackage{subfiles}
\usepackage{graphics}
\usepackage{setspace}
\usepackage{amsmath}
\usepackage{graphicx,subfigure}
\usepackage[left=1in,top=1in,right=1in,bottom=1in]{geometry}
\usepackage{subfigure}    %% For putting multiple figures together.
\usepackage{caption}
\usepackage{makeidx}      %% creating index
\usepackage{nomencl}	  %% For list of abbreviations
\usepackage{url}
\usepackage[colorlinks=true,linkcolor=blue,citecolor=magenta]{hyperref}

\usepackage{longtable}

\usepackage{lettrine}
\usepackage[T1]{fontenc}
\usepackage{multirow}
\usepackage{rotating}
\usepackage{subfigure}
\usepackage{times}
\usepackage{wasysym}
\usepackage[normalem]{ulem}
\usepackage{tfrupee}
\usepackage[official]{eurosym}

\newcommand\blfootnote[1]{%
  \begingroup
  \renewcommand\thefootnote{}\footnote{#1}%
  \addtocounter{footnote}{-1}%
  \endgroup
}

\ifCLASSINFOpdf
\else
\fi
\begin{document}
\title{Design and Implementation of Path Trackers for Ackermann Drive based Vehicles}
\author{
Adarsh Patnaik\textsuperscript{*}\textsuperscript{a}\and
Manthan Patel\textsuperscript{*}\textsuperscript{a}\and 
Vibhakar  Mohta\textsuperscript{*}\textsuperscript{a}\and 
Het Shah\textsuperscript{*}\textsuperscript{a}\and
Shubh Agrawal\textsuperscript{*}\textsuperscript{a}\and
Aditya Rathore\textsuperscript{*}\textsuperscript{b}\and
Ritwik Malik\textsuperscript{*}\textsuperscript{c} Debashish Chakravarty\textsuperscript{c} Ranjan Bhattacharya\textsuperscript{a}\
% \thanks{* Denotes Equal Contribution \\
% a Department of Mechanical Engineering, IIT Kharagpur\\
% b Department of Electrical Engineering, IIT Kharagpur\\
% c Department of Mining Engineering, IIT Kharagpur
% }
}
% \author[1]{Adarsh Patnaik}
% \author[1]{Manthan Patel}
% \author[1]{Shubh Agrawal}
% \author[1]{Het Shah}
% \author[1]{Vibhakar Mohta}
% \author[2]{Aditya Rathore}
% \author[3]{Ritwik Malik}
% \author[3]{Debashish Chakravarty}
% \author[1]{Ranjan Bhattacharya}

% \affil[1]{Department of Mechanical Engineering, IIT Kharagpur}
% \affil[2]{Department of Electrical Engineering, IIT Kharagpur}
% \affil[3]{Department of Mining Engineering, IIT Kharagpur}
\maketitle

\begin{abstract}
This article is an overview of the various literature on path tracking methods and their implementation in simulation and realistic operating environments. The scope of this study includes analysis, implementation, tuning, and comparison of some selected path tracking methods commonly used in practice for trajectory tracking in autonomous vehicles. Many of these methods are applicable at low speed due to the linear assumption for the system model, and hence, some methods are also included that consider non-linearities present in lateral vehicle dynamics during high-speed navigation. The performance evaluation and comparison of tracking methods are carried out on realistic simulations and a dedicated instrumented passenger car, Mahindra e2o, to get a performance idea of all the methods in realistic operating conditions and develop tuning methodologies for each of the methods. It has been observed that our model predictive control-based approach is able to perform better compared to the others in medium velocity ranges.  
\end{abstract}%\\

\IEEEpeerreviewmaketitle
\section{Introduction}
\label{sec:intro}

\blfootnote{
\phantom{ll} \textsuperscript{*} Denotes Equal Contribution \\
\phantom{ll} \textsuperscript{a} Department of Mechanical Engineering, IIT Kharagpur \\
\phantom{ll} \textsuperscript{b} Department of Electrical Engineering, IIT Kharagpur \\
\phantom{ll} \textsuperscript{c} Department of Mining Engineering, IIT Kharagpur  \\
}

In past years, the development in the area of autonomous vehicles has been stupendous through technological advances made in the areas of communication, computing, sensors, and electronic devices along with advanced processing algorithms. In this section, we discuss the importance and literature review of implementing an autonomous system module, path tracker, that is necessary for accurate and safe manoeuvring of an autonomous vehicle. It is required for the proper functioning of modules like collision avoidance, lateral and longitudinal motion control.

Precise path following by a car-like robot is only possible if it is able to follow the desired steering angle and desired velocity at each time step during the traversal. A club of all such vehicle controllers is scientifically termed as path trackers. The expected goal of a path tracking algorithm is to generate actuator commands that minimize the cross-track error or simply the lateral distance between the vehicle's center point and the geometric path estimated by the motion planner. Apart from this, it is also expected that the controller minimizes the difference between the vehicle's orientation and path's heading while also limiting the rate of steering input for overall lateral stability and smooth driving experience. Path tracking controllers have been existing since the advent of autonomous vehicles and recent developments have made them very extensive and robust to system assumptions. 

For each of the controller presented in this work, its theory with system model is discussed first before implementation of the controller itself. To ease comparison based study, we limit the performance tuning parameters and analyze the performance of trackers on four different types of courses. Rest of this work is divided as follows. Section \ref{sec:sim_setup_vehicle_platform} describes the experimental setup used for simulation, various courses designed for analysis and computational platform used for carrying out the experiments. Path tracking algorithm, in general, can be defined as a combination of two control systems, i.e., lateral control and longitudinal control of the vehicle. Section \ref{sec:longi_control_vehicle} presents the development of longitudinal control of the vehicle, which is a pre-requisite for lateral control to perform at its best. This section introduces two types of velocity controller commonly found in industry, i.e., PID and Adaptive PID controller, and their comparison before selecting one as the base longitudinal controller for our lateral control study. Section \ref{lateral_control} then presents the complete study of lateral control starting from introduction to different vehicle driver models, then discusses the synthesis of commonly found lateral control theory in practice before finally discussing their analysis and head-to-head performance comparison. The experiments carried out in sections \ref{sec:longi_control_vehicle}, and  \ref{lateral_control} is carried within a computer simulation for quick prototyping of the complete control system framework before using the same framework on our real passenger car. Section \ref{results} then presents some of the pre-requisite experimentation needed on the passenger car and finally discusses the results of complete path tracking implementation. Ultimately, section \ref{sec:conc} concludes the presented work, in brief, its importance and illustrates how commonly used primitive control techniques are used successfully. It also discusses the need for developing more advanced path tracking theories as autonomous vehicles are moving towards high-speed operations and complicated objectives.

\section{Literature Review}
\label{sec:lit_review}

The importance of path tracking systems has already been discussed in the previous section. Two important survey papers \cite{snider2009automatic,paden2016survey} have been published so far that discusses the merits and demerits of various controllers developed in this domain. \cite{snider2009automatic} has classified the path tracking systems into three approaches viz. Geometric model-based, Kinematic model-based, and Dynamic model-based approaches. The geometric model revolves around tracking capability only by involving geometry of the vehicle(s) and path(s). Though this method seems naive, it was a core component of winning team's vehicle in DARPA grand challenge in the year 2004 \cite{thrun2006stanley}. The kinematic model discussed in this work derives an approximate kinematic bicycle model equivalent to the car, on which controllers were developed. Further, a dynamic model for the vehicle is also developed for incorporating the vehicle dynamics for enhanced precision tracking in high-speed navigation. \cite{paden2016survey} have discussed motion planning along with the path control approaches adjunctly as they mention that both of these fields to be closely related. They have divided the path tracking systems based on control objective, either path stabilization or trajectory stabilization. Similar to the kinematic model approximation, this work has explained different approaches separately as path stabilization using the kinematic model and trajectory stabilization using the kinematic model. This study has also discussed the application of predictive control approaches and parameter varying controllers as modern techniques to tackle the path tracking problem.

One of the most famous geometric model controller, the pure pursuit algorithm \cite{coulter1992implementation} was originally developed as a process for calculating the local path to get a robot back to the global path. Terragator, a six-wheeled skid steered robot used for outdoor vision experiments was the first application of this method. Modeling and control for a three wheeled mobile robot has been presented in \cite{pandey2017modeling}. Some of the studies, like the one done by \cite{andersen2016geometric} already have pointed out the problem with this tracker. That is, it faces problems like corner-cutting and significant steady-state error if an appropriate look-ahead distance is not selected, especially at higher speeds(greater than 30 kmph). Another famous geometric model, Stanley method \cite{thrun2006stanley}, is known to perform better than pure pursuit by avoiding corner-cutting and reducing steady-state error. It estimates the steering angle based on a nonlinear control law that considers the heading error of vehicle with respect to the path as well as the cross-track error that is measured from the front axle of the vehicle to the path. It is further shown that it has a higher tendency for oscillation and is not as robust to discontinuity in the path as compared to the pure pursuit method. In the past, researchers have also come up with a modified version of these geometric models like
% \textcolor{red}{replace embresten with agv paper}
\cite{andersen2016geometric, ohta2016pure,amer2018adaptive, ollero1994fuzzy} that focused on eradicating the cons of the mentioned methods. Feedback methods are also a plausible replacement to the geometric models as they are more robust to external disturbances and initial errors. An interesting method for controlling the kinematic model of non-holonomic systems (car-like robot) is discussed in \cite{de1998feedback}. This study derived a kinematic model of a car-like robot along with the feedback control law, which can also be easily extended to control 'n' trailers attached to the system. However, at higher speeds, this model too gives significant cross track error as it neglects the vehicle's lateral dynamics. At this level, the vehicle dynamics are considerably sophisticated, and high fidelity models are quite nonlinear, time-taking and vary with changing environments. Their negligence in vehicle system modeling leads to increased cross-track error in high-speed navigation. This has been very well investigated in the studies made by \cite{tavan2015optimal,heng2015trajectory}. They have also devised an Linear Quadratic Regulator(LQR) method that utilizes the inclusion of the lateral dynamics into the state-space model and gives fairly better results than the geometric and kinematic models. However, these methods include an approximate linearized vehicle dynamics and thus leaves a scope of improvement. Model predictive control \cite{garcia1989model} is known to be a general control design methodology which can be easily extended to path tracking problem. Intuitively, the approach is to resolve the motion planning problem over a short period horizon, then take a small time frame window of the output open-loop control, and apply it to the vehicle system. Like discussed in the previous section, the availability of high-end computational devices has made the computation of predictive control algorithm feasible in real-time and comfortable for use on-board vehicle. Some variations of model predictive control framework that are found in past literature include a) Unconstrained Model Predictive Control (MPC) with Kinematic Models \cite{ollero1991predictive},   b) Path Tracking Controllers \cite{kim2014model},   c) Trajectory Tracking Controllers \cite{falcone2007predictive}.

\section{Simulation Setup and Vehicle Platform}
\label{sec:sim_setup_vehicle_platform}

This work is focused towards implementation and performance evaluation of different path trackers, and hence, a typical experimental setup is necessary to minimize any error/bias induced by variation in hardware setup or software framework. Sufficient time was spent on choosing these platforms that are compliant with the academic literature and are also commonly used in industry to confirm for their usage reliability.

\subsection{MATLAB for Plant and Controller analysis}

Before implementing any of the path trackers, it is best to undertake some analysis on the model for its stability check and system behavior. We chose MATLAB Control Toolbox for the analysis of the models due to it's accurate results and it's easy interface. We were able to specify the path tracking controller as a transfer function, state-space, or frequency-response model. The toolbox also helped in tuning the SISO and MIMO compensators, like the PID based tracker and LQR based tracker. Validation of the design by verifying overshoot, settling time, rise time, and phase margins was also quick using this platform. Since we were not able to find any satisfactory resources using which, we could assemble the controller model defined in MATLAB with a car or could model the car-environment interaction, this platform remained useful only for mathematical analysis and observations. 

\subsection{Simulation world setup}
 \begin{figure}[!tbh]
\centering
\begin{subfigure}[Simulation's planned urban scenario]
{
  \includegraphics[width=35mm]{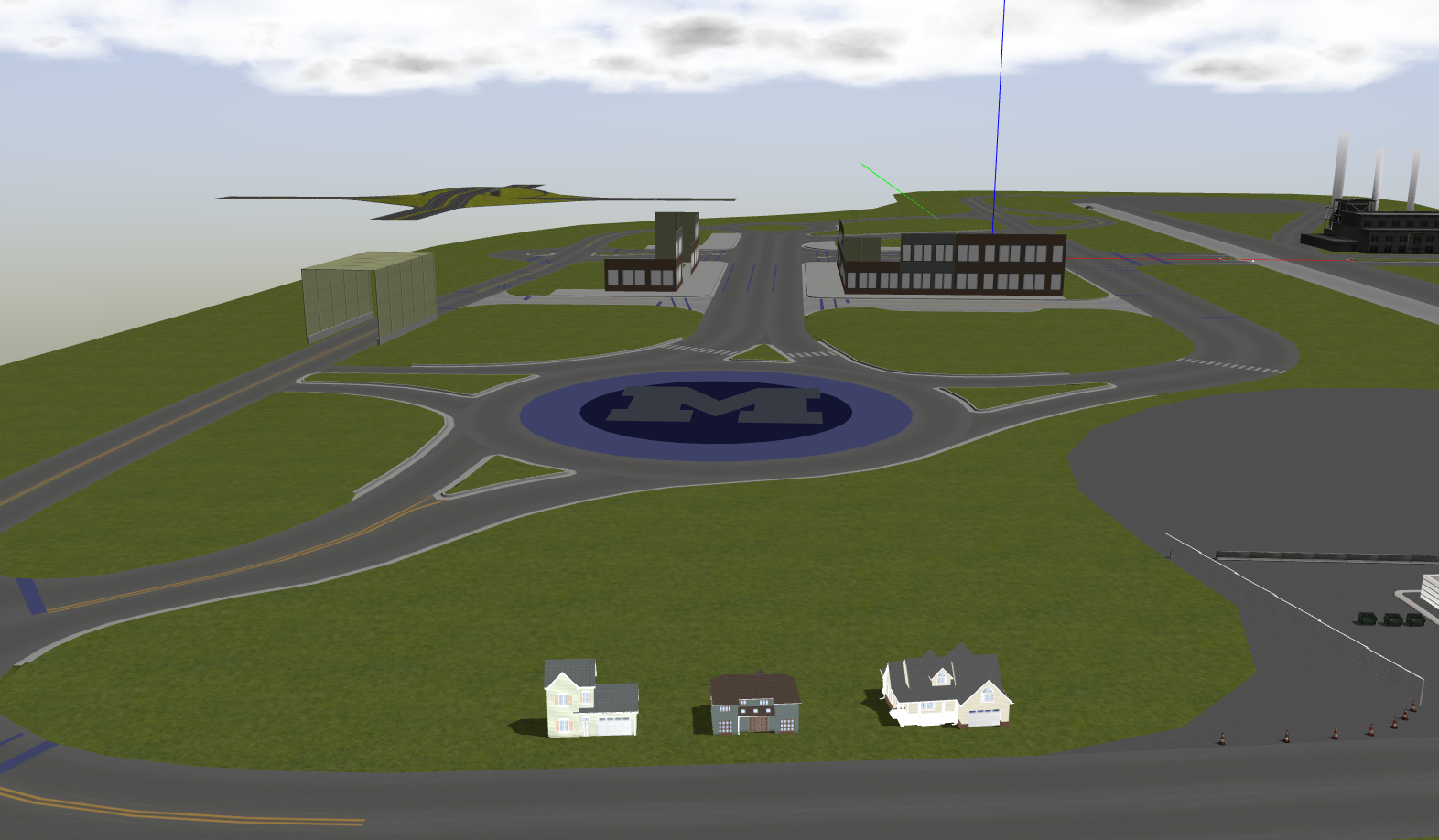}
}
\end{subfigure}
\begin{subfigure}[Simulation's vehicle for conducting experiments]
{
 \includegraphics[width=35mm]{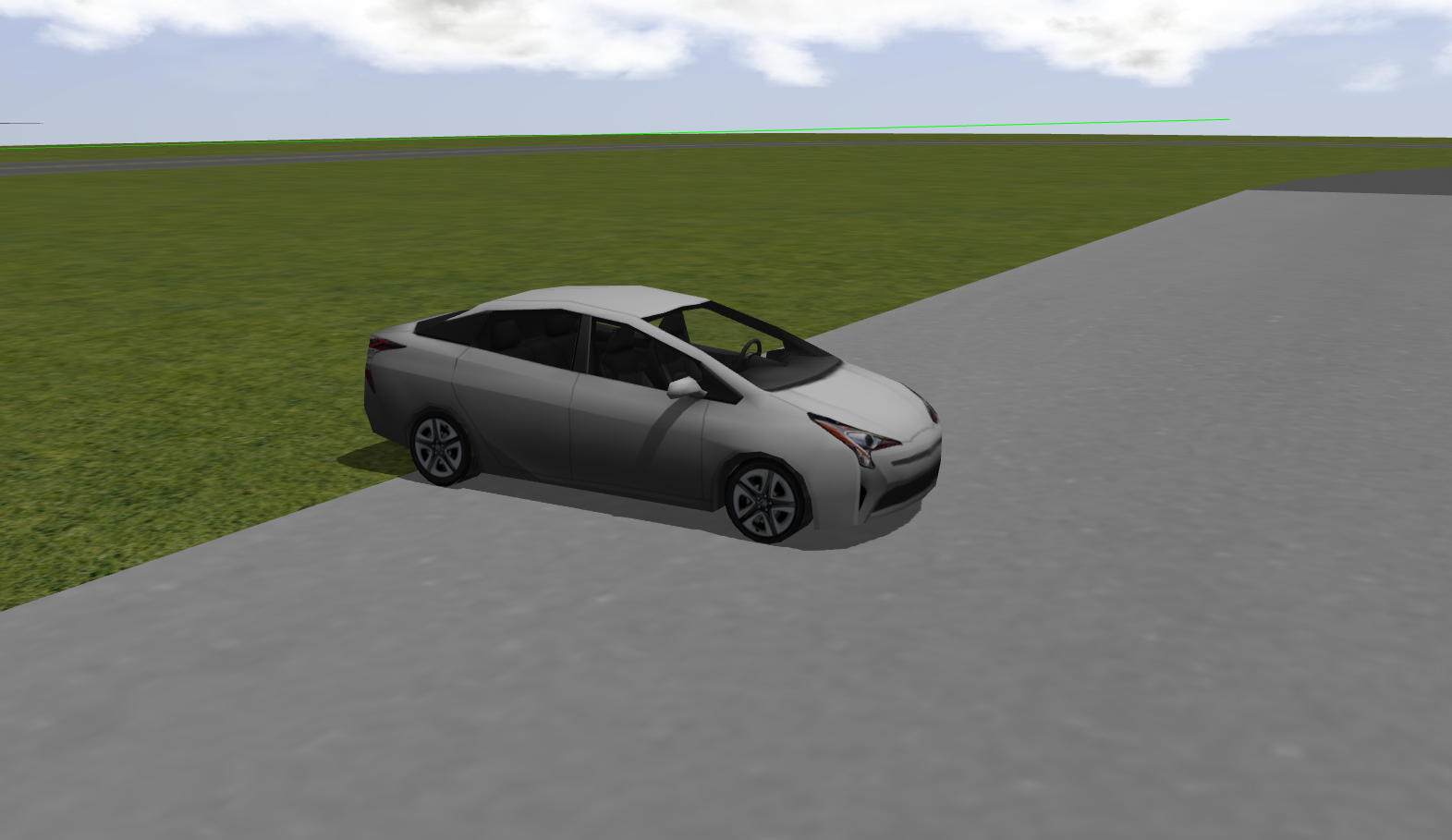}
}
\end{subfigure}
\caption{OSRF Car Simulator screenshot}
\label{fig:simu_env}
\end{figure}
% Autonomous vehicles have complexity right from mechanical domain to embedded system and then to software. Testing any one of these modular entity requires others entity to be fully functional. In the presented work, we are implementing path tracking controllers that can be tested easily if only the mechanical model and embedded system of the autonomous vehicle are ready to use. Even before testing the algorithms on some hardware, it is preferred testing on a simulation environment first to observe behaviour of vehicle in different situations and terrain. Devising a perfect controller would come from modifying a naively defined controller over and over, and every time uploading the model definitions for actual hardware testing (vehicle computer) will incur unnecessary spending of time and effort. However, with simulation, a single line of command can launch the virtual environment and test the controller. A common approach would be to test the controller model on MATLAB first, then rewrite it for the simulation platform.

Before testing the algorithms on some hardware, it is preferred to test in a simulation environment. We chose Gazebo with the Robot Operating System (ROS) as backend as it is widely used as a simulation stack in the robotics research community. There exist various packages pre-compiled that control the robots so that we focus towards only the module of our interest, in our case, path tracking. 
% Miscellaneous tools that are required to access sensor data, process it and generate appropriate response for the actuators is also provided by ROS. It is also designed to act as a fully distributed system in terms of computation wherein different computers can take part in computing processes or control. This makes the whole ROS system act together as a single entity. 
For our purpose, we were able to find a simulation world that included a car and urban environment, just enough to satisfy the requirements. This simulation software released by OSRF realistically models various parameters of the car and its interaction with the surrounding environment quite similar to the real-world scenario. Figure \ref{fig:simu_env} is a screenshot of car\_demo animation. The controller programs were implemented using Python and C++ language, and they communicate with this simulator at a high rate with a visualization window that helps in investigating their performances.

All the demonstrated experiments viz. analysis using MATLAB and controller performance in the simulation world were carried out on an Intel Core i5 @1.7 GHz x 2 machine loaded with 8 GB RAM and Ge-Force 820M GPU running Ubuntu 16.04 OS.
%Simulation's planned urban scenario
%Simulation's vehicle for conducting experiments

\begin{figure}[h]

\centering
\begin{subfigure}[Straight line course]{
   \includegraphics[width=35mm]{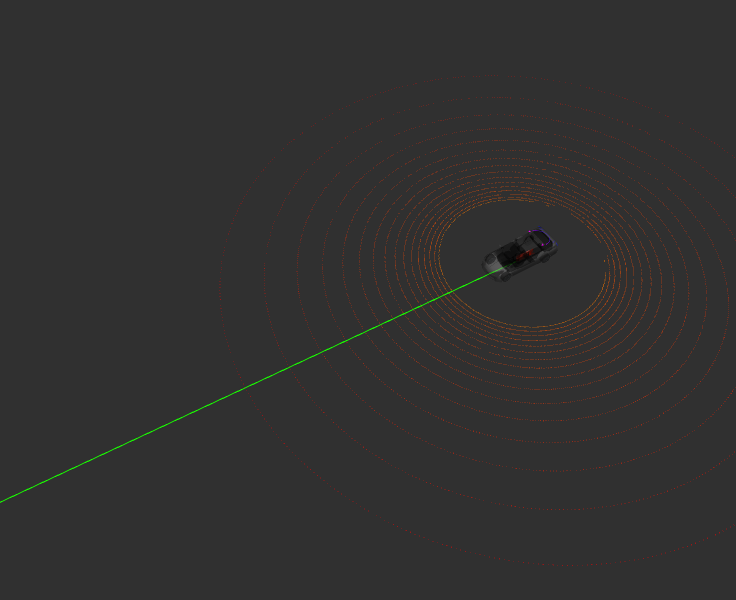}
   %\caption{Conventional feature based monocular visual odometry}
   } 
\end{subfigure}
\begin{subfigure}[Circular course]{
   \includegraphics[width=35mm]{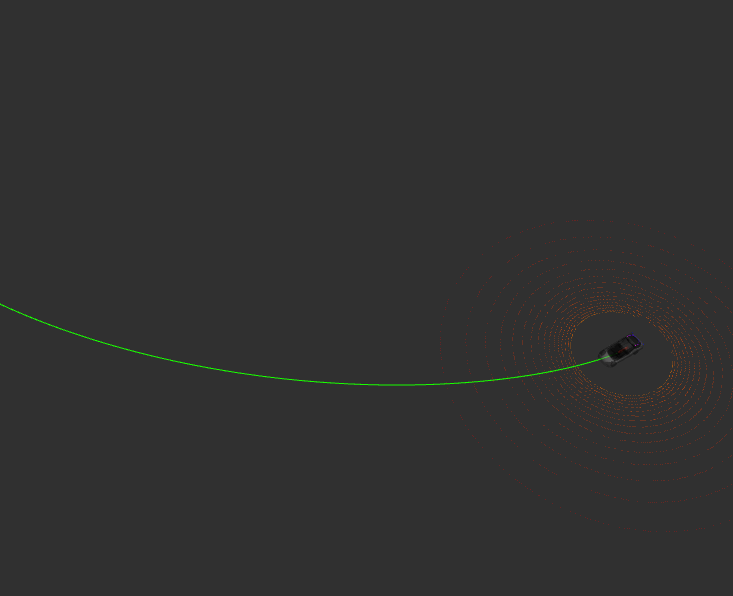}
   %\caption{Deep learning based end-to-end framework}
   }
\end{subfigure}
\begin{subfigure}[Lane shift course]{
   \includegraphics[width=35mm]{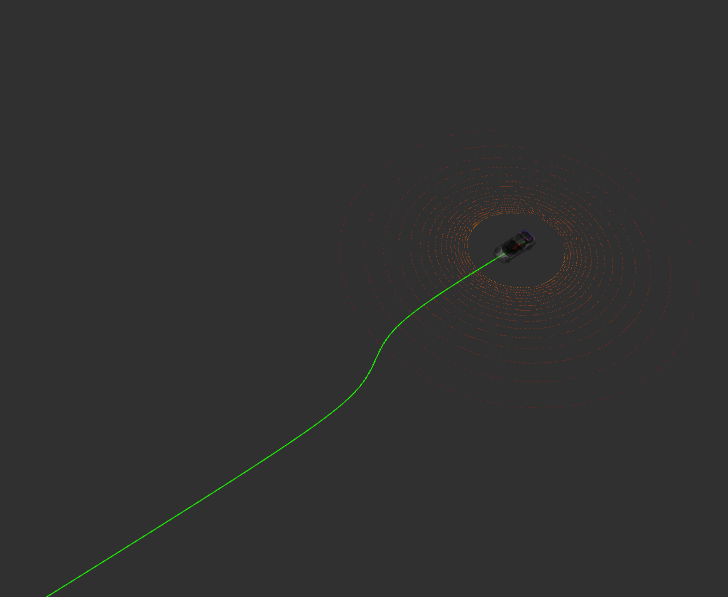}
   %\caption{Conventional feature based monocular visual odometry}
   } 
\end{subfigure}
\begin{subfigure}[Sinusoidal course]{
   \includegraphics[width=35mm]{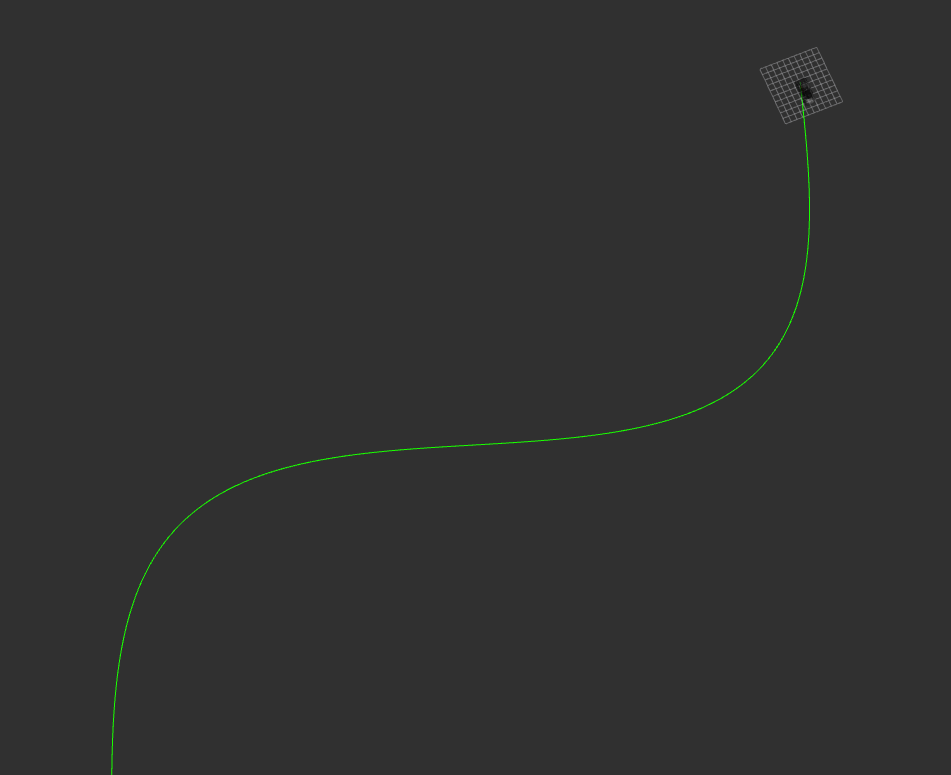}
   %\caption{Deep learning based end-to-end framework}
   }
\end{subfigure}
\caption{Courses used for validating path tracking capabilities}
\label{fig:courses}
\end{figure}

All of the trackers were to be tested on some predetermined paths. Rather than having them tested on a random driving course, they were validated against some specific geometry of paths to provide insights into their relative advantages and disadvantages in an organized manner. Figure \ref{fig:courses} shows a pictorial representation of the defined courses.

\subsubsection{Straight line course}
This is the most straightforward course for the vehicle to follow. This course basically validates if the vehicle can maneuver a given path without much attention given to tuning of parameters. 
% It also helps us verify if the sign conventions are appropriate for the controller, and there is no mistake in the sign assumptions.
Once the vehicle is fine-tuned, the maneuver on a straight-line path can give expectation value of minimal cross-track, settling time, and peak cross-track error. This track was performed with constant velocities of 10 kmph, 20 kmph and 35 kmph indicating slow, medium, and high-speed scenarios respectively.

\subsubsection{Lane shift course}
The lane shift course is a scenario in which the vehicle is expected to perform a simple lane change procedure. This course is a standard test for a vehicle undergoing lane change on a highway or a collision avoidance scenario. It is also chosen to demonstrate the tracking capability as well as response to an instantaneous, continuous section. Maneuvers on this course were also carried at constant velocities of 10 kmph, 20 kmph and 35 kmph similar to the cruise control of the vehicle on a highway.

\subsubsection{Circular course}
This course consists of a circular path of a constant radius (higher than the vehicle's minimum turning radius). It can be related to a roundabout in an urban driving scenario near the square. Interestingly, this course can provide valuable insight into the control of vehicle like the steady-state characteristics while maneuvering a constant non-zero curvature path like this. Experiments on this course were carried at constant velocities of 10 kmph, 20 kmph and 35 kmph like the previous courses.

\subsubsection{Sinusoidal course}
This course consists of a sinusoidal curve-shaped path, which is usually difficult to encounter in an urban driving scenario. The main highlight of this course is continuously varying curvature that changes in both magnitude and sign over the path length. It is expected that a controller's response to this type of continuous variance provides an insight into its robustness and responsiveness. Maneuvers on this course were performed at constant velocities of 10 kmph, 20 kmph and 35 kmph like the previous courses.

\subsection{Passenger car setup for laboratory-scale deployment}
\begin{figure}[!tbh]
\centering
\includegraphics[width=60mm]{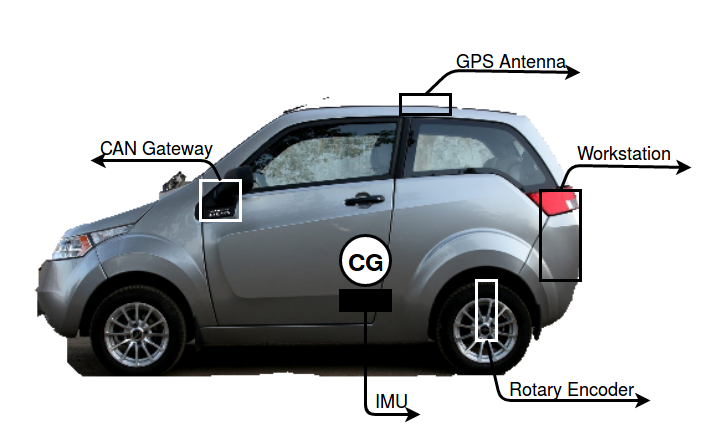} 
\caption{Vehicle Instrumentation}
\label{fig:vehicle_instru}
\end{figure}

While the simulator's physics engine tries its best to simulate the interaction between the car and the environment, there will always be some noisy processes and unmodelled vehicle dynamics. Also, we are able to observe non-linear mapping which is usually considered linear in simulation world like the mapping between steering column to actual steering angle, the mapping between throttle to vehicle's acceleration and effect of unequal loading on the vehicle tires. Hence, we also prepare to test out some of the controllers on a dedicated instrumented passenger car in real world scenario. We employed the passenger car "Mahindra e2o", with CAN based drive by wire system installed for the tests. It is possible to input steering commands and throttle value to this vehicle electronically via the CAN bus, which makes it easier to use for research purposes. However, a separate driver is needed that allows conversion of commands from the USB port of the in-vehicle computer to the CAN bus. We assume it to be predesigned as the details of this device driver is not within the scope of this study.
%\subsection{Vehicle Instrumentation}

For accurate localization and control signal generation, the test vehicle was equipped with various sensors and computation unit. All of the software-based workflows were performed using an Intel core Xeon @1.7GHz x 8 machine loaded with 32GB RAM and operating Ubuntu 16.04. 
% A desktop screen for real-time monitoring of vehicle's state and program debugging was also set up at the back seat. This complete workstation was powered using an inverter with 150 Ah 12V battery configured inside the vehicle's trunk for safety reasons. 
IMU, GPS, and rotary encoders provided necessary data for vehicle's localization whereas Kvaser CAN module helped in communicating actuation signals to the vehicle system. Figure \ref{fig:vehicle_instru} shows overall setup of different instrumentation.

\subsection{Vehicle Localization}
Accurate localization of the vehicle is necessary to calculate the deviation of the path followed by the vehicle from the predetermined path.
% Motion capture systems are best suited for this purpose. However, they are generally restricted to indoor environments (most of them are IR based) and also won't be able to cover the vast area required for a passenger vehicle sized robot testing. Hence, 
We resorted to localization using the standard method of fusing wheel encoder data with the IMU data. For this, we utilized the popular robot\_localization package under the ROS framework that internally used an Extended Kalman Filter (EKF) for state estimation. EKF uses Gaussian noise assumption for input sensor signals that improve its ability to deal with uncertainty while estimating vehicle state. In this work, EKF was used to estimate X, Y (current position), and $\theta$ (current orientation) given the planar movement assumption of vehicle. After having assembled the rear wheels with rotary encoders and the CG of the vehicle with IMU, we were able to localize the vehicle using EKF for sufficiently long distance accurately.

\begin{figure}[!tbh]
\centering
\includegraphics[width=73mm]{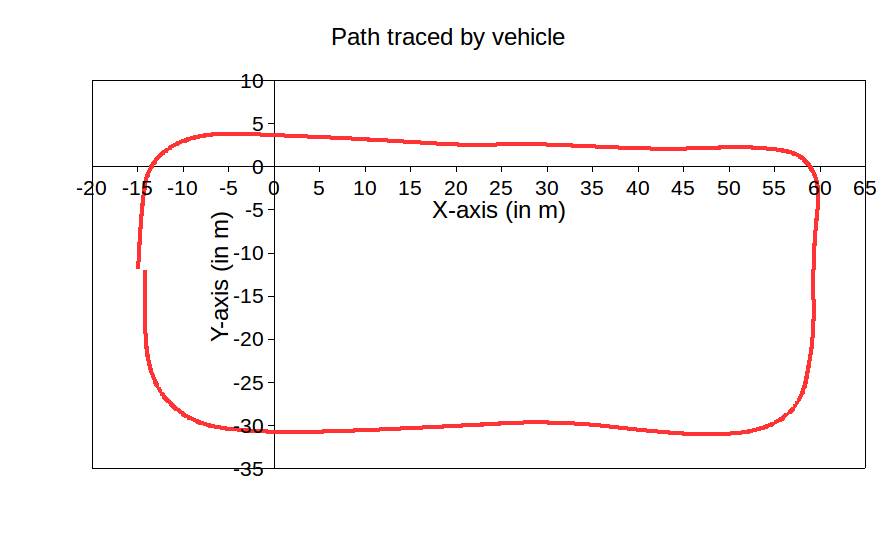} 
\caption{Loop closure given the path traced by vehicle}
\label{fig:loop_closure}
\end{figure}

To verify the precise nature of vehicle localization using EKF, we carried out a loop closure test for sufficiently long distances. A small orientation or displacement error at any point along the loop can show significant deviation from the initial position at the end of the test. Figure \ref{fig:loop_closure} shows the path trail calculated using EKF where the change in position error approximates $1.2$ m for a total traversed the length of 235 m. 

%loop closure with google maps background

\section{Longitudinal Control of Vehicle}
\label{sec:longi_control_vehicle}

As previously stated, path tracking is a problem majorly focused on lateral control. However, the longitudinal velocity does hold importance when a particular path tracking algorithm uses a velocity term for computing steering angle. Hence, for completeness, we demonstrate the velocity control capability of the simulated vehicle and realistic vehicle.

\subsection{Controller Synthesis}
\label{sec:longi_controller}
%change this
Following methods are chosen based on their simplicity and industry-wide acceptance as a control method. The theoretical basis is presented with simplification using necessary assumptions.

\subsubsection{PID Control}

\begin{figure}[!tbh]
\centering
\includegraphics[width=73mm]{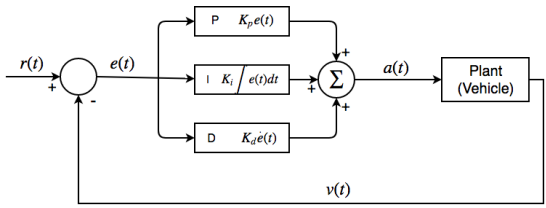} 
\caption{Illustration of PID velocity controller}
\label{fig:pidVelocity}
\end{figure}

Proportional-Integral-Derivative Controller (PID Controller) is one of the most popular control loop feedback mechanism for implementing cruise control of a vehicle. For our case, we considered the process variable as velocity and the manipulated variable as throttle (for both simulated vehicle and realistic vehicle). Thus, according to PID control theory, the error term $e(t)$, that is a difference between the setpoint and process variable, controls the throttle of vehicle proportionally. Figure \ref{fig:pidVelocity} shows the controller implemented for our purpose.

\noindent Here,

$a(t)$ be the throttle input to vehicle

$v(t)$ be measured velocity or current velocity

$r(t)$ be reference or set point

$e(t)$ be the error signal

$K_p$, $K_i$ $K_d$ be the proportional, integral and derivative gain constants respectively \\
\noindent Following equation was embedded into program for velocity control,

\begin{equation}
a(t) = K_p e(t) + K_i \int e(t) dt  + K_d \dot{e}(t)
\end{equation}

\noindent However, the above equation was converted to discrete form for actual use.

\begin{equation}
a(t) = K_p e(t) + K_i \sum_0^t e(t) + K_d \frac{e(t) - e(t-\Delta t)}{\Delta t} 
\end{equation} 

\noindent where $\Delta t$ is the sampling interval used for discretization.

% Figure here

%\noindent PID constant values can be found for the following specifications that gives descent response behaviour and charachteristics for the path tracking purpose.
\begin{itemize}
\item Rise time  <  100 ms
\item Steady-state error  <  2\%
\item Overshoot  <  10\%
\end{itemize}

\subsubsection{Adaptive PID Control}

PID control performs best when the experimental operating conditions are close to the tuning operating conditions. Deviations like change in total mass, the slope of the terrain, terrain-wheel interaction, etc. could make the controller a sub-optimal performer. These kinds of problems ought to occur in real-life application of the cruise control. Therefore, the PID constants need to change as a change in operating conditions occur. We resorted to adaptive PID control law, wherein, the constants similar to the ones defined in the previous section can change depending upon the response of the controller. One of the popular methods for such a self-tuning controller is the MIT rule \cite{sastry2011adaptive}. However, with this method, the system is known to get unstable with time-varying and noisy inputs. Finally, we implemented a technique that modifies the MIT rule and reduces the risk of instability. Following equations form the set of rules as explained in the work \cite{alonso2013self}.

\noindent Here,

$e_m (t)$ be the filter version (low pass first order filer in this case) of error signal $e(t)$

$D(t)$ be the derivative of error signal $e(t)$

$D_m (t)$ be the derivative of filtered error signal $e_m (t)$

\noindent Then, the modification in PID constants can be given as,

\begin{equation}
K_p (k+1) = K_p (k) + \gamma_p | e(k) - e_m (k) |
\end{equation}
\begin{equation}
K_i (k+1) = K_i (k) + \gamma_i e_m (k)
\end{equation}
\begin{equation}
K_d (k+1) = K_d (k) + \gamma_d | D(k) - D_m (k) |
\end{equation}

where $\gamma$ is the learning rate for the $K$ constants.

\subsection{Simulation Results and Discussion}

\begin{figure}[!tbh]
\centering
\begin{subfigure}[Unit step response of simple PID controller]{
   \includegraphics[width=35mm]{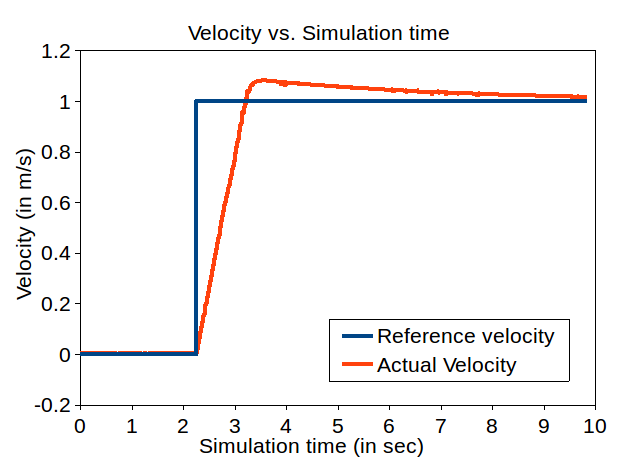}
   %\caption{Conventional feature based monocular visual odometry}
   } 
\end{subfigure}
\begin{subfigure}[Unit step response of Adaptive PID controller]{
   \includegraphics[width=35mm]{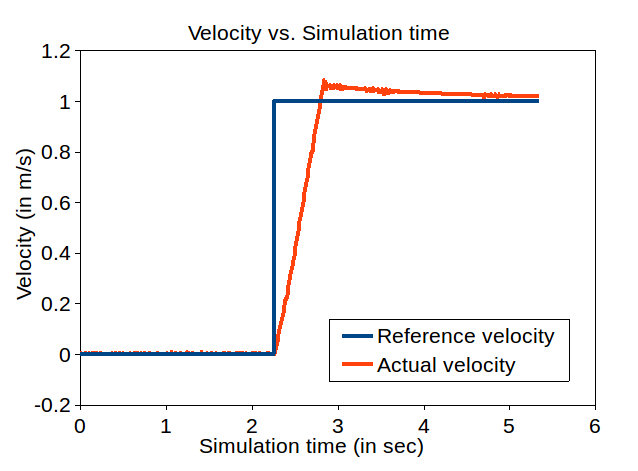}
   %\caption{Deep learning based end-to-end framework}
   }
\end{subfigure}
\caption{Unit step responses after manually tuning the controllers}
\label{fig:manual_step_response}
\end{figure}

Methods described in section \ref{sec:longi_controller} were implemented in our dedicated simulation environment. The parameters were tuned as per the desired specifications having a decent response behavior for path tracking purpose. Firstly, the trial-and-error method of tuning was carried out to observe the extent of tuning and error with such a manual method. In this method, the PID constants are initialized with some value for which we observe the response and improvise these values by repetitively changing the constant values until a desired sweet spot is reached. With a couple of observations, it became intuitive to tune the parameters as we observe that each of the constants $K_p$, $K_i$, $K_d$ correspond to response characteristics namely rise time, steady-state error and value oscillations respectively. With that intuition, the described simple PID controller was tuned for unit step response for which is shown in Figure \ref{fig:manual_step_response} (a). The same PID constants were then used for modified MIT-rule adaptive PID controller along with its learning parameter fixed to a nominal value of 0.1 each. This helped us observe the effect of using a learning parameter in conjunction with a simple PID controller's constants that characterizes an adaptive controller. Figure \ref{fig:manual_step_response} (b) shows the unit step response for Adaptive PID controller utilizing modified MIT rule.
 
\begin{figure*}[!tbh]
\centering
\begin{subfigure}[Velocity response of vehicle to throttle input as square wave]{
   \includegraphics[width=73mm]{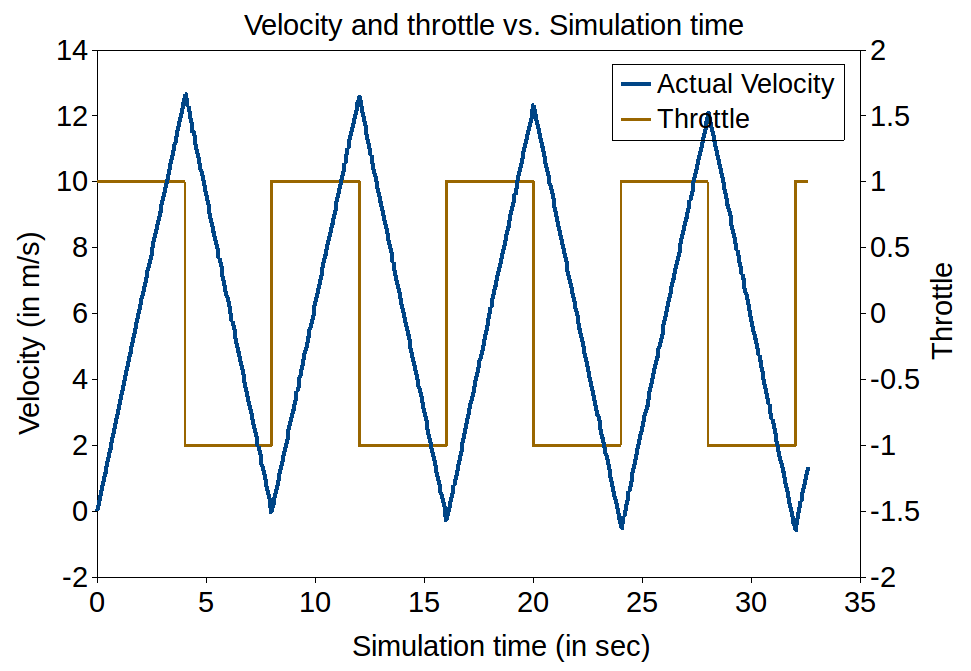}
   %\caption{Conventional feature based monocular visual odometry}
   } 
\end{subfigure}
\begin{subfigure}[Performance of simple PID controller using estimated gain values]{
   \includegraphics[width=73mm]{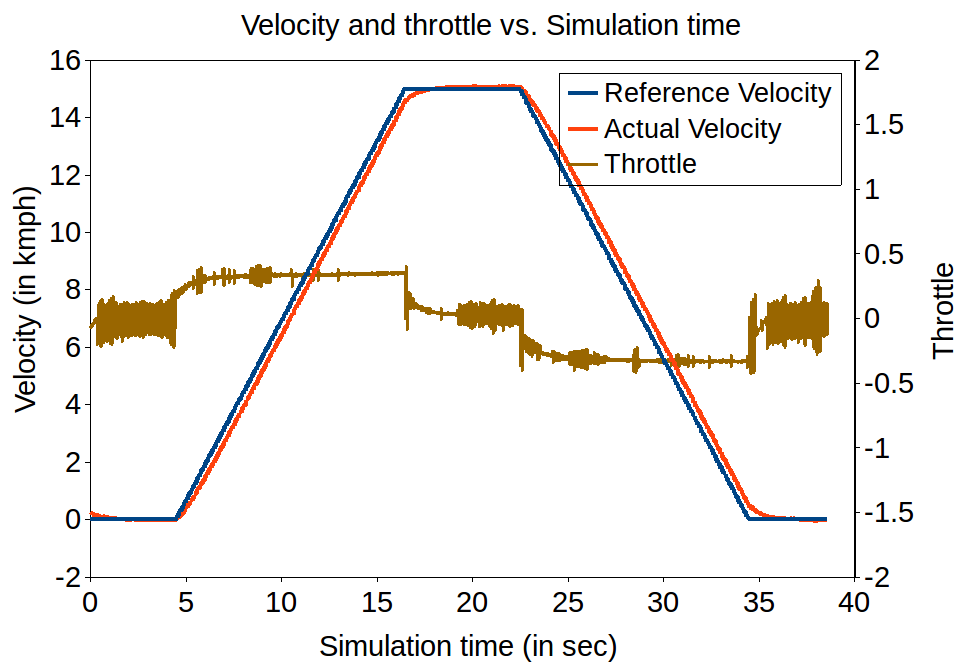}
   %\caption{Deep learning based end-to-end framework}
   }
\end{subfigure}
\caption{Parameter tuning using plant model identification}
\label{fig:modelled_step_response}
\end{figure*}
Tuning PID constants manually using the trial-and-error method is tedious and doesn't guarantee to settle to the expected desired specification. It truly depends on the person's experience tuning the controller and requires much practice to meet the specifications. It can be observed that having learning parameters as an adaptive PID controller does improve over the simple PID controller, though it can be expected to give much better response if PID constants are tuned appropriately.

\subsubsection{Plant Modelling}

Another method that can be used to tune PID constants is through modeling the plant by utilizing the mapping between input throttle to output velocity of the vehicle. For this purpose, we used MATLAB's control toolbox that included the functionality of fitting a model given a mapping. After a thorough review of existing practices used in industry, it was found that two types of input waveforms are majorly used to identify plant model. One is square wave input while the other is sinusoidal wave input. We went ahead with the usage of both the waveforms and observed which one performs better. 

\paragraph{Square wave input}

Using the gazebo simulation platform for the car, we scheduled a square wave input throttle to the vehicle and collected the actual velocity simultaneously. Figure \ref{fig:modelled_step_response} (a) shows the mapping between the throttle and observed the actual velocity of the vehicle. This was further processed in MATLAB's PID tuner using a state-space model of order 2 for model identification. Table \ref{table:gain_values} shows the PID constants obtained.

%\begin{figure}[!tbh]
%\centering
%\includegraphics[width=100mm]{results/longi_control/sw_response.png} 
%\caption{Velocity response of vehicle to throttle input as square wave}
%\label{fig:sw_response}
%\end{figure}

Having calculated the gain values, we tested it out on our simulation car with a given velocity reference to follow. Figure \ref{fig:modelled_step_response} (b) shows the performance of a tuned simple PID controller along with the throttle input to follow the velocity reference.

%\begin{figure}[!tbh]
%\centering
%\includegraphics[width=100mm]{results/longi_control/trapezium_response_sw_pid.png} 
%\caption{Performance of simple PID controller using estimated gain values}
%\label{fig:trapezium_sw_response}
%\end{figure}

\paragraph{Sinusoidal wave input}

\begin{figure*}[!tbh]
\centering
\begin{subfigure}[Velocity response of vehicle to throttle input as sinusoidal wave]{
   \includegraphics[width=51mm]{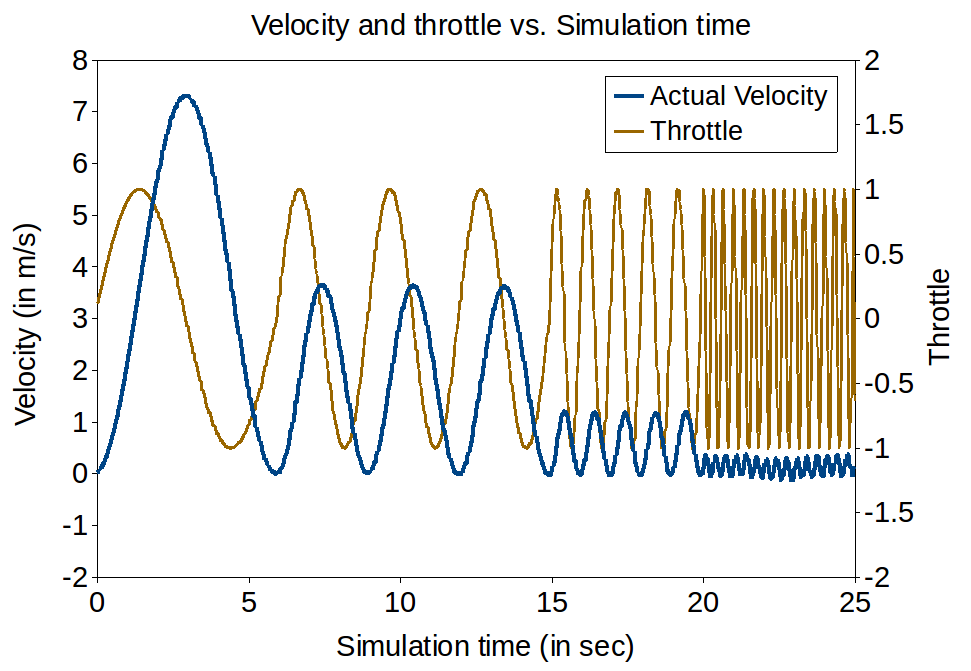}
   %\caption{Conventional feature based monocular visual odometry}
   } 
\end{subfigure}
\begin{subfigure}[Performance of simple PID controller using estimated gain values]{
   \includegraphics[width=51mm]{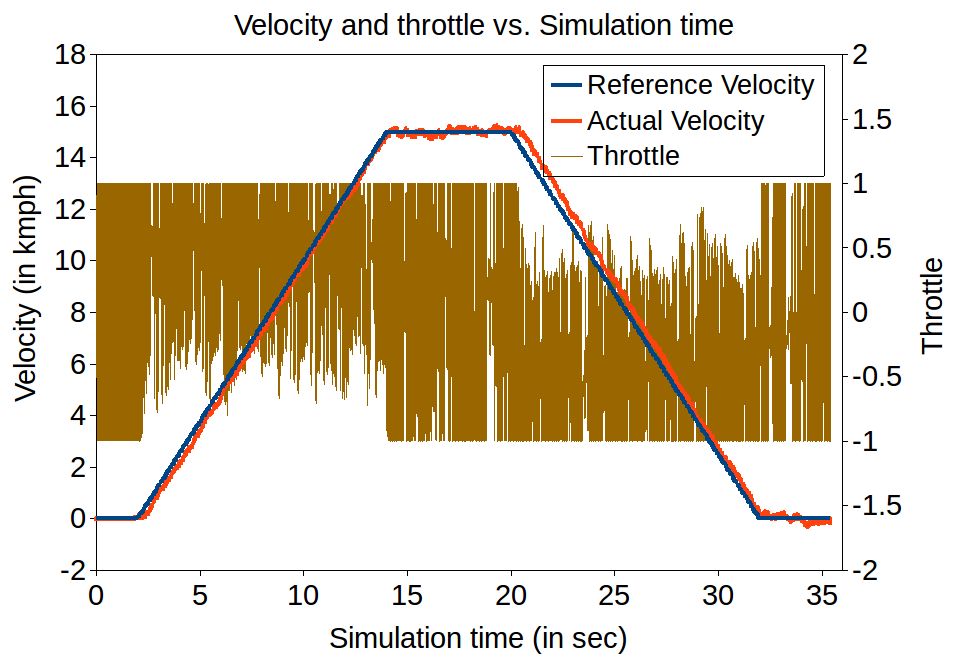}
   %\caption{Deep learning based end-to-end framework}
   }
\end{subfigure}
\begin{subfigure}[Performance of Adaptive PID controller using estimated gain values]{
    \includegraphics[width=51mm]{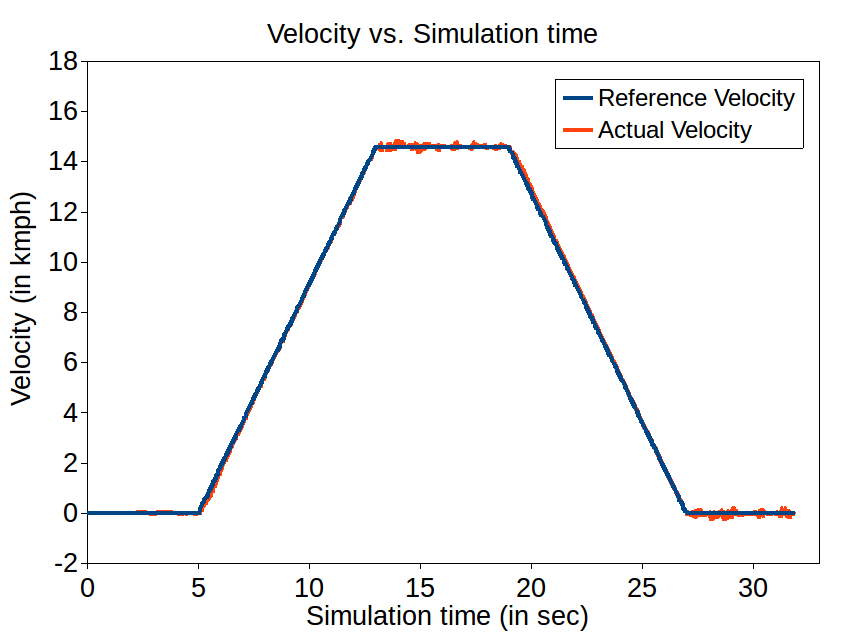} 
\label{fig:trapezium_adaptive_response}
}
\end{subfigure}
\caption{Parameter tuning using plant model identification}
\label{fig:modelled_sin_response}
\end{figure*}

Similarly, we scheduled a sinusoidal wave input throttle to the vehicle and collected the actual velocity simultaneously. Figure \ref{fig:modelled_sin_response} (a) shows the mapping between the throttle and observed the actual velocity of the vehicle. This was further processed in MATLAB's PID tuner using a state-space model of order 2 for model identification. Table \ref{table:gain_values} shows the PID constants obtained.

Having calculated the gain values, we tested it out on our simulation car with a given velocity reference to follow. Figure \ref{fig:modelled_sin_response} (b) shows the performance of a tuned simple PID controller along with the throttle input to follow the velocity reference.

\begin{table}
\begin{center}
\caption{Gain values estimated using MATLAB's PID tuner}
\label{table:gain_values}
\begin{tabular}{| c | c | c | c |}
\hline
\textbf{Waveform} & $K_p$ & $K_i$ & $K_d$ \\
\hline
\textbf{Square wave} & $0.63359$ & $0.00015$ & $26.31990$ \\
\hline
\textbf{Sinusoidal wave} & $2.236$ & $0.00186$ & $49.35450$ \\ 
\hline
\end{tabular}
\end{center}
\end{table}

\noindent As evident from Figure \ref{fig:modelled_step_response} (b) and Figure \ref{fig:modelled_sin_response} (b), reference velocity do get followed with a decent accuracy for both set of gain values but the throttle values requirement differ. Much fluctuation was observed in throttle value with gain values calculated using sinusoidal wave, and therefore it doesn't suit to practical implementation. Throttle actuators in a vehicle have certain response time and sensitivity, and with such a fluctuating throttle value it can damage the actuator, while also giving a different response than expected. Therefore, we went ahead with using square wave for estimating the gain values for further experiments.

\noindent Gain values calculated for simple PID controller (square wave) was also then incorporated into MIT rule adaptive controller with setting learning parameter values as $\gamma = 0.1$. Figure \ref{fig:trapezium_adaptive_response} shows the performance of this controller that is found to be better than a simple PID controller.

\subsubsection{Performance comparison}

\begin{figure}[!tbh]
\centering
\begin{subfigure}[Unit step response before weight addition]
{
   \includegraphics[width=36.6mm]{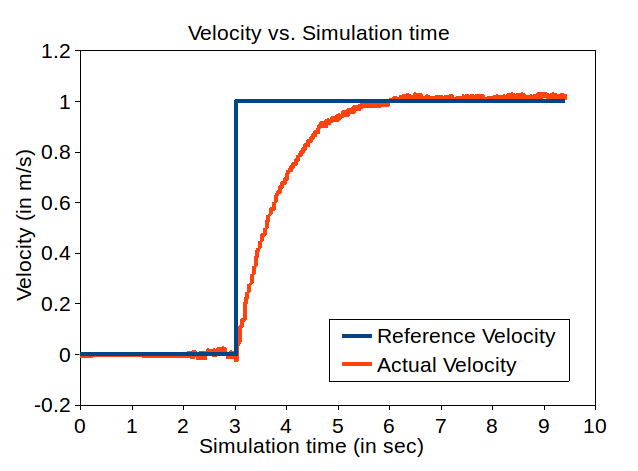}
   %\caption{Unit response before weight addition}
} 
\end{subfigure}
\begin{subfigure}[Unit step response after weight addition]
{
   \includegraphics[width=36.6mm]{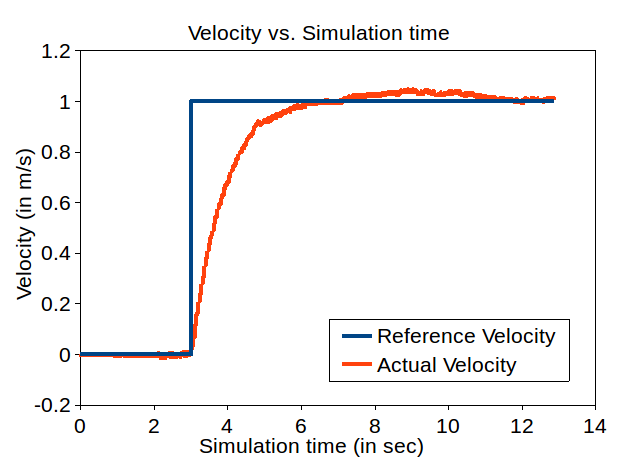}
   %\caption{Unit response after weight addition}
}
\end{subfigure}
\caption{Performance comparison of simple PID controller with vehicle weight change}
\label{fig:simple_pid_wt_change}
\end{figure}

\begin{figure}[!tbh]
\centering
\begin{subfigure}[Unit step response before weight addition]
{
   \includegraphics[width=36.6mm]{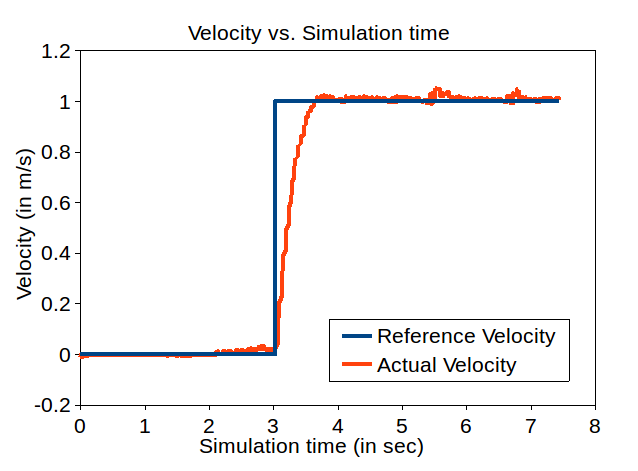}
   %\caption{Unit response before weight addition}
} 
\end{subfigure}
\begin{subfigure}[Unit step response after weight addition]
{
   \includegraphics[width=36.6mm]{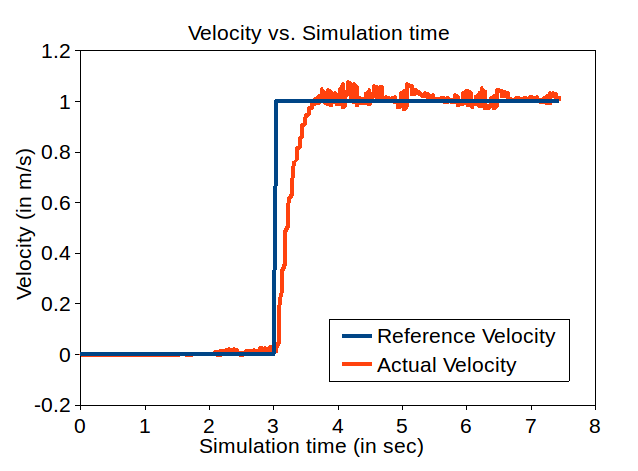}
   %\caption{Unit response after weight addition}
}
\end{subfigure}
\caption{Performance comparison of modified MIT-rule adaptive PID controller with vehicle weight change}
\label{fig:adap_pid_wt_change}
\end{figure}

Autonomous vehicles can be utilized for different purposes like public transportation, goods supply, or recreation. There is a definite possibility that these vehicles are subjected to change in weight during any of these purposes. This can be considered a system change. Another change the vehicle can witness is a change in the driving surface that can affect the friction coefficient between tire and road surface. This can be considered as environmental change. A simple PID controller is known to have resilience/robustness property absent with respect to system or environment change. To verify this, we manipulated the weight of the test simulation car by 180 kg (weight of 3 passengers) and plotted the velocity control result. Figure \ref{fig:simple_pid_wt_change} shows the performance comparison of simple PID Controller before and after the weight change.
%please shorten this para 3 words
\noindent It can be observed that after the weight addition, the Controller's rise time changes by an observable amount while the settling time changes significantly. This can be problematic for path tracking in a real-world scenario if the vehicles are tested within the same conditions only before deploying it in actual use case scenarios. Class of adaptive controllers specifically came into existence to resolve the issue of system change given the Controller is tuned to a specific condition. We carried out the same test for the described adaptive PID controller using MIT rule in the simulation environment. Figure \ref{fig:adap_pid_wt_change} shows the performance comparison of this adaptive controller before and after the weight change.

\noindent For adaptive PID controller, it can be observed that the change in rise time and settling time is much lower than that of the simple PID controller. This verifies the fact that adaptive controllers are better than standard controllers like PID in terms of robustness and resilience. Hence, for all our path tracking algorithms further described, we used modified MIT-rule adaptive PID controller as a longitudinal velocity controller for the vehicle in our work.

%\subsection{Conclusion}
%
%Using PID as a velocity controller was useful as it performed quite optimally on simulation world data. Though the manually tuned controller wasn't in par with industry grade performance, after having the plant modelled, significant increase in performance was observed. Plant modelling was most effective using unit step response that was modified to square wave input for consensus calculation. Sinusoidal wave input also did model plant well however the steering angle calulation using that model comprised of high frequency oscillations. This won't be effective on real world platform as there exists a degree of latency  between calculation and performing the computed throttle. Hence, the plant modelled using sinusoidal wave input was ruled out. We verified the claim that adaptive PID improved performance over the simple PID method as it utilized learning gain values at every time step. Adaptive PID method also shown some resilience to change in vehicle mass that will be a frequently encountered scenario in real world application.
\section{Lateral Control of Vehicle}
\label{lateral_control}
As previously stated, path tracking is a problem majorly focused on lateral control than longitudinal control. Hence, we separately tackled this problem after achieving the necessary performance on the longitudinal velocity control. From the literature review, it was shown that modeling of the vehicle holds critical importance before defining the controller itself. The following section describes various vehicle models taken into consideration.

\subsection{Vehicle Modelling}
\begin{figure}[!tbh]
\centering
\begin{subfigure}[Geometric Bicycle Model]
{
  \includegraphics[width=36.6mm]{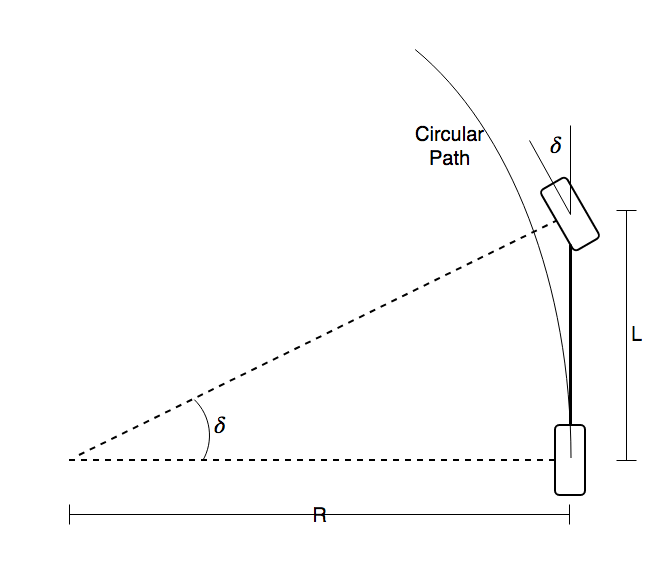}
  \label{fig:geoBiMo}
}
\end{subfigure}
\begin{subfigure}[Kinematic Bicycle Model]
{
  \includegraphics[width=36.6mm]{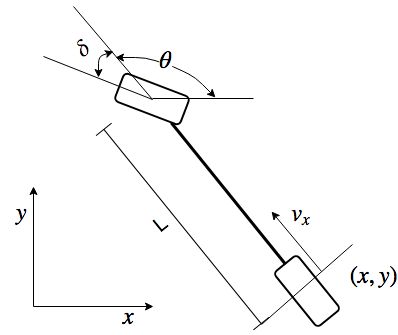}
  \label{fig:kinematicModel}
}
\end{subfigure}
\caption{}
\end{figure}
Different path tracking algorithms rely on various vehicle models depending upon the level of involvement of lateral dynamics into the equation of motion. We rederive these models here, only to an extent, that is required by the presented path tracking methods in the subsequent section.

\subsubsection{Geometric Vehicle Model}

Bicycle model approximation is a popular method to handle ackermann steering vehicles. Since we are considering only the geometrical aspects here, this kind of assumption which may violate some vehicle dynamics can be considered as a good enough approximation. The model assumes a two wheeler approximation to the vehicle assuming it to be symmetric about the central longitudinal axis. Another assumption is that the vehicle navigates in a plane so that we can neglect three-dimensional entities which are much difficult to operate with. Figure \ref{fig:geoBiMo} shows result of above mentioned approximations and assumptions.

\noindent Here,

$\delta$ be the steering angle

$R$ be the turning radius

$L$ be the wheel base

\begin{equation} 
tan(\delta) = \frac{L}{R} 
\end{equation} 

\subsubsection{Kinematic Vehicle Model}

For some path tracking models directed towards better performance than geometry-based path trackers, kinematics of the vehicle is included. In order to get the vehicle back to path at higher speeds, a low steering angle suffices whereas, for lower speeds, it is required to steer hard towards the path. This explains the importance of involving kinematics for vehicle modeling. Figure \ref{fig:kinematicModel} shows the variable assignments for model derivation.\\

\noindent Here,

$v_{x}$ be the longitudinal velocity

$\theta$ be the heading of the vehicle

$(x, y)$ be the real axle position \\

\noindent Intuitively from the Figure \ref{fig:kinematicModel}, we can write the expressions,

\begin{equation} R = \frac{v_{x}}{\dot{\theta}} \end{equation}
\begin{equation} \frac{v_{x} tan(\delta)}{L} = \frac{v_{x}}{R} \end{equation}
\begin{equation} tan(\delta) = \frac{L}{R} \end{equation}

\noindent Considering the bicycle model in the x-y plane with $v_{x}$ as rear-wheel axle velocity or longitudinal velocity and $(x, y)$ its instantaneous position, we can write

\begin{equation} \dot{x} = v_{x} cos(\theta) \end{equation}
\begin{equation} \dot{y} = v_{x} sin(\theta) \end{equation}

\noindent It is beneficial to write these equations in a condensed state-space model format for an easy practical implementation of control laws, 

\begin{equation}
\begin{bmatrix}
\dot{x} \\
\dot{y} \\
\dot{\delta} \\
\dot{\theta}
\end{bmatrix} 
=
\begin{bmatrix}
cos(\theta) \\
sin(\theta) \\
0 \\
\frac{tan(\delta)}{L}
\end{bmatrix} v
+
\begin{bmatrix}
0 \\
0 \\
1 \\
0
\end{bmatrix} \dot{\delta}
\end{equation}

\noindent where $\dot{\delta}$ and $v$ are the angular velocity of steered wheel and longitudinal velocity respectively.

\subsubsection{Dynamic Vehicle Model}
\begin{figure*}[!tbh]
\centering
\begin{subfigure}[Dynamic bicycle model]
{
  \includegraphics[width=50mm]{Figs/dynamic_force_diagram.png}
  \label{fig:dynModel}
}
\end{subfigure}
\begin{subfigure}[Relative coordinate system between vehicle and path]
{
  \includegraphics[width=50mm]{Figs/dynamic_model_path_coord.png}
  \label{fig:dynPathCoord}
}
\end{subfigure}
\begin{subfigure}[Plot of lateral force on the wheel vs. slip angle]
{
  \includegraphics[width=50mm]{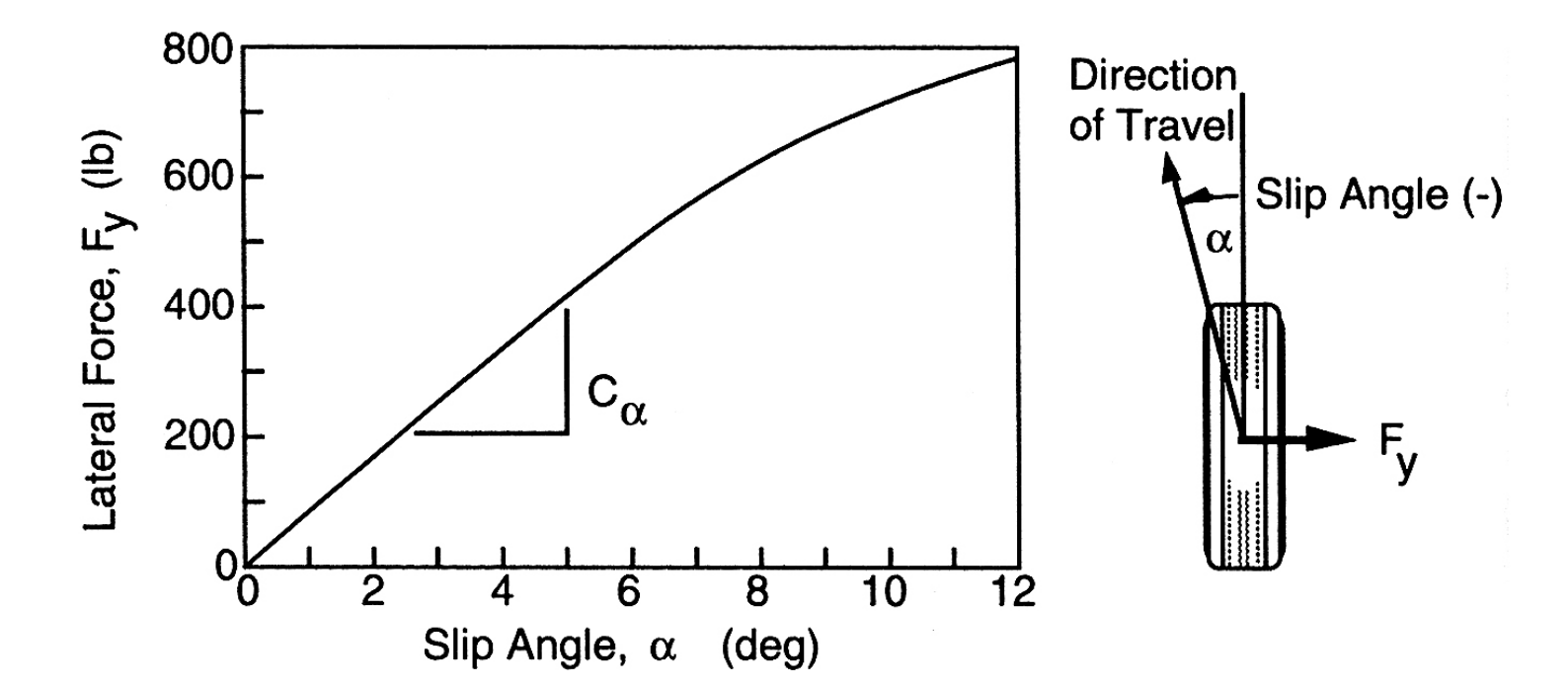}
  \label{fig:slipAngle}
}
\end{subfigure}
\caption{}
\end{figure*}
Forces between the wheels of the vehicle and the road surface play a significant role in path tracking capabilities at a higher speed. Hence, it becomes important to derive vehicle model by also considering newton's law of motion. Given that the path tracking problem is a problem in lateral control, the lateral forces acting on the vehicle must be the dominating forces. Like the previous sub-sections, we assume the longitudinal velocity is controlled separately. Figure \ref{fig:dynModel} shows different forces acting on the car.

%\begin{figure}[!tbh]
%\centering
%\includegraphics[width=70mm]{Figs/dynamic_force_diagram.png} 
%\caption{Dynamic bicycle model}
%\label{fig:dynModel}
%\end{figure}

\noindent Here, 

$F_{fx}$, $F_{fy}$ be the forces acting on front wheel

$F_{rx}$, $F_{ry}$ be the forces acting on the rear wheel

$v_x$, $v_y$ be the longitudinal and lateral velocity respectively

$I_z$ be the moment of inertia about CG of vehicle

$\omega_{z}$ be the angular velocity of the vehicle in the x-y plane 

$l_f$, $l_r$ be the distance of CG from the front wheel and rear wheel respectively \\ 

\noindent Writing the forces in lateral direction of vehicle, given the mass $m$ of vehicle,
\begin{equation} F_{ry} + F_{fy} cos(\delta) - F_{fx} sin(\delta) = m \dot{v_{y}} + m v_{x} \omega_{z} \end{equation}

\noindent Balancing yaw moments,
\begin{equation} F_{fy} l_{f} cos(\delta) - F_{ry} l_{r} + F_{fx} l_{r} sin{\delta} = I_{z} \dot{\omega_{z}} \end{equation}

\noindent Using the definition of slip angle of wheel tires, we can write
\begin{equation} \alpha_{r} = tan^{-1}(\frac{v_{y} - l_{r} \omega_z}{v_{x}}) \end{equation}
\begin{equation} \alpha_{f} = tan^{-1}(\frac{v_{y} + l_{f} \omega_z}{v_{x}}) - \delta \end{equation}

\noindent where $\alpha_{r}$, $\alpha_{f}$ are the slip angle at rear wheel and front wheel respectively.

\noindent Figure \ref{fig:slipAngle} shows the typical variation of lateral force on the wheel against slip angle. \cite{doumiati2010method} in their study, show how to use the topology of this plot along with a method to calculate the linear constant.

\noindent For small angles, it is observed that the variation between lateral force and slip angle can be considered linear. As defined in the literature, the constant of proportionality between these two entities is defined as cornering stiffness $c_{r}$ and $c_{f}$. \\
\begin{equation} F_{ry} = - c_{r} \alpha_{r} \end{equation}
\begin{equation} F_{fy} = - c_{f} \alpha_{f} \end{equation}
\noindent Applying assumptions like small steering angle ($ \delta \approx 0 $) and constant longitudal velocity ($\dot{v_{x}} \approx 0$), we get the following set of equations after simplifying Eq. 4.8 and Eq. 4.9, \\
\begin{equation} \dot{v_{y}} = - \frac{c_{f} v_{y} + c_{f} l_{f} \omega_{z}}{m v_{x}}  + \frac{c_{f} \delta}{m} + \frac{-c_{r} v_{y} + c_{r} l_{r} \omega_{z}}{m v_{x}} - v_{x} \omega_{z} \end{equation}
\begin{equation} \dot{\omega_{z}} = \frac{l_{r} c_{r} v_{y} - l^2_{r} c_{r} \omega_{z}}{I_{z} v_{x}} - \frac{l_{f} c_{f} v_{y} + l^2_{f} c_{f} \omega_{z}}{I_{z} v_{x}} \omega_{z} + \frac{l_{f} c_{f}}{m} \delta \end{equation}\\
\noindent This can be further simplified to state-space model as also described in the study by \cite{snider2009automatic} and \cite{ryu2005state},

\begin{equation}
\begin{bmatrix}
\dot{v_{y}} \\
\dot{\omega_{z}}
\end{bmatrix} 
=
\begin{bmatrix}
-\frac{c_{f} + c_{r}}{m v_{x}} & \frac{l_{r} c_{r} - l_{f} c_{f}}{m v_{x}} - v_{x} \\
\frac{l_{r} c_{r} - l_{f} c_{f}}{I_{z} v_{x}} & - \frac{l^2_{f} c_{f} + l^2_{r} c_{r}}{I_{z} v_{x}}
\end{bmatrix}
\begin{bmatrix}
v_{y} \\
\omega_{z}
\end{bmatrix}
+
\begin{bmatrix}
\frac{c_{f}}{m} \\
\frac{l_{f} c_{f}}{m}
\end{bmatrix}
\delta
\end{equation}

\noindent It was essential to estimate the constant values like stiffness constant and weight distribution for the vehicle setup. We used the optimization technique, least squares method, as described in Vehicle Parameter Estimation section in the work \cite{ryu2005state}.

\noindent While the above model incorporates vehicle dynamics, it is also possible to include path dynamics in the state-space modeling process. Figure \ref{fig:dynPathCoord} shows a relative system defined between the vehicle and path to be followed. Inclusion of both vehicle dynamics and path dynamics could lead to a better response of vehicle as some prior knowledge of the path will be known. Again assuming constant longitudinal velocity ($\dot{v_{x}} \approx 0$), the yaw rate derived from the path ($\omega_{p}$) can be given as,

%\begin{figure}[!tbh]
%\centering
%\includegraphics[width=70mm]{Figs/dynamic_model_path_coord.png}
%\caption{Relative coordinate system between vehicle and path}
%\label{fig:dynPathCoord}
%\end{figure}

\begin{equation} \omega_{p}(s) = K(s) v_{x} \end{equation}

\noindent Similarly, lateral acceleration derived from the path can be given as,

\begin{equation} \dot{v_{y}}(s) = K(s) v^2_{x} \end{equation}

\noindent Let $e_{cg}$ be the cross-track error or the orthogonal distance between C.G of vehicle and path,

%\begin{equation} \ddot{e}_{cg} = (\dot{v_{y}} + v_{x} \omega_{z}) - \dot{v_{y}}(s) \end{equation}
%\begin{equation} \ddot{e}_{cg} = \dot{v_{y}} + v_{x}( \omega_{z} - \omega_p(s) \end{equation}
%\begin{equation} \ddot{e}_{cg} = \dot{v_{y}} + v_{x} \dot{\theta}_{e} \end{equation}
\begin{equation} \dot{e}_{cg} = \dot{v_{y}} + v_{x} sin(\theta_{e}) \end{equation}

\noindent where $\theta_{e} = \theta - \theta_{p}$. After plugging ($e_{cg}$, $\theta_{p}$) into Eq. 4.14 and Eq. 4.15, we get,

\begin{equation}
\begin{aligned}
\ddot{e}_{cg} = & - \frac{c_f + c_r}{m v_x} \dot{e}_{cg}
                + \frac{c_f + c_r}{m} \theta_e 
                + \frac{l_r c_r - l_f c_f}{m v_x} \dot{\theta}_e \\
                & + \frac{l_r c_r - l_f c_f}{m v_x} \omega_p 
                - v_x \omega_p + \frac{c_f}{m} \delta
\end{aligned}
\end{equation} 

\begin{equation}
\begin{aligned}
\ddot{\theta}_e = & \frac{l_r c_r - l_f c_f}{I_z v_x} \dot{e}_{cg} 
                + \frac{l_f c_f - l_r c_r}{I_z} \theta_e \\
                & - \frac{l^2_f c_f + l^2_r c_r}{I_z v_x}(\dot{\theta}_e 
                + \omega_p) + \frac{l_f c_f}{m} \delta 
                - \dot{\omega_p} \\
\end{aligned}
\end{equation}

\noindent Using the above equations, we can form the state-space model as,

\begin{equation}
\begin{aligned}[t]
\begin{bmatrix}
\dot{e}_{cg} \\
\ddot{e}_{cg} \\
\dot{\theta}_e \\
\ddot{\theta}_e
\end{bmatrix} 
= & \begin{bmatrix}
0 & 1 & 0 & 0 \\
0 & -\frac{c_f + c_r}{m v_x} & \frac{c_f + c_r}{m} & \frac{l_r c_r - l_f c_f}{m v_x} \\
0 & 0 & 0 & 1 \\
0 & \frac{l_r c_r - l_f c_f}{I_z v_x} & \frac{l_f c_f - l_r c_r}{I_z} & -\frac{l^2_f c_f + l^2_r c_r}{I_z v_x}
\end{bmatrix} 
\begin{bmatrix}
e_{cg} \\
\dot{e}_{cg} \\
\theta_{e} \\
\dot{\theta}_{e}
\end{bmatrix}
\\
& +
\begin{bmatrix}
0 \\
\frac{c_{f}}{m} \\
0 \\
\frac{l_{f} c_{f}}{I_{z}}
\end{bmatrix} \delta
+
\begin{bmatrix}
0 \\
\frac{l_{r} c_{r} - l_{f} c_{f}}{m v_{x}} - v_{x} \\
0 \\
-\frac{l^{2}_{f} c_{f} + l^{2}_{r} c_{r} }{I_{z} v_{x}}
\end{bmatrix} \omega_{p} 
\end{aligned}
\end{equation}
\subsection{Controller Synthesis}

The presented controllers have been widely researched in the past decade after the rise of autonomous vehicle technology. Their implementation remains complicated for industrial and real-world utilization. Following subsections present them in a simplified manner for the purpose of analysis and quick implementation.

\begin{figure}[!tbh]
\centering
\begin{subfigure}[Geometry for pure pursuit]
{
  \includegraphics[width=36.6mm]{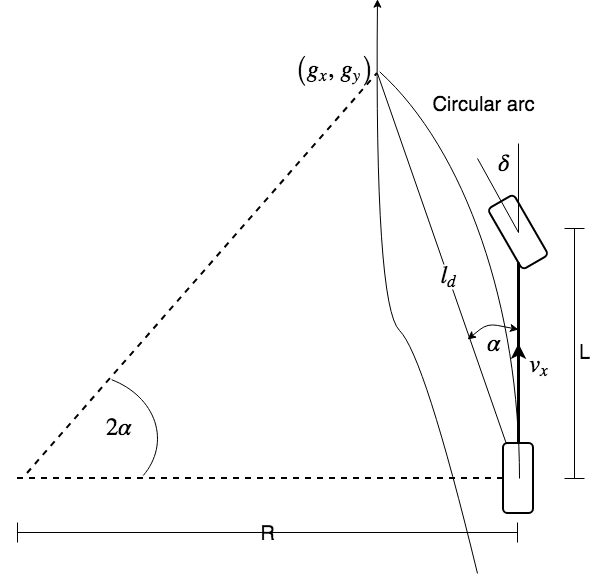}
  \label{fig:purePursuit}
}
\end{subfigure}
\begin{subfigure}[Geometry for stanley method]
{
 \includegraphics[width=36.6mm]{Figs/stanley_method.png}
 \label{fig:stanleyMethod}
}
\end{subfigure}
\caption{}
\end{figure}
\subsubsection{Pure Pursuit}

Pure Pursuit method is based on the geometric model of the vehicle we defined previously in vehicle modeling section. This model is known to approximate the tracking capability of the vehicle reasonably well at low speeds and up to moderate steering angles. In this method, the goal point on the path to be followed is determined by the look-ahead distance from the rear axle of the vehicle. Then, a circular arc that connects the rear axle of the vehicle and the goal point is determined. Figure \ref{fig:purePursuit} shows the construction of these geometrical entities.

%\begin{figure}[!tbh]
%\centering
%\includegraphics[width=70mm]{Figs/pure_pursuit.png} 
%\caption{Geometry for pure pursuit implementation}
%\label{fig:purePursuit}
%\end{figure}
\noindent Here,

$\alpha$ be the angle between vehicle's heading and look ahead distance vector

$R$ be the radius of curvature of the path

$l_{d}$ be the look-ahead distance \\

\noindent Applying the sine law,

$$ \frac{l_d}{sin(2\alpha)} = \frac{R}{sin(\frac{\pi}{2} - \alpha)} $$ 
$$ \frac{l_d}{2 sin(\alpha) cos(\alpha)} = \frac{R}{cos(\alpha)} $$
$$ \frac{l_d}{sin(\alpha)} = 2 R $$

\noindent Curvature $K$ can be defined as,

$$ K = \frac{1}{R} = \frac{2 sin(\alpha)}{l_d} $$

\noindent We had already shown steering angle relation with turning radius as,

$$ \delta = tan^{-1}(\frac{L}{R}) = tan^{-1}(K L) $$

\noindent After combining the above equations, we get,

\begin{equation} \delta = tan^{-1} \bigg( \frac{2 L e_{ld}}{l^2_d} \bigg) \end{equation}

\noindent where $e_{ld}$ is the lateral distance between the heading vector of the vehicle and goal point of the path.

\noindent Implementing this controller involved tuning of the parameter $l_d$, i.e., look ahead distance. It was simple to conclude that for the overall stability of the vehicle if the longitudinal velocity of the vehicle is higher, the look-ahead distance needs to be larger, and similarly, for lower longitudinal velocity the look-ahead distance will be smaller. Thus, it was necessary to express look-ahead distance as a function of velocity; for our purpose, we chose a linear function. So the final equation we used for tracking controller was,
\begin{equation} \delta = tan^{-1} \bigg( \frac{2 L e_{ld}}{ k v^2_x (t)} \bigg) \end{equation}
where $k$ is a tuning parameter. \\

\noindent Before implementing this control law on a simulated vehicle or passenger car, we validated the model's convergence using MATLAB. Figure \ref{fig:pure_conv} shows the quick reduction in cross-track error with time, thus allowing us to move ahead with this controller.

\subsubsection{Stanley Method}

This method comes from the robot system Stanley \cite{thrun2006stanley} that won the DARPA Grand Challenge. This method defines the steering control law as a non-linear function of the cross-track error $e_{fa}$ (taken from front axle here), that is, the distance between the front axle of the vehicle and the nearest point on the path. Intuitively, the law consists of two terms, one that helps vehicle align to the heading of nearest path point ($c_x$, $c_y$) and the other that reduce the cross-track error. From Figure \ref{fig:stanleyMethod}, we can write the control law as,

$$ \theta_e = \theta - \theta_p $$

\noindent Combining the two terms as per the definition,

\begin{equation}
\delta(t) = \theta_e(t) + tan^{-1} \bigg( \frac{k e_{fa}}{v_x (t)} \bigg)
\end{equation}

\noindent Comparing the above law with pure pursuit, there is an additional term, $\theta_e$, that makes all the difference. Intuitively, this term adds more stability as it tries to correct the heading. Hence, it was expected that this method performs better than the pure pursuit for our set of experiments.\\

\noindent Convergence test for Stanley Controller shows decent results as shown in \ref{fig:stanley_conv}

\subsubsection{PID Control}

We defined the control law very similar to the previously derived PID velocity control. For this control, we assigned the process variable as cross-track error $e_{cg}$, that is, the distance between the C.G of the vehicle and nearest path point. Similarly, the manipulated variable was assigned as the steering angle input $\delta$. Reference signal was considered constant of value 0, as for the path tracking capability we needed the cross-track error to converge.

\noindent Here,

$\delta(t)$ be the steering angle input to vehicle

$e_{cg}$ be measured cross track error

$r(t)$ be reference or set point

$e(t)$ be the error signal

$K_p$, $K_i$ $K_d$ be the proportional, integral and derivative constants respectively \\

\noindent Following equation was embedded into path tracking program,

\begin{equation}
\delta(t) = K_p e(t) + K_i \int e(t) dt  + K_d \dot{e}(t)
\end{equation}

\noindent However, the above equation was converted to discrete form for actual use

\begin{equation}
\delta(t) = K_p e(t) + K_i \sum_0^t e(t) + K_d \frac{e(t) - e(t-\Delta t)}{\Delta t} 
\end{equation} 

\noindent where $\Delta t$ is the sampling interval used for discretization.

\noindent We tuned the PID constant values for low cross-track error and different desired specification than previously defined. They were,
\begin{itemize}
\item Steady-state error  <  2\%
\item Overshoot  <  10\%
\item settling time  <  2 s
\end{itemize}

\begin{figure*}[!tbh]
\centering
\begin{subfigure}[Pure Pursuit's convergence of cross track error with time]
{
  \includegraphics[width=70mm]{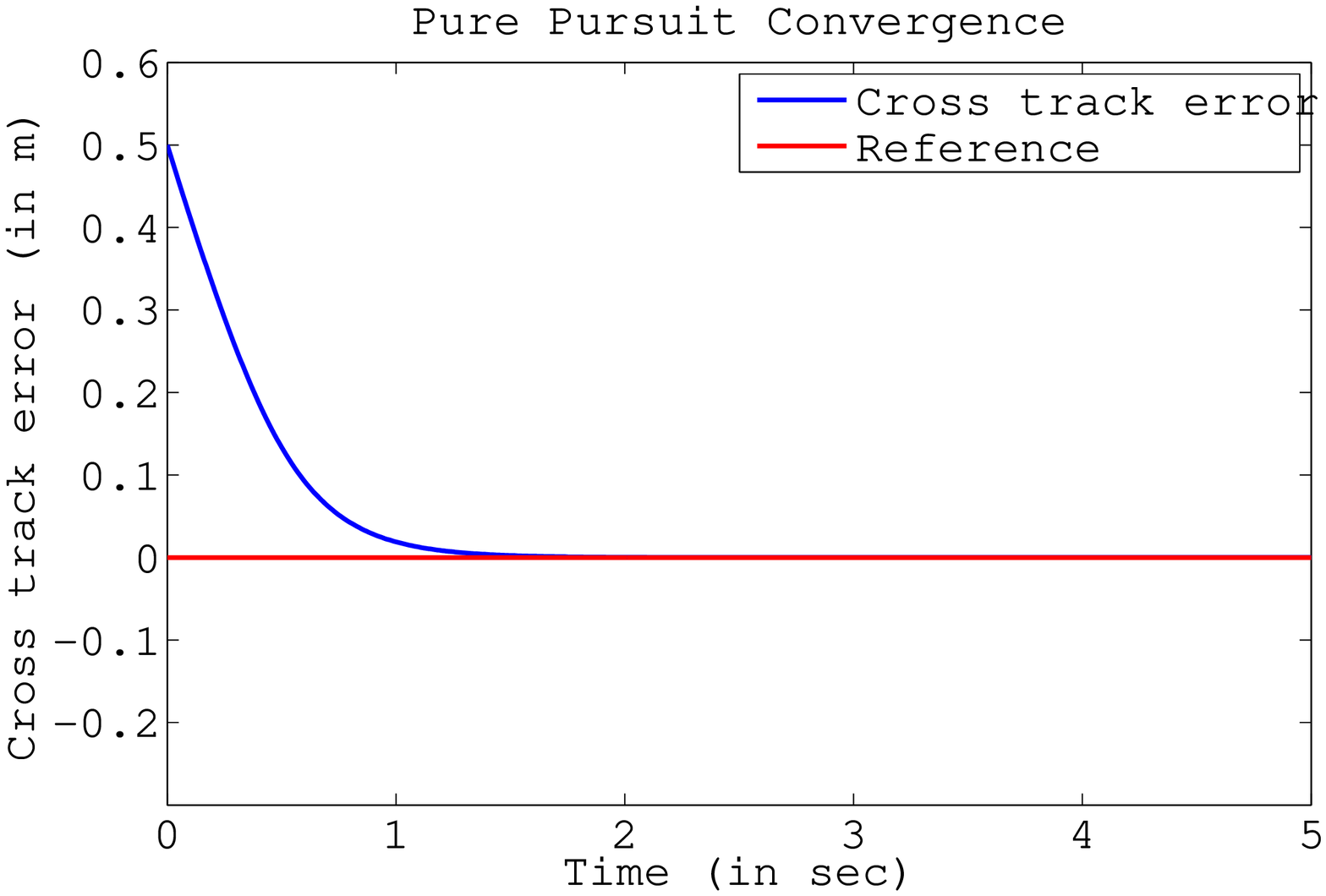}
  \label{fig:pure_conv}
}
\end{subfigure}
\begin{subfigure}[Stanley Method's convergence of cross track error with time]
{
  \includegraphics[width=70mm]{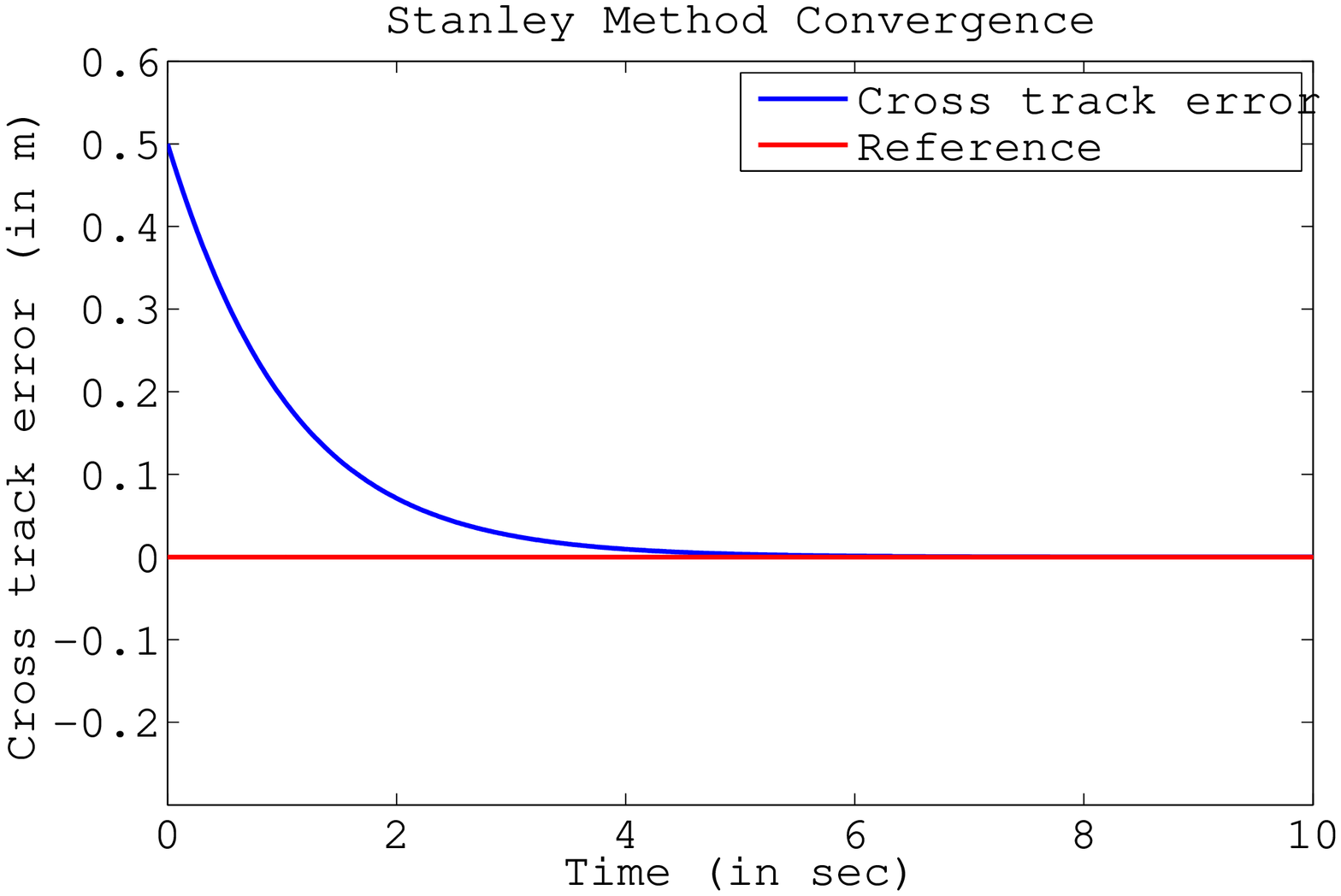}
  \label{fig:stanley_conv}
}
\end{subfigure}
\begin{subfigure}[PID controller's convergence of cross track error with time]
{
  \includegraphics[width=70mm]{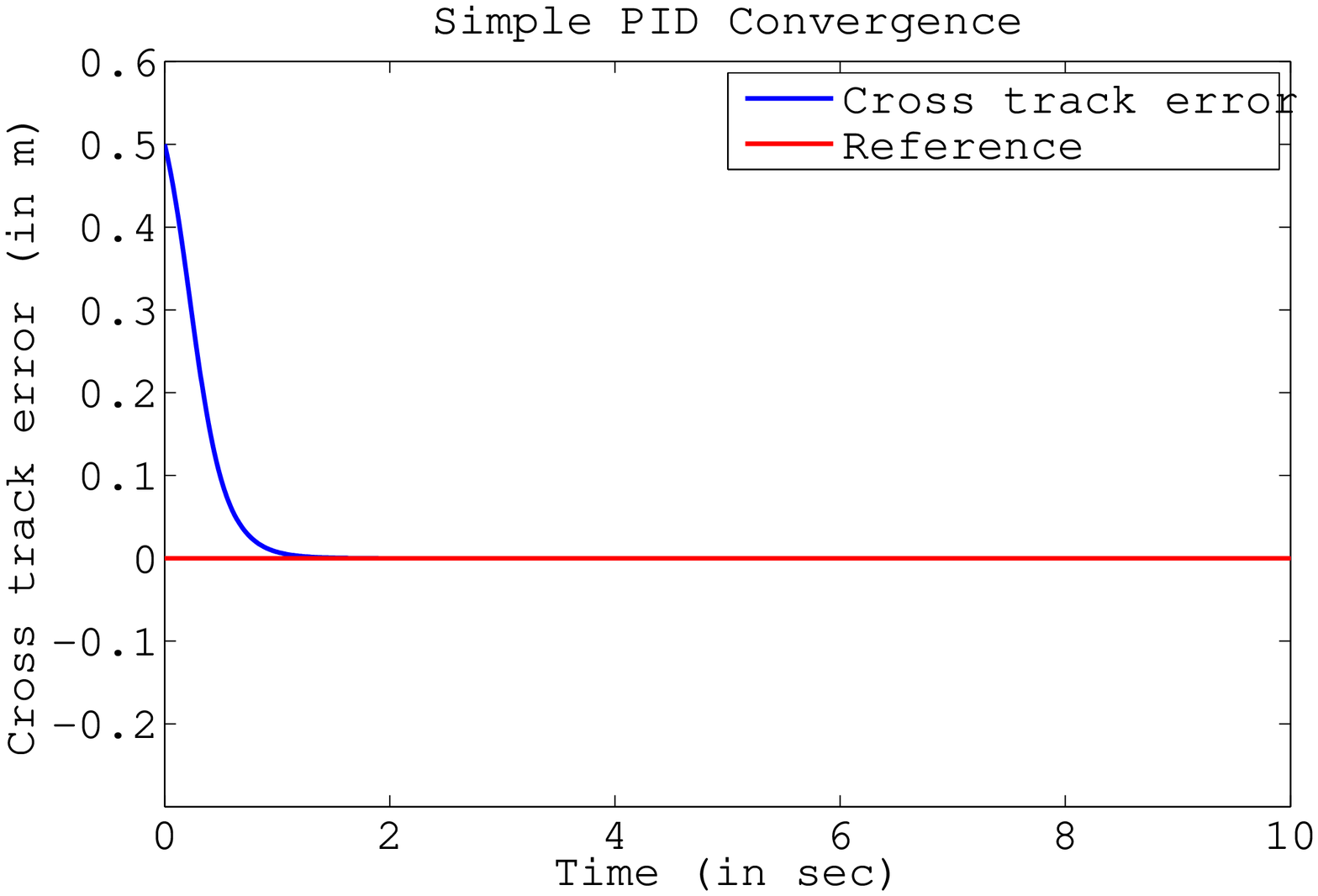}
  \label{fig:pid_conv}
}
\end{subfigure}
\begin{subfigure}[LQR controller's convergence of cross track error with time]
{
  \centering
  \includegraphics[width=70mm]{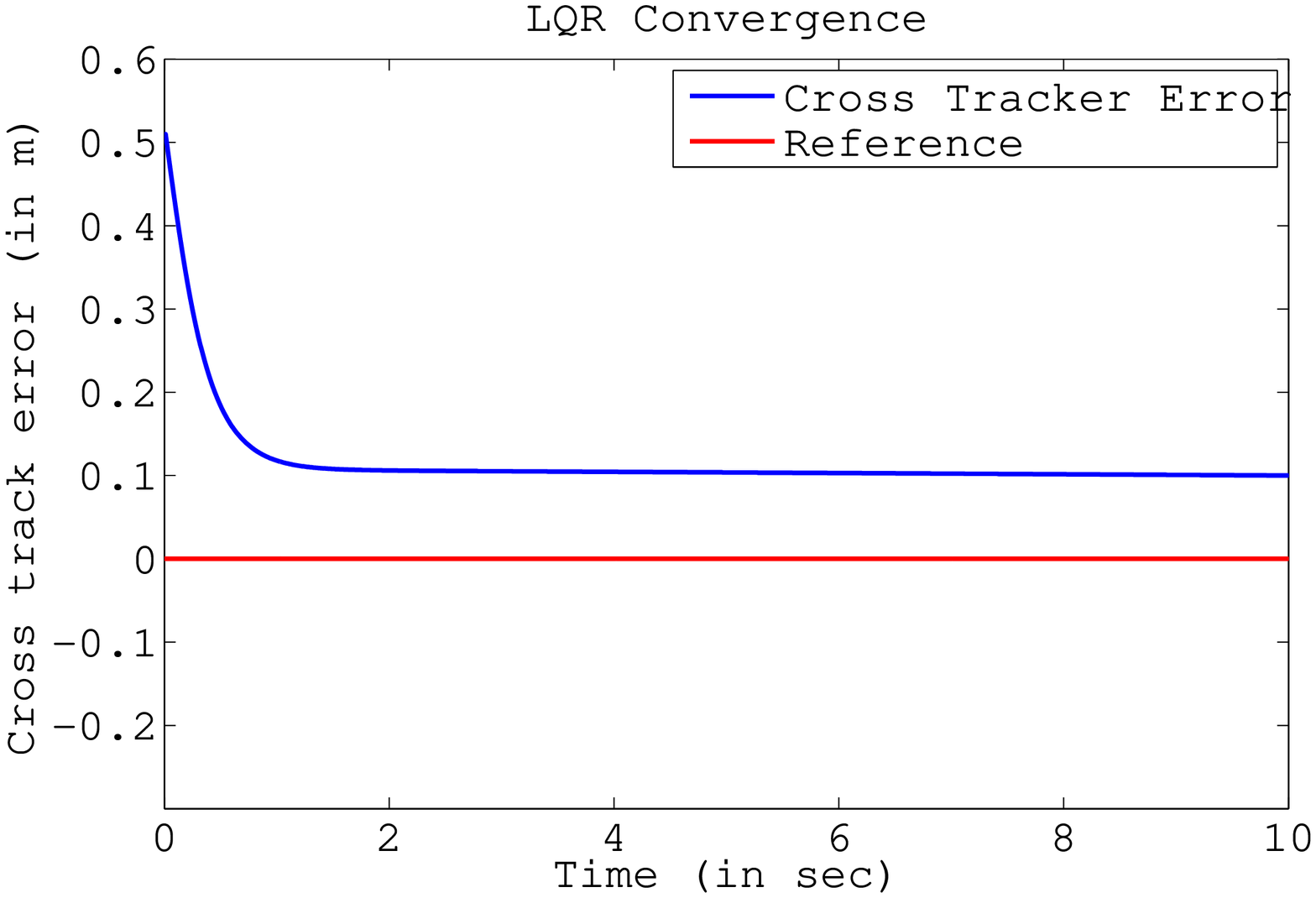}
  \label{fig:lqr_conv}
}
\end{subfigure}
\caption{}
\end{figure*}

\noindent PID controller showed a very quick convergence time \ref{fig:pid_conv} as compared to other methods as it can be tuned to be more aggressive while maintaining stability.

\subsubsection{Linear Quadratic Regulator}

LQR Control is one of the most popular controller practiced in the field of optimal control theory. This controller is concerned with operating a dynamic system (which in our case, the vehicle is) at minimum cost. The cost function is generally defined as a function of a property of the system that is targeted for attenuation. Similar to a PID controller or pole placement method, the true essence of applying this method is to find an optimal gain matrix. Here, the optimality refers to the situation where the gains are not too high to saturate the signal input to the actuator or too low to be vulnerable to instability. The method involves no tuning parameters except the weights associated with minimizing the cost function.

We utilized the dynamic model of the vehicle as previously defined for implementation of LQR method. Eq. 4.25 can be written in the standard form of linear state space equation as,

\begin{equation}
\dot{x} = A x + B \delta + C \omega_p
\end{equation}

\noindent where, A is the state transition matrix, B is the input matrix, and C is the residual matrix\\  
\\ \\$A = \begin{bmatrix}
0 & 1 & 0 & 0 \\
0 & -\frac{c_f + c_r}{m v_x} & \frac{c_f + c_r}{m} & \frac{l_r c_r - l_f c_f}{m v_x} \\
0 & 0 & 0 & 1 \\
0 & \frac{l_r c_r - l_f c_f}{I_z v_x} & \frac{l_f c_f - l_r c_r}{I_z} & -\frac{l^2_f c_f + l^2_r c_r}{I_z v_x}
\end{bmatrix}$

% \\ \\ \\ 
$ B = \begin{bmatrix}
0 \\
\frac{c_f}{m} \\
0 \\
\frac{l_f c_f}{I_z}
\end{bmatrix}$ 
$ C = \begin{bmatrix}
0 \\
\frac{l_r c_r - l_f c_f}{m v_x} - v_x \\
0 \\
-\frac{l^2_f c_f + l^2_r c_r }{I_z v_x}
\end{bmatrix}$ \\

\noindent All other variables hold the same terminology as defined in vehicle modeling section.

\noindent And we defined the cost function as per the standard in literature as,

\begin{equation}
J = \int x^T Q_s x + \int u^T Q_u u
\end{equation}

\noindent where $Q_s$ is the state cost matrix, and $Q_u$ is the input cost matrix.

For linear systems like $ \dot{x} = A x + B u$ it suffices to define the feedback law as $u = -G x$. However, we needed a feedforward term in addition to the feedback term to cancel out the residual term in the Eq. 4.31. So, we defined the feedback law as,

\begin{equation}
\delta = -G x + \delta_{ff}
\end{equation}

\noindent The study made by \cite{snider2009automatic} includes a procedure to obtain the appropriate expression for feedforward term for canceling such residual terms. Using the same method, the final state-space model can be obtained as

\begin{equation}
\dot{x} = (A - B G) x
\end{equation}

\noindent The above system is defined in continuous time domain. It was necessary to convert it to discrete time domain for actual implementation. For example, a simple system like $\dot{x} = A x + B u$ in discrete form can be presented as $x[n+1] = A_d x + B_d u$. Here, the continuous time domain matrices ($A$ and $B$) are discretized using zero-order hold method to matrices $A_d$ and $B_d$.\\

\noindent After following similar procedure as of continuous time domain, we get the ultimate expression for embedding into our program,

\begin{equation}
x[n+1] = (A_d - B_d G) x
\end{equation}

\noindent And the steering angle to update the vehicle position,

\begin{equation}
\delta = - G x + \frac{L}{R}
\end{equation} 

\noindent where $\delta_{ff} = \frac{L}{R}$ is the calculated feedforward term.\\

\noindent The tuning parameters used for this controller were,

$$ Q_s = 
\begin{bmatrix}
q_1 & 0 & 0 & 0 \\
0 & q_2 & 0 & 0 \\
0 & 0 & q_3 & 0 \\
0 & 0 & 0 & q_4
\end{bmatrix}
$$

\noindent where we defined $q_2 = q_3 = q_4 = 0$, implying that the cross track error $e_{cg}$ is only being weighted towards the control effort.\\
\noindent and $Q_u$ was chosen as,

$$ Q_u = 1 $$

\noindent to neglect the scaling between $Q_u$ and $Q_s$. \\

\noindent The convergence test for LQR shows the importance of the feed forward term as we observe a steady state crosstrack error developing with the simple LQR controller without feed forward term.

\subsubsection{Model Predictive Control}

Model Predictive Control (MPC) is another optimal control algorithm proficient with multivariable systems where manipulated variables and outputs are subjected to specific constraints in an optimized fashion. MPC follows a prolonged success in process industries and therefore is rapidly accepted by the community in domains such as automotive, aerospace, city resources planning, and financial engineering. MPC can be considered to be a variant of LQR control strategy with additional features like constraints on manipulated variables (input to a plant) and optimization over $N$ future steps instead of infinity. Recently, an increase has been observed in employing MPC for tasks of robot navigation \cite{schwarting2017parallel,kim2011control,katrakazas2015real}. The works which used MPC for steering control tasks in autonomous driving include \cite{babu2018model,borrelli2005mpc}. While the focus in \cite{babu2018model} was on the inclusion of actuator dynamics and its effect in overtaking and lane change maneuvres, \cite{borrelli2005mpc} focused on the task of following a reference trajectory similar to our case.

The name suggests the manner in which the control output is computed. At the present time $t$, the behavior of the system over the horizon $N$ is considered. The model predicts the state over the horizon using control outputs. The cost function is thereafter minimized, satisfying the given constraints and control output for the next $N$ steps are calculated. Only the first computed predicted output is used. At the time $t+1$, the computations are repeated with the horizon shifted by a one-time interval. The mathematical formulation can be given as,

\begin{equation}
U_{t}^{*}(x(t)) : = argmin \sum_{k=0}^{N-1} q(x_{t+k}, u_{t+k})
\end{equation}

\noindent subjected to,
\begin{subequations}
\begin{align}
& x_t = x(t) \label{eq:subeq1} \\
& x_{t+k+1} = A x_{t+k} + B u_{t+k}\label{eq:subeq2} \\
& x_{t+k}   \epsilon  \mathit{X} \label{eq:subeq3}\\
& u_{t+k}  \epsilon \mathit{U} \label{eq:subeq4}\\
& U_{t} = \{ u_t, u_{t+1},..., u_{t+N-1} \} \label{eq:subeq5}
\end{align}
\end{subequations}

\noindent The formulation uses prediction model \ref{eq:subeq2} to estimate possible state for $N$ iterations. Equations \ref{eq:subeq3} and \ref{eq:subeq4} define the state and input constraints respectively. For our purpose, the model is defined based on kinematic bicycle model as for other methods described in previous subsections. This includes state of six parameters as $x-y$ position, $\theta$ orientation, $v_x$ longitudinal velocity, $cte$ cross track error and $\psi$ orientation error.

\noindent Model is defined as,
% forgot to define the function f(x(t))
$$ x_{t+1} = x_{t} + v_{x} cos(\theta_{t}) * dt $$
$$ y_{t+1} = y_{t} + v_{x} sin(\theta_{t}) * dt $$
$$ \theta_{t+1} = \theta_{t} + \frac{v_x}{L} \delta_t * dt $$
$$ v_{t+1} = v_x $$
$$ cte_{t+1} = f(x_{t}) - y_{t} + (v_x sin(\psi_t) * dt) $$
$$ \psi_{t+1} = \theta_{t} - \theta_{p} + (\frac{v_x}{L} \delta_t * dt) $$

\noindent While the model is non-linear, the control method is still called linear MPC as the resulting cost function is convex. The distinction between Non-Linear MPC and MPC does not reside in the fact that the model of the system is linear or non-linear, but the fact that the resulting cost function is convex or non-convex.

\noindent The cost function is defined such that it minimizes the error in position, orientation along with penalizing more use and a sudden change of steering. The cost function is calculated over the horizon of N time intervals. All the terms are weighted by their respective constants. Our work uses the IPOPT library for optimizing the cost function. It is an implementation of the primal-dual point-integral algorithm with a filter line-search method for non-linear programming \cite{wachter2006implementation}. The further explanation of the optimization process is beyond the scope of this work.\\ 
\begin{equation}
\begin{aligned}
J = & \sum_{t=1}^{N} ( w_{cte} \| cte_t \|^2 + w_{\psi} \| \psi_t \|^{2} ) +                  \sum_{t=1}^{N-1} w_{\delta} \| \delta_t \|^2 \\
    & + \sum_{t=2}^{N} \| \delta_t - \delta_{t-1} \|^2 
\end{aligned}
\end{equation}

%----------------------------------------------------------------------------------------------------------------------------------------------

\subsection{Simulation Results and Discussion}

After the successful demonstration of control laws convergence in MATLAB, we went ahead with using the same laws for validating the results in a simulation environment. The results are in correspondence with the experimental setup defined in Chapter 2 for simulation.

\subsubsection{Pure Pursuit}
\begin{figure*}[!tbh]
\centering
\begin{subfigure}[Straight course with velocity 10 kmph]{
  \includegraphics[width=50mm]{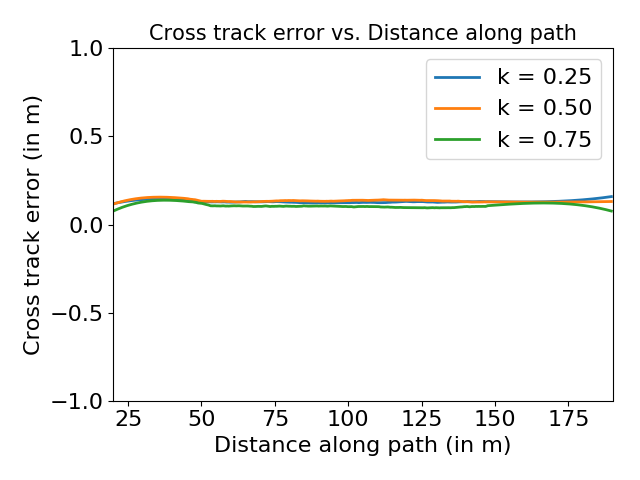}
  %\caption{Conventional feature based monocular visual odometry}
  } 
\end{subfigure}
\begin{subfigure}[Straight course with velocity 25 kmph]{
  \includegraphics[width=50mm]{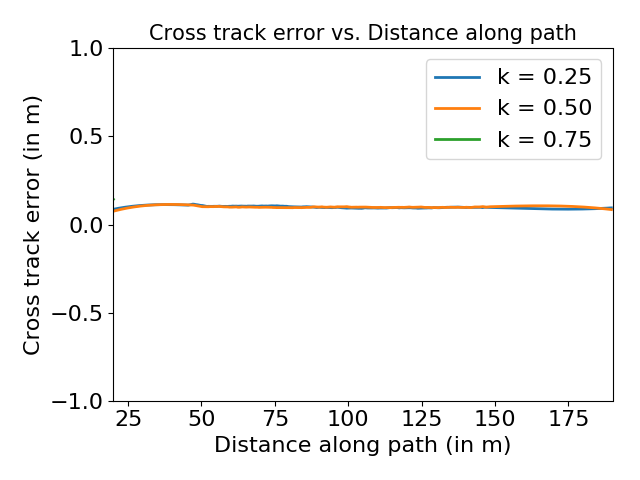}
  %\caption{Deep learning based end-to-end framework}
  }
\end{subfigure}
\begin{subfigure}[Straight course with velocity 35 kmph]{
  \includegraphics[width=50mm]{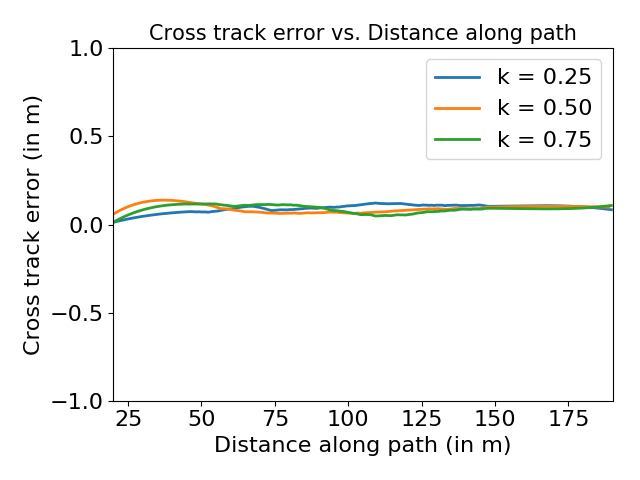}
  %\caption{Deep learning based end-to-end framework}
  }
\end{subfigure}
\begin{subfigure}[Circular course with velocity 10 kmph]{
  \includegraphics[width=50mm]{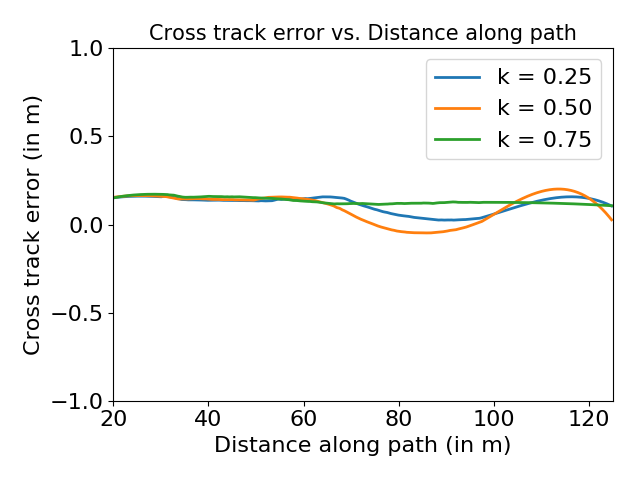}
  %\caption{Conventional feature based monocular visual odometry}
  } 
\end{subfigure}
\begin{subfigure}[Circular course with velocity 25 kmph]{
  \includegraphics[width=50mm]{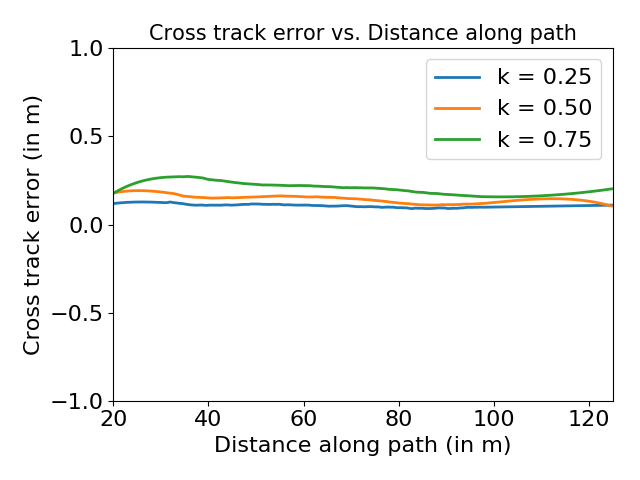}
  %\caption{Deep learning based end-to-end framework}
  }
\end{subfigure}
\begin{subfigure}[Circular course with velocity 35 kmph]{
  \includegraphics[width=50mm]{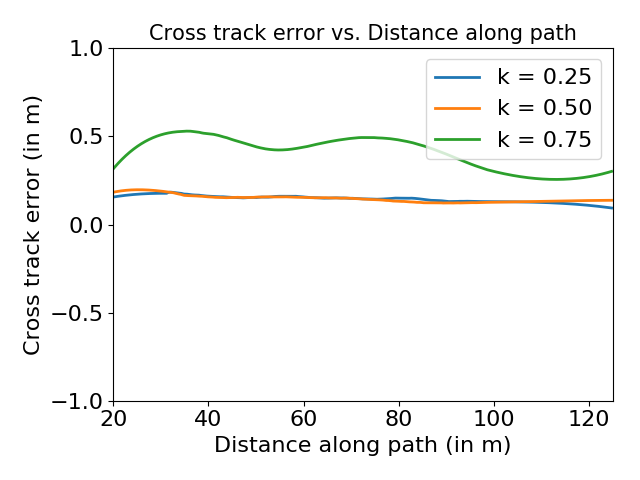}
  %\caption{Deep learning based end-to-end framework}
  }
\end{subfigure}
\begin{subfigure}[Lane change course with velocity 10 kmph]{
  \includegraphics[width=50mm]{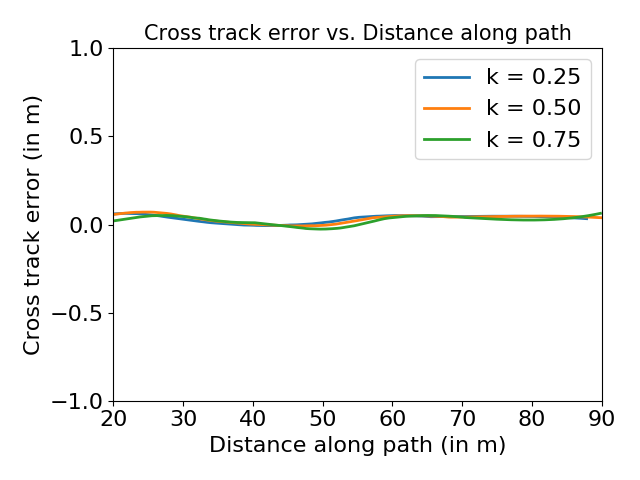}
  %\caption{Conventional feature based monocular visual odometry}
  } 
\end{subfigure}
\begin{subfigure}[Lane change course with velocity 25 kmph]{
  \includegraphics[width=50mm]{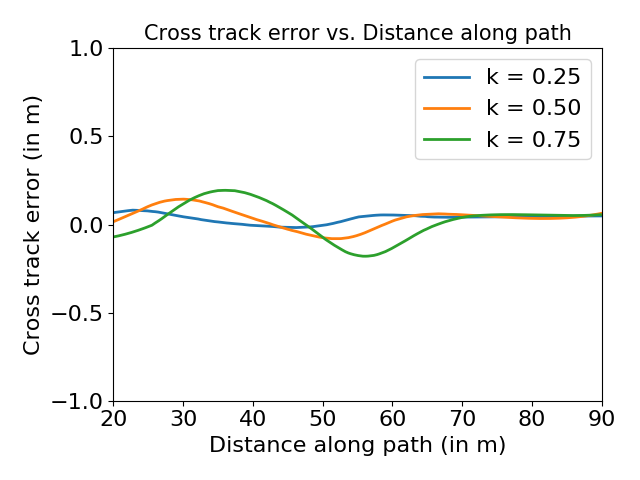}
  %\caption{Deep learning based end-to-end framework}
  }
\end{subfigure}
\begin{subfigure}[Lane Change course with velocity 35 kmph]{
  \includegraphics[width=50mm]{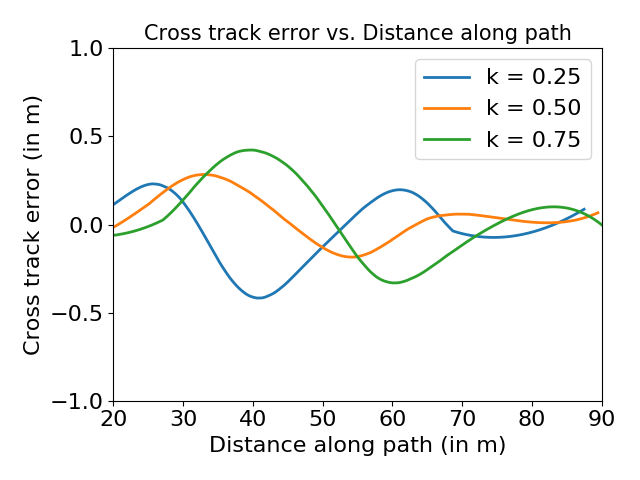}
  %\caption{Deep learning based end-to-end framework}
  }
\end{subfigure}
\begin{subfigure}[Sinusoidal course with velocity 10 kmph]{
  \includegraphics[width=50mm]{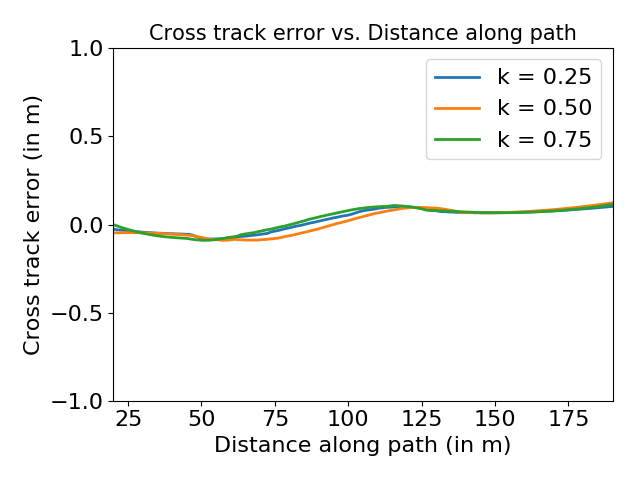}
  %\caption{Conventional feature based monocular visual odometry}
  } 
\end{subfigure}
\begin{subfigure}[Sinusoidal course with velocity 25 kmph]{
  \includegraphics[width=50mm]{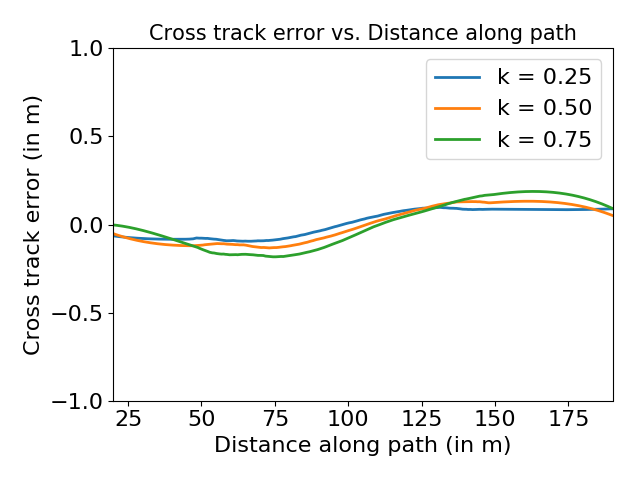}
  %\caption{Deep learning based end-to-end framework}
  }
\end{subfigure}
\begin{subfigure}[Sinusoidal course with velocity 35 kmph]{
  \includegraphics[width=50mm]{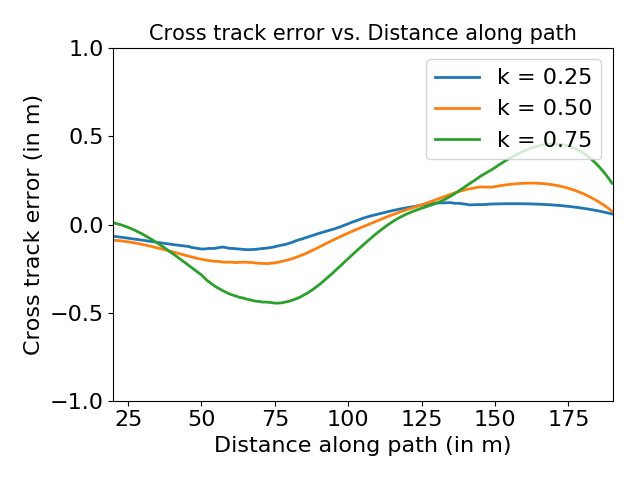}
  %\caption{Deep learning based end-to-end framework}
  }
\end{subfigure}
\caption{Pure Pursuit method's plots for cross track error vs. distance travelled on straight course(a,b,c), circular course(d,e,f), lane change course(g,h,i) and sinusoidal course(j,k,l)}
\label{fig:pure_pursuit}
\end{figure*}
\begin{figure*}[!tbh]
\centering
\begin{subfigure}[Straight course with velocity 10 kmph]{
  \includegraphics[width=50mm]{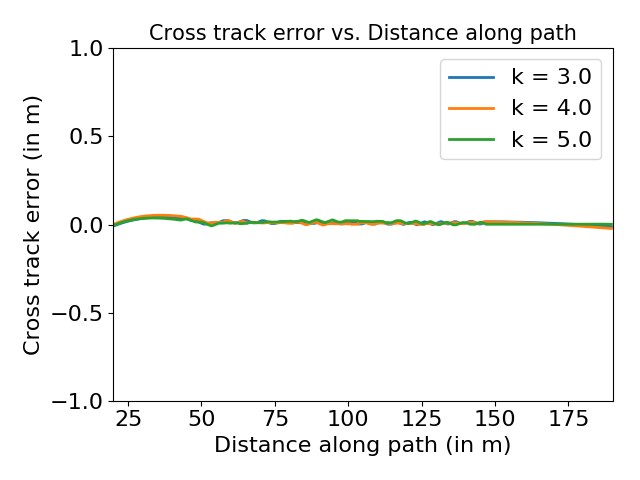}
  %\caption{Conventional feature based monocular visual odometry}
  } 
\end{subfigure}
\begin{subfigure}[Straight course with velocity 25 kmph]{
  \includegraphics[width=50mm]{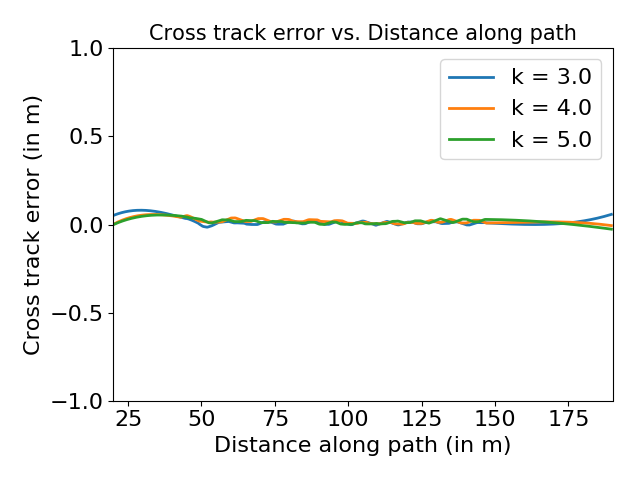}
  %\caption{Deep learning based end-to-end framework}
  }
\end{subfigure}
\begin{subfigure}[Straight course with velocity 35 kmph]{
  \includegraphics[width=50mm]{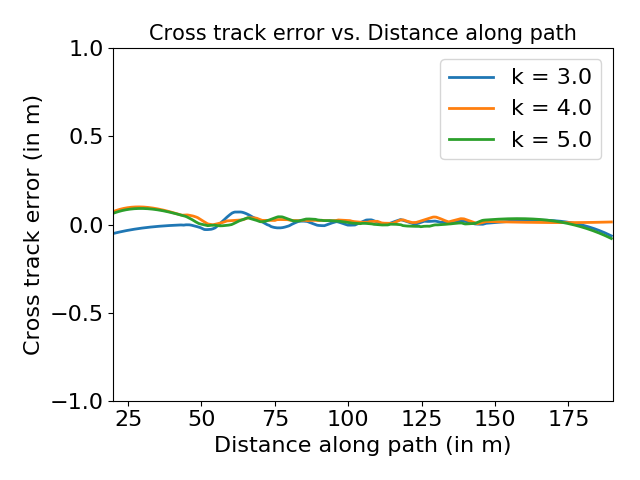}
  %\caption{Deep learning based end-to-end framework}
  }
\end{subfigure}
\begin{subfigure}[Circular course with velocity 10 kmph]{
  \includegraphics[width=50mm]{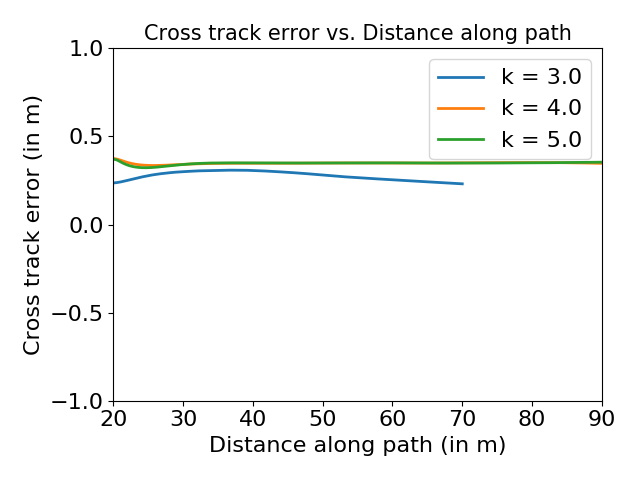}
  %\caption{Conventional feature based monocular visual odometry}
  } 
\end{subfigure}
\begin{subfigure}[Circular course with velocity 25 kmph]{
  \includegraphics[width=50mm]{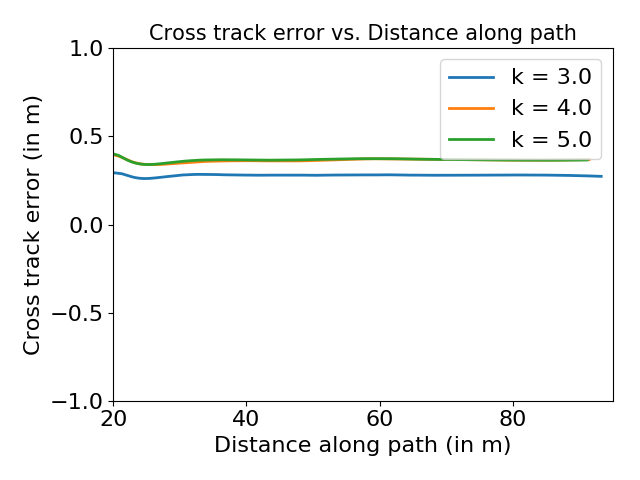}
  %\caption{Deep learning based end-to-end framework}
  }
\end{subfigure}
\begin{subfigure}[Circular course with velocity 35 kmph]{
  \includegraphics[width=50mm]{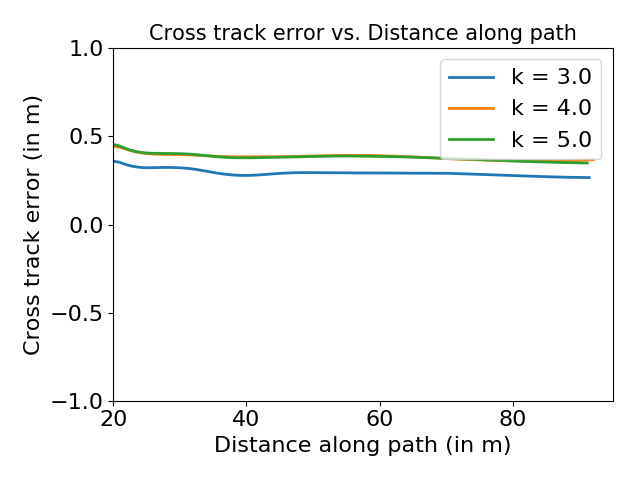}
  %\caption{Deep learning based end-to-end framework}
  }
\end{subfigure}
\begin{subfigure}[Lane change course with velocity 10 kmph]{
  \includegraphics[width=50mm]{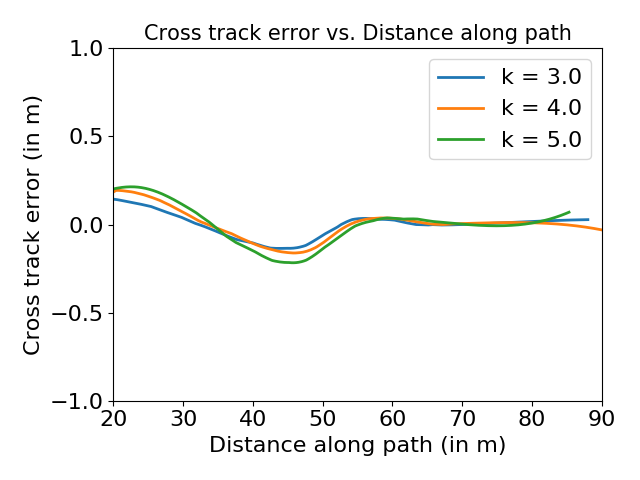}
  %\caption{Conventional feature based monocular visual odometry}
  } 
\end{subfigure}
\begin{subfigure}[Lane change course with velocity 25 kmph]{
  \includegraphics[width=50mm]{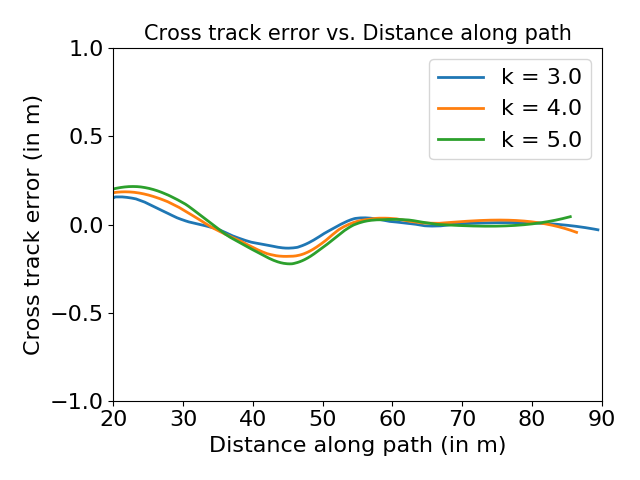}
  %\caption{Deep learning based end-to-end framework}
  }
\end{subfigure}
\begin{subfigure}[Lane Change course with velocity 35 kmph]{
  \includegraphics[width=50mm]{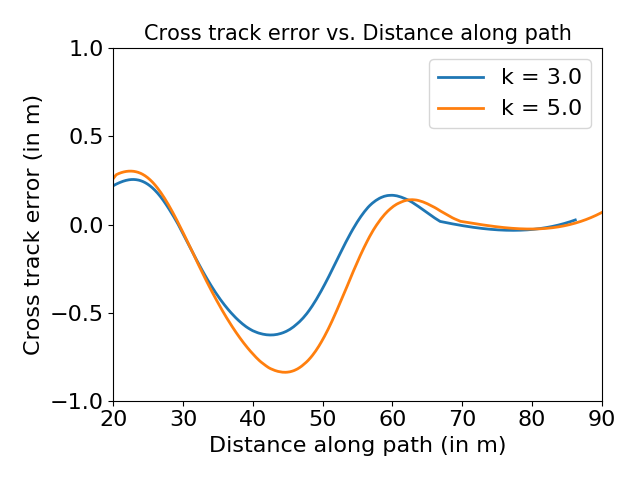}
  %\caption{Deep learning based end-to-end framework}
  }
\end{subfigure}
\begin{subfigure}[Sinusoidal course with velocity 10 kmph]{
  \includegraphics[width=50mm]{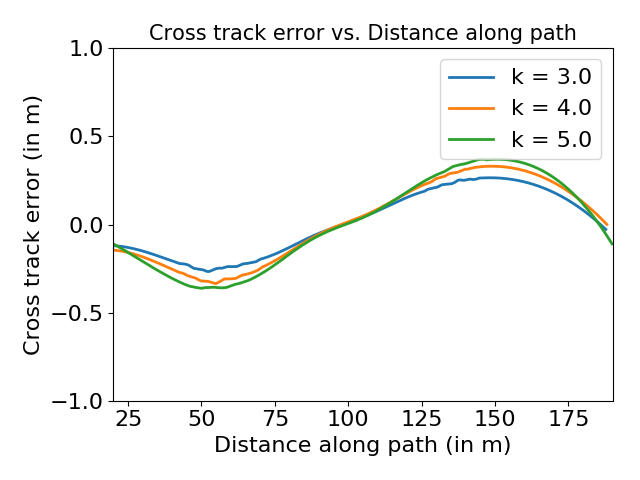}
  %\caption{Conventional feature based monocular visual odometry}
  } 
\end{subfigure}
\begin{subfigure}[Sinusoidal course with velocity 25 kmph]{
  \includegraphics[width=50mm]{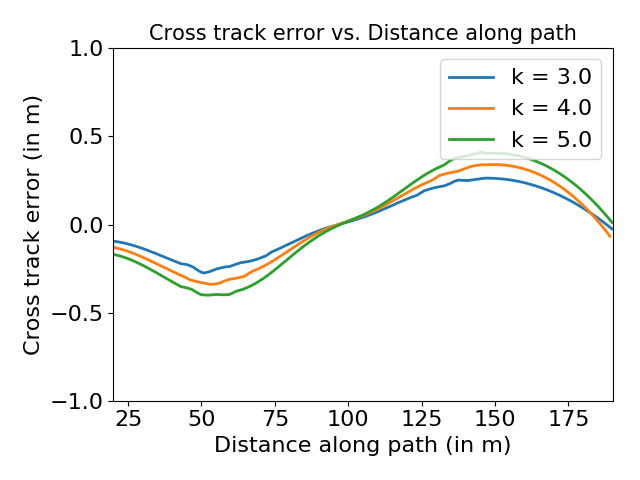}
  %\caption{Deep learning based end-to-end framework}
  }
\end{subfigure}
\begin{subfigure}[Sinusoidal course with velocity 35 kmph]{
  \includegraphics[width=50mm]{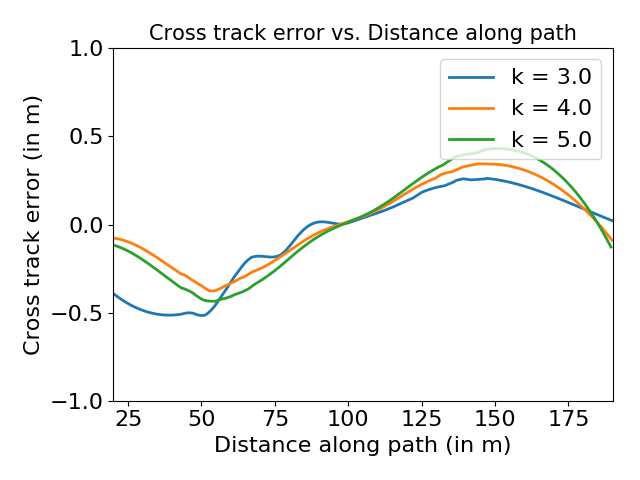}
  %\caption{Deep learning based end-to-end framework}
  }
\end{subfigure}
\caption{Stanley method's plots for cross track error vs. distance travelled on straight course(a,b,c), circular course(d,e,f), lane change course(g,h,i) and sinusoidal course(j,k,l)}
\label{fig:stanley}
\end{figure*}
Since the Pure pursuit method assumes a circular trajectory between the current position and the target point at a specific look ahead distance, the method was highly path specific. Initially, the look-ahead distance was taken to be a product of the longitudinal velocity and a constant. But when the results were observed, the parameter which was tuned for a specific velocity didn't work well for the other velocities. So after a lot of observations, it was concluded that the look-ahead distance should be,

$$ L_d = k * v_x + d $$

	The introduction of a fixed constant $d$ led to a significant improvement in the results. To find the value of $k$ and $d$, observations were made on the cross-track error on various paths and velocities for a fixed parameter which was to be tuned. After finding this parameter for different velocities, linear regression was applied to get a relation between the look-ahead distance and longitudinal velocity. 

The value of fixed constant $d$ was found out by trial-and-error, and the cross-track error was plotted for different paths and velocities by varying the value of $k$. Intuitively, it was thought that with an increase in the value of look-ahead distance, the car would take longer time to converge to the path, and hence the cross-track error would be more significant for the initial phase, and a steady-state error may be observed. By reducing the value of $k$, the tracking would become more aggressive. Reducing it beyond a particular value would lead to oscillations of the car about the path, and the car would not be able to return back to the path. 

On a straight course as observed from Figure \ref{fig:pure_pursuit} (a) (b) (c), there was a steady-state error of around 0.1m. Much change in results was not obtained by varying the value of $k$, and the error remained similar for all the values. This was due to the fact that the method calculated an infinite radius curve between the current position and the point at a look-ahead distance. The radius of curvature would remain infinity irrespective of the look-ahead distance due to the straight path and hence the controller would give an almost zero degree steering angle. For a constant radius circular path as observed from Figure \ref{fig:pure_pursuit} (d) (e) (f), the results were almost similar for low velocities irrespective of the value of $k$ within a certain range. However, at higher velocities, a significant change was observed by varying the values of $k$. For high speeds, the higher values of $k$ led to a very high steady-state error (~0.5m) as compared to that at lesser values of $k$ where the error was around 0.2m.

The sinusoidal path contained all types of features, including varying radius of curvature. Thus, if the car was correctly tuned for the sinusoidal path, it worked with decent performance on all the other paths as well. Specific features were observed in the cross-track error vs. distance along the path plot. The cross-track error jumped to a higher value when the car approached the maxima and minima of the curve, whereas the error was less in magnitude during the remaining path. The effect of $k$ was clearly visible for higher speeds as the cross-track error shooted to 0.5m at the maxima and minima of the curve. Since the pure pursuit controller assumes a circular path between two points, it didn't perform satisfactorily on a lane switching curve. The results were oscillatory, and errors were peaking up to 0.5m for high speeds. By increasing the value of $k$, oscillations were attenuated but increasing it beyond a specific value skipped the lane switching curve leading to a significant amount of error.

Thus, overall, the pure-pursuit algorithm is a good and computationally inexpensive geometrical method for path tracking. It works satisfactorily for low speed and simple curves but fails on higher speeds and complex curves. Moreover, the tuning parameters were observed to be weakly path-dependent as well.

\subsubsection{Stanley Method}

The Stanley controller contains two terms whose sum is used to calculate the steering angle. The first term is used to make the steering wheel parallel to the tangent to the closest point on the path, whereas the second term is used to steer the car inwards towards the path. The first term is the difference between the heading of the vehicle and the heading of the path, whereas the second term is dependent on the cross-track error needs to be tuned. It was observed that the value of the second term has some relation with velocity, but its contribution is very less, so we have not made the term velocity dependent for the simple reason of easy tuning. Since the testing of the vehicle was done below 35 kmph, the contribution has been ignored. Intuitively, it can be thought that the more the velocity of the vehicle, the less should be the extra steering angle towards the path; otherwise, it can lead to overshoot from the path. Thus, the velocity contribution if accounted for should be added in the denominator of the second term. Figure \ref{fig:stanley} shows the performance of this controller with tuning achieved within our capacity.

Tuning of the constant $k$ was done for the sinusoidal path as it contained varying radius of curvature. With the increase in the value of $k$, the second term magnitude decreased, and the vehicle was less responsive, and it took more time to converge to the path. On decreasing the value of $k$ below a certain optimum number, the oscillations caused by it were visible. Thus, by trial and error method, the constant $k$ was found out, and the results were plotted for three different values of $k$. The cross-track error plot was similar to the sinusoidal waveform. It reached maximum value corresponding to the maxima of the sine wave. Moreover, the magnitude of error was less than $0.5$ m in all the cases. The controller worked fairly for this path. For the circular path of a constant radius, a steady-state error of around 0.3-0.4 m was observed. To reduce this steady state constant error, the value of $k$ was reduced, but it lead to oscillations, and thus, the error could not be minimized further. Hence, the controller didn't work well for such a constant radius curve.
On a straight path, the results were quite good with the error remaining zero in almost all cases. The minor oscillations were also not observed. The lane change path was tracked quite well below 35 kmph. The effect of change in the value of $k$ was clearly visible in this case. The maximum error for $k=5$ was more significant than the maximum error for $k=3$. The maximum error remained below 0.5m.

Comparing the Stanley method with pure pursuit and PID, the controller worked comparatively better at higher velocities. Not much change was observed in the tracking results with the change in velocity except for the lane change case. Thus, this controller is better out of the three for high-speed velocities. It was also found to be path-dependent up to some extent.

\subsubsection{PID Control}
\begin{figure*}[!tbh]
\centering
\begin{subfigure}[Straight course with velocity 10 kmph]{
  \includegraphics[width=50mm]{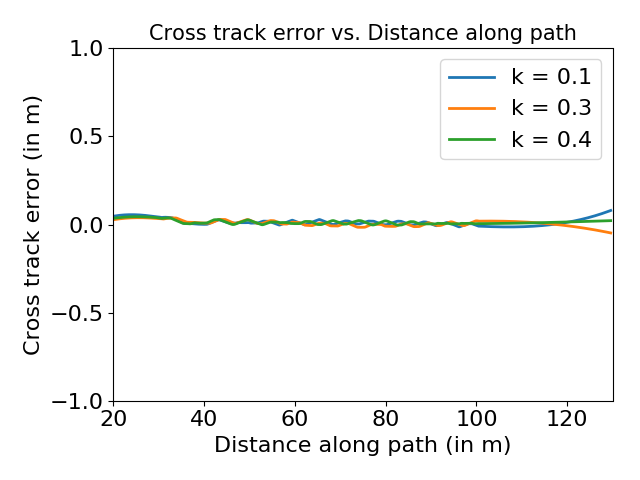}
  %\caption{Conventional feature based monocular visual odometry}
  } 
\end{subfigure}
\begin{subfigure}[Straight course with velocity 25 kmph]{
  \includegraphics[width=50mm]{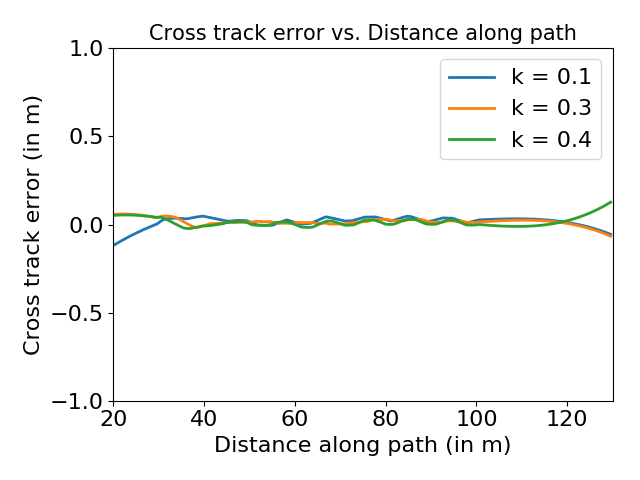}
  %\caption{Deep learning based end-to-end framework}
  }
\end{subfigure}
\begin{subfigure}[Straight course with velocity 35 kmph]{
  \includegraphics[width=50mm]{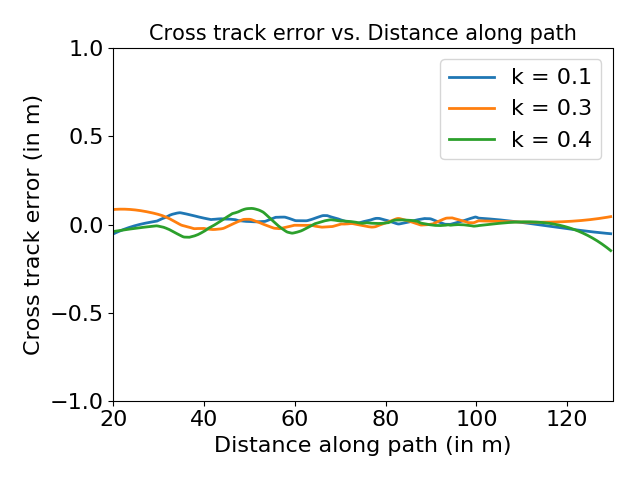}
  %\caption{Deep learning based end-to-end framework}
  }
\end{subfigure}
\begin{subfigure}[Circular course with velocity 10 kmph]{
  \includegraphics[width=50mm]{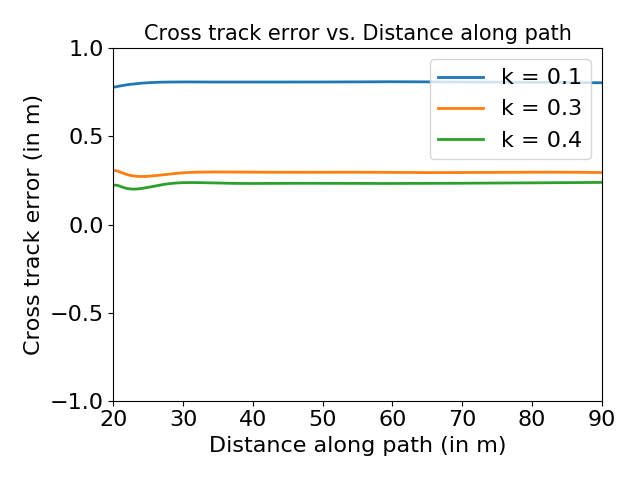}
  %\caption{Conventional feature based monocular visual odometry}
  } 
\end{subfigure}
\begin{subfigure}[Circular course with velocity 25 kmph]{
  \includegraphics[width=50mm]{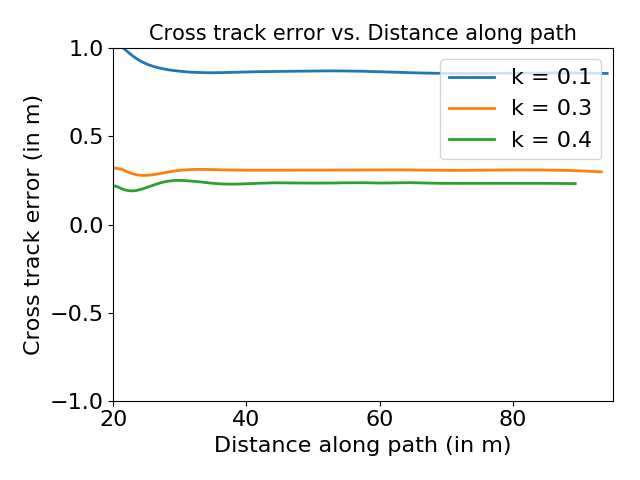}
  %\caption{Deep learning based end-to-end framework}
  }
\end{subfigure}
\begin{subfigure}[Circular course with velocity 35 kmph]{
  \includegraphics[width=50mm]{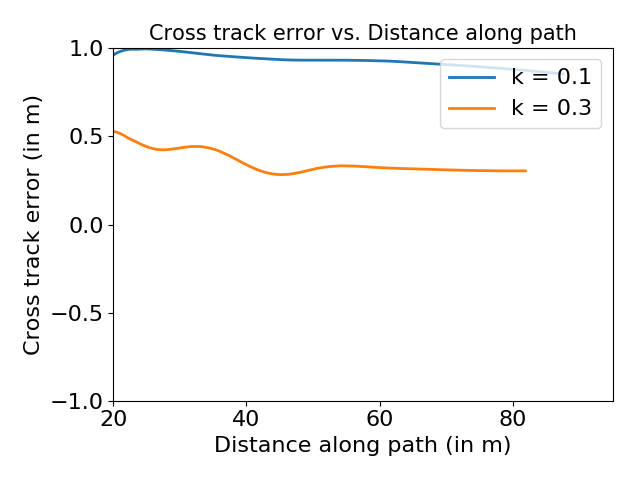}
  %\caption{Deep learning based end-to-end framework}
  }
\end{subfigure}
\begin{subfigure}[Lane change course with velocity 10 kmph]{
  \includegraphics[width=50mm]{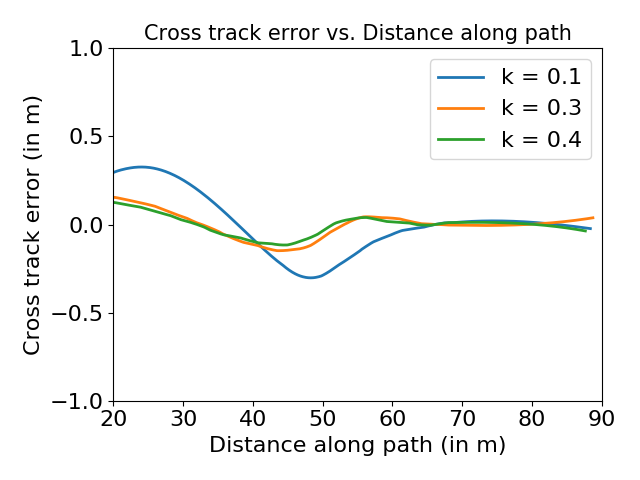}
  %\caption{Conventional feature based monocular visual odometry}
  } 
\end{subfigure}
\begin{subfigure}[Lane change course with velocity 25 kmph]{
  \includegraphics[width=50mm]{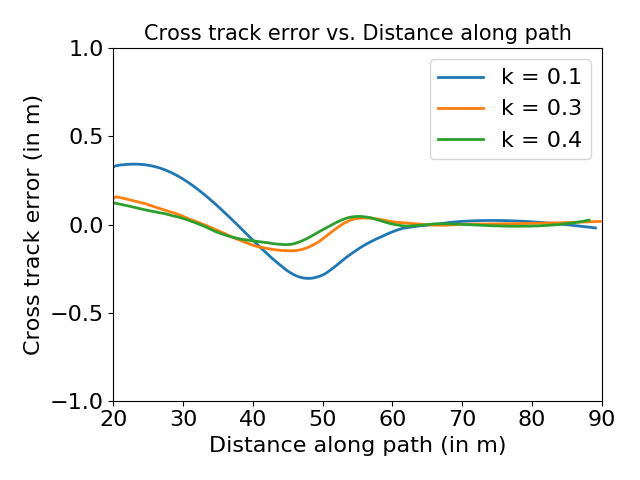}
  %\caption{Deep learning based end-to-end framework}
  }
\end{subfigure}
\begin{subfigure}[Lane Change course with velocity 35 kmph]{
  \includegraphics[width=50mm]{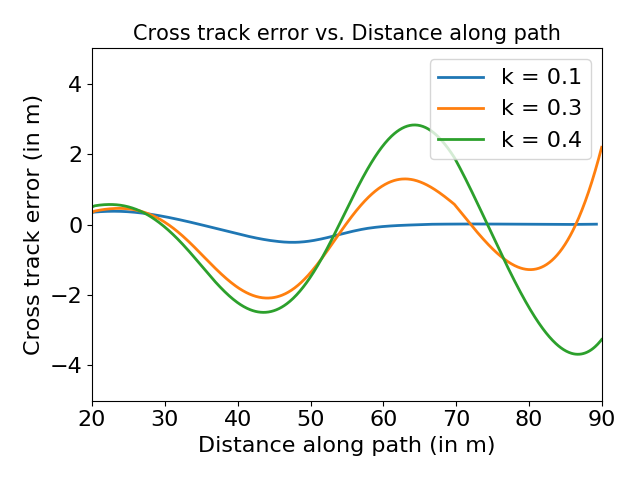}
  %\caption{Deep learning based end-to-end framework}
  }
\end{subfigure}
\begin{subfigure}[Sinusoidal course with velocity 10 kmph]{
  \includegraphics[width=50mm]{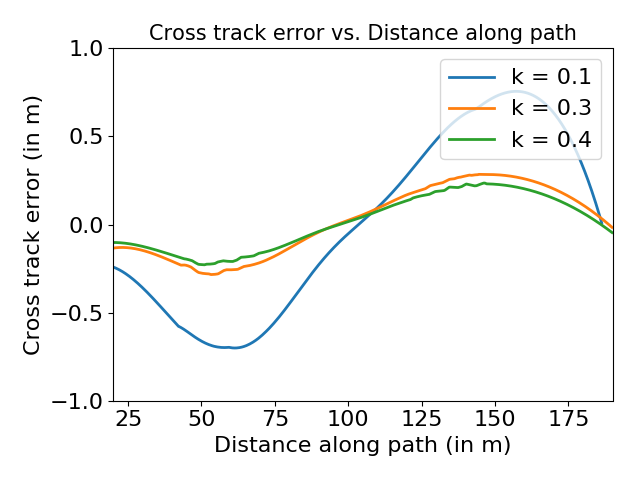}
  %\caption{Conventional feature based monocular visual odometry}
  } 
\end{subfigure}
\begin{subfigure}[Sinusoidal course with velocity 25 kmph]{
  \includegraphics[width=50mm]{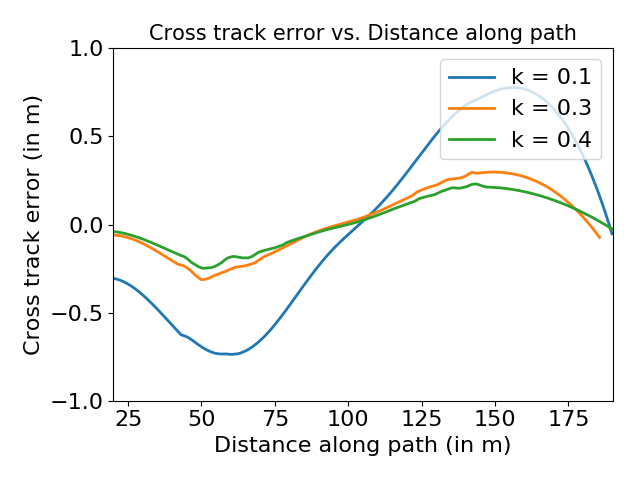}
  %\caption{Deep learning based end-to-end framework}
  }
\end{subfigure}
\begin{subfigure}[Sinusoidal course with velocity 35 kmph]{
  \includegraphics[width=50mm]{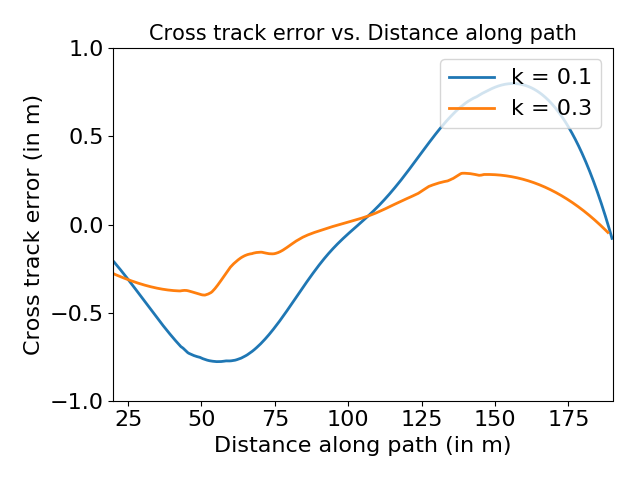}
  %\caption{Deep learning based end-to-end framework}
  }
\end{subfigure}
\caption{PID method's plots for cross track error vs. distance travelled on straight course(a,b,c), circular course(d,e,f), lane change course(g,h,i) and sinusoidal course(j,k,l)}
\label{fig:pid}
\end{figure*}
\begin{figure*}[!tbh]
\centering
\begin{subfigure}[Straight course with velocity 10 kmph]{
  \includegraphics[width=50mm]{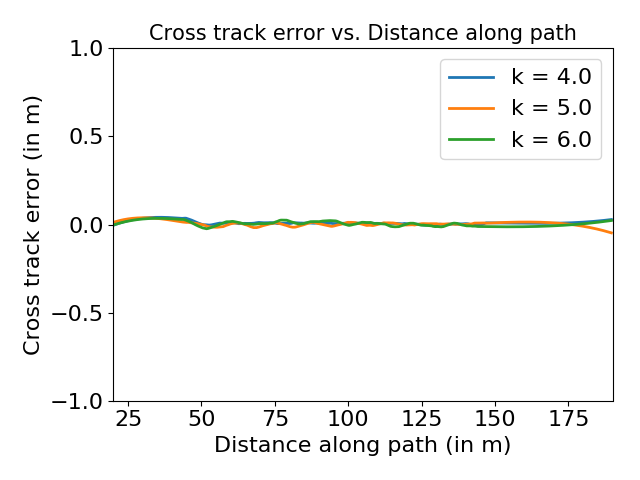}
  %\caption{Conventional feature based monocular visual odometry}
  } 
\end{subfigure}
\begin{subfigure}[Straight course with velocity 25 kmph]{
  \includegraphics[width=50mm]{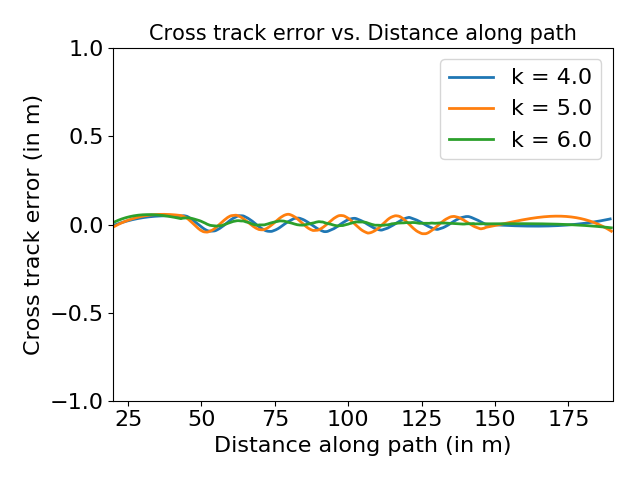}
  %\caption{Deep learning based end-to-end framework}
  }
\end{subfigure}
\begin{subfigure}[Straight course with velocity 35 kmph]{
  \includegraphics[width=50mm]{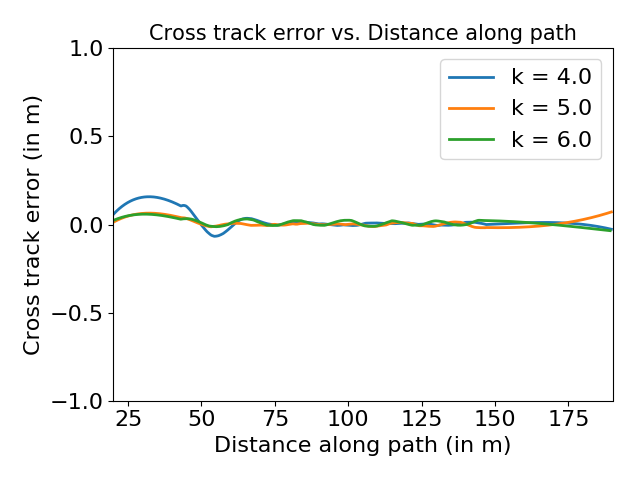}
  %\caption{Deep learning based end-to-end framework}
  }
\end{subfigure}
\begin{subfigure}[Circular course with velocity 10 kmph]{
  \includegraphics[width=50mm]{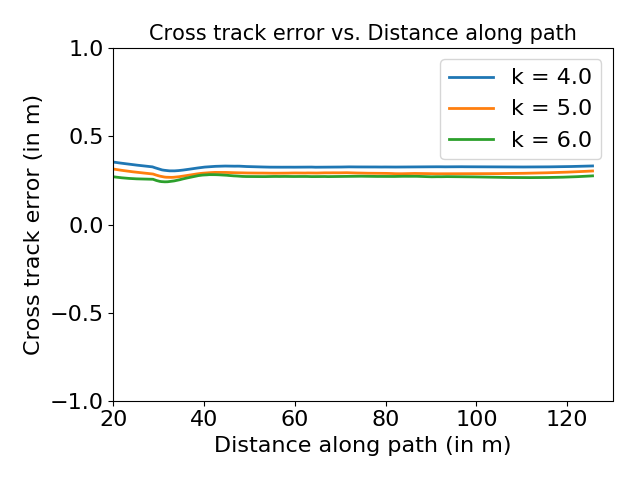}
  %\caption{Conventional feature based monocular visual odometry}
  } 
\end{subfigure}
\begin{subfigure}[Circular course with velocity 25 kmph]{
  \includegraphics[width=50mm]{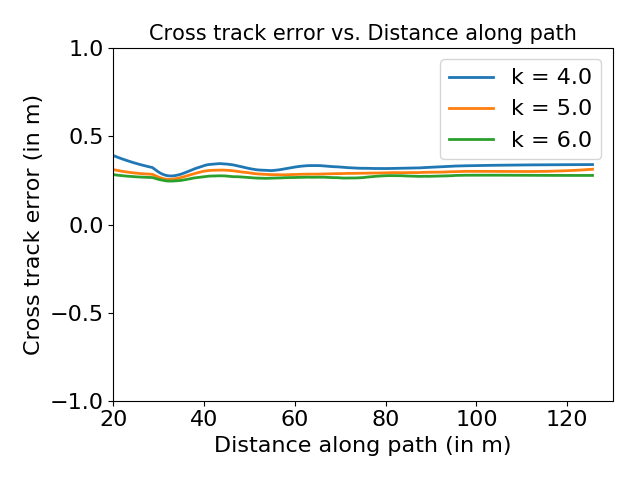}
  %\caption{Deep learning based end-to-end framework}
  }
\end{subfigure}
\begin{subfigure}[Circular course with velocity 35 kmph]{
  \includegraphics[width=50mm]{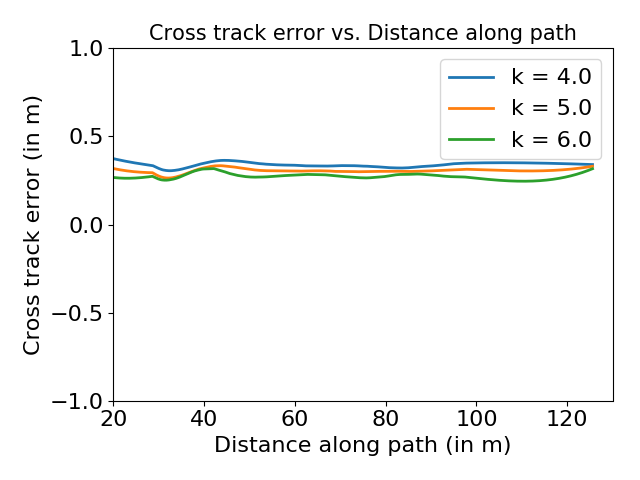}
  %\caption{Deep learning based end-to-end framework}
  }
\end{subfigure}
\begin{subfigure}[Lane change course with velocity 10 kmph]{
  \includegraphics[width=50mm]{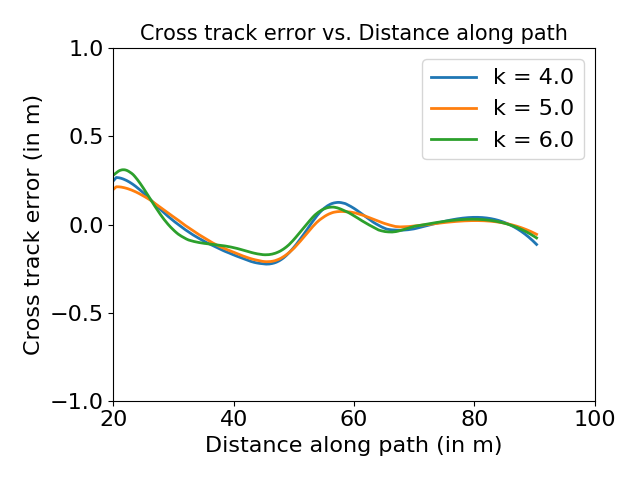}
  %\caption{Conventional feature based monocular visual odometry}
  } 
\end{subfigure}
\begin{subfigure}[Lane change course with velocity 25 kmph]{
  \includegraphics[width=50mm]{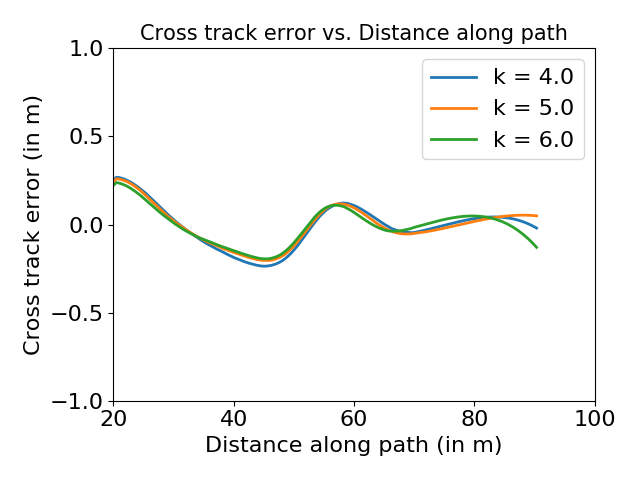}
  %\caption{Deep learning based end-to-end framework}
  }
\end{subfigure}
\begin{subfigure}[Lane Change course with velocity 35 kmph]{
  \includegraphics[width=50mm]{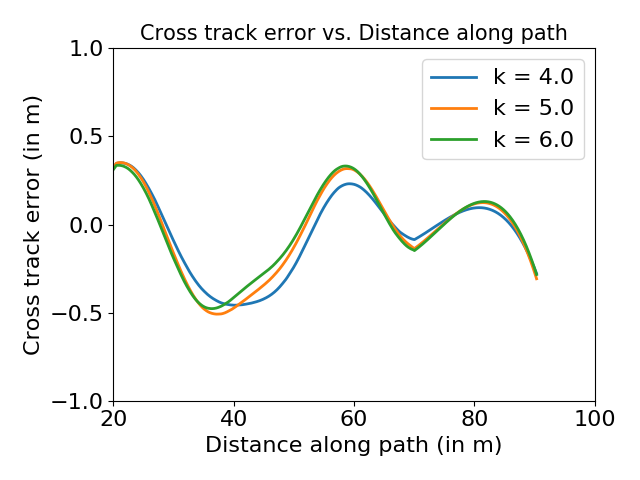}
  %\caption{Deep learning based end-to-end framework}
  }
\end{subfigure}
\begin{subfigure}[Sinusoidal course with velocity 10 kmph]{
  \includegraphics[width=50mm]{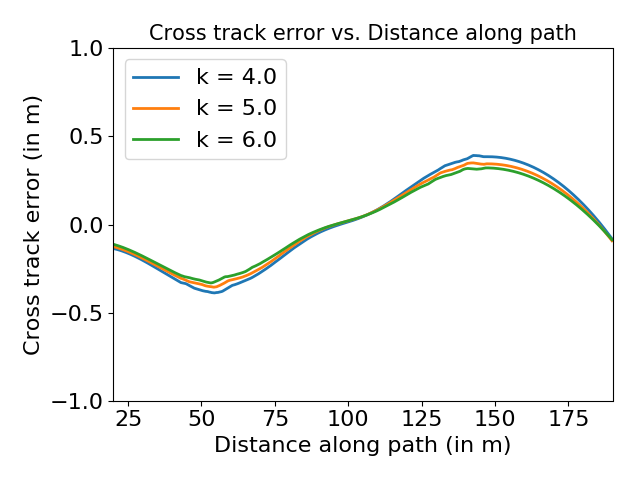}
  %\caption{Conventional feature based monocular visual odometry}
  } 
\end{subfigure}
\begin{subfigure}[Sinusoidal course with velocity 25 kmph]{
  \includegraphics[width=50mm]{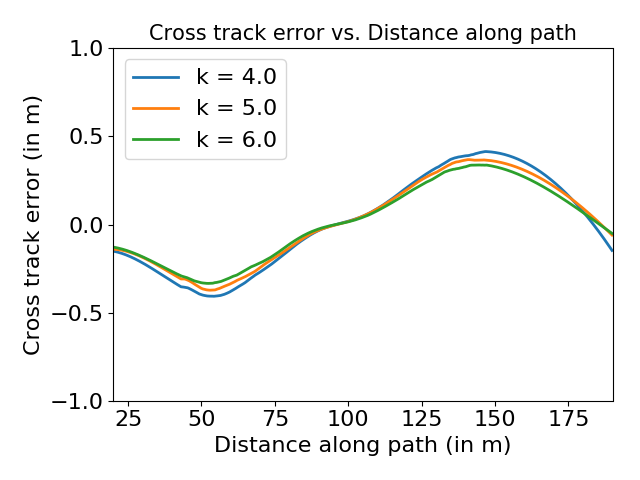}
  %\caption{Deep learning based end-to-end framework}
  }
\end{subfigure}
\begin{subfigure}[Sinusoidal course with velocity 35 kmph]{
  \includegraphics[width=50mm]{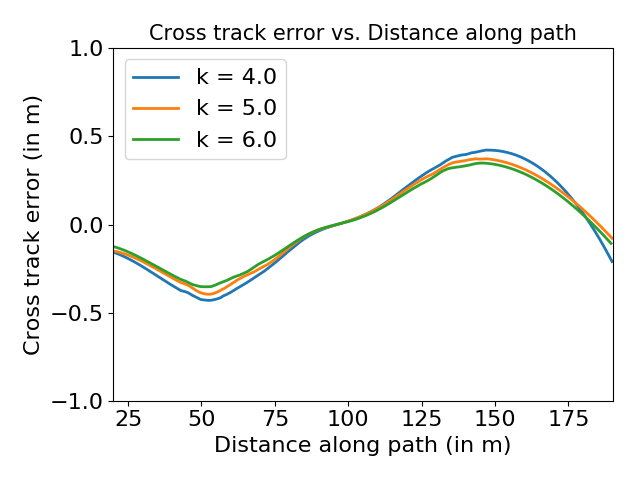}
  %\caption{Deep learning based end-to-end framework}
  }
\end{subfigure}
\caption{LQR controller's plots for cross track error vs. distance travelled on straight course(a,b,c), circular course(d,e,f), lane change course(g,h,i) and sinusoidal course(j,k,l)}
\label{fig:lqr}
\end{figure*}
The PID steering method applied a simple PID feedback controller to the steering error w.r.t the path. It was observed that the gains $K_i$ and $K_d$ didn't have a significant effect on a cross-track error while the $K_p$ gain had a noticeable impact. Hence, we fixed the gains $K_i$ and $K_d$ to a nominal value and only varied $K_p$ (manifested as $k$ in legend) for further tuning observations.

For the straight path, as observed in Figure \ref{fig:pid} (a) (b) (c), the cross-track error kept oscillating within the range $\pm 0.1$ m at a lower speed. These oscillations increased with both an increase in speed and gain value. A low steady-state error was observed similar to the Pure pursuit method with the help of gain $K_i$, but the vehicle underwent small oscillations as it couldn't steer small-angle appropriately. For the circular path, the cross-track error was within a range of 0.3 m for tuned gains. Unusually, for a low value of $K_p$ the cross-track error boomed for all the speed ranges.

For the sinusoidal path, the error was within 0.25 m for tuned constants. However, for a slightly lower $K_p$ the error boomed to ~0.8 m. As seen earlier, higher $K_p$ became unstable at higher speeds. This is again due to the fact that the system overcompensated for a small error which leads to a resonating kind of error curve. For the lane change path, the performance profile was similar for all the 3 tried $K_p$ values with only change in the max error of this profile. However, for high speed, the tracking method failed to converge any of the gain $K_p$ values. This is due to the rapid change in the curvature of the lane change path, which at high speed cannot be tracked using a simple PID steering controller.

The results were highly parameter specific, and proper tuning was required to achieve low average cross-track error and convergent tracking of the path. With a small change in parameters, the cross-track error varied by a noticeable margin. The tuning of parameters was also a trial-and-error process, and unlike other PID systems not intuitive for this purpose. Overall, we can label the PID steering method as a simple error reducing feedback controller, which is only useful to use at low speeds for relatively simple paths (no sharp changes in curvature). The tuning of gains is a tedious and non-intuitive process and often fails at different speeds.

\subsubsection{Linear Quadratic Regulator}
\begin{figure*}[!tbh]
\centering
\begin{subfigure}[Lane change course with velocity 10 kmph]{
  \includegraphics[width=50mm]{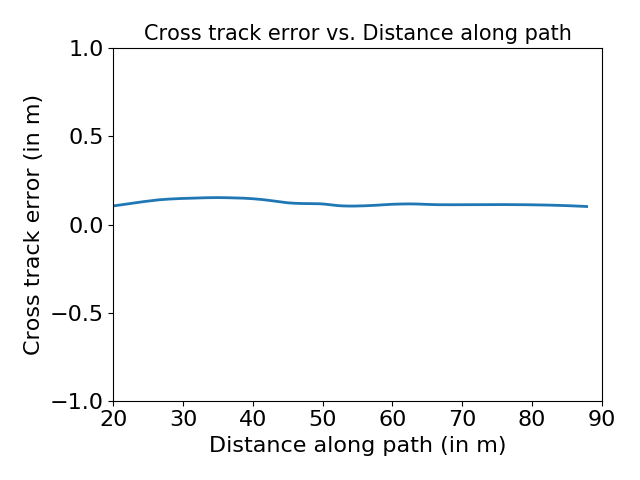}
  %\caption{Conventional feature based monocular visual odometry}
  } 
\end{subfigure}
\begin{subfigure}[Lane change course with velocity 25 kmph]{
  \includegraphics[width=50mm]{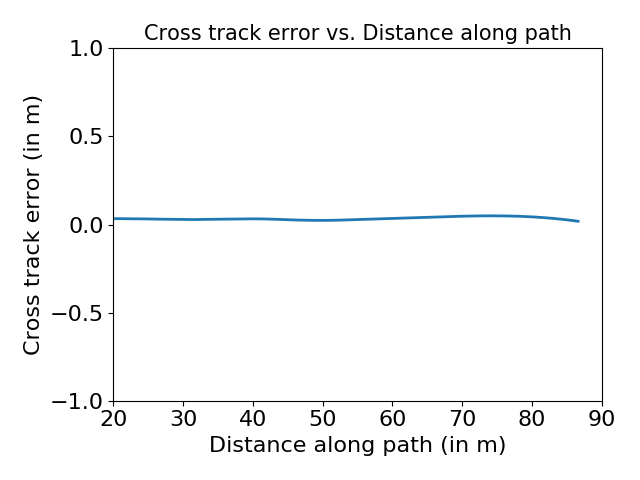}
  %\caption{Deep learning based end-to-end framework}
  }
\end{subfigure}
\begin{subfigure}[Lane Change course with velocity 35 kmph]{
  \includegraphics[width=50mm]{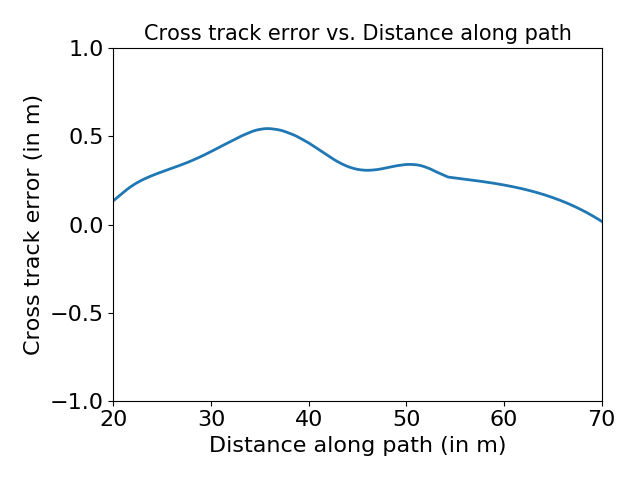}
  %\caption{Deep learning based end-to-end framework}
  }
\end{subfigure}
\begin{subfigure}[Sinusoidal course with velocity 10 kmph]{
  \includegraphics[width=50mm]{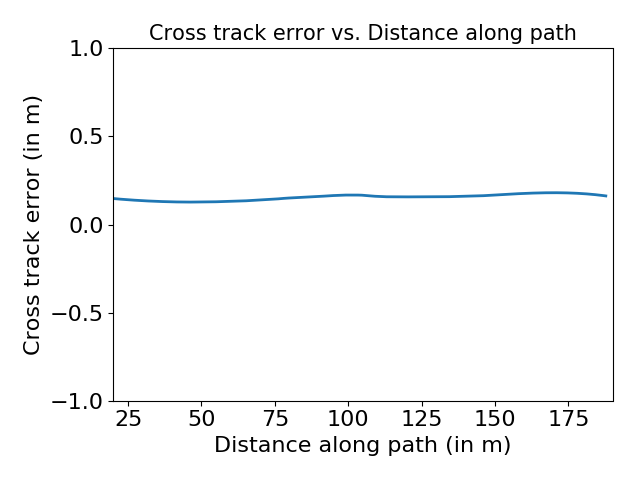}
  %\caption{Conventional feature based monocular visual odometry}
  } 
\end{subfigure}
\begin{subfigure}[Sinusoidal course with velocity 25 kmph]{
  \includegraphics[width=50mm]{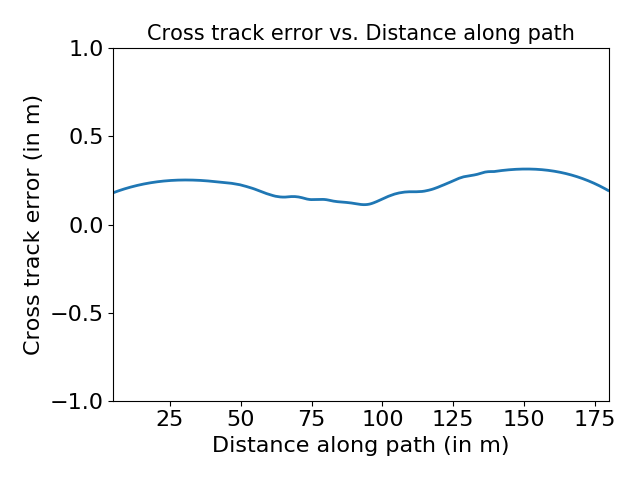}
  %\caption{Deep learning based end-to-end framework}
  }
\end{subfigure}
\begin{subfigure}[Sinusoidal course with velocity 35 kmph]{
  \includegraphics[width=50mm]{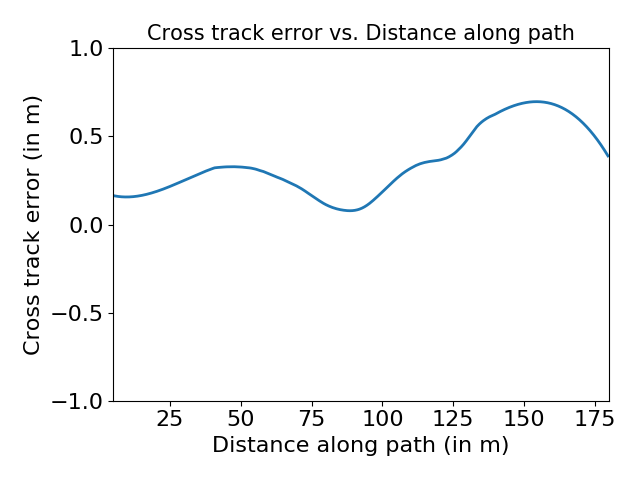}
  %\caption{Deep learning based end-to-end framework}
  }
\end{subfigure}
\caption{Model predictive controller's plots for cross track error vs. distance travelled on lane change course and sinusoidal course}
\label{fig:mpc}
\end{figure*}
The LQR method considers a linearized model for the vehicle and calculates the optimal input for the system at any particular instant. It guarantees asymptotic stability by finding the solution to the Discrete Algebraic Ricatti Equation and using the solution to find the optimal control action. We weight the respective errors to decide the aggressiveness of the control action to minimize the respective error. From our synthesis described in the previous section, we keep parameter $k = q_1$ as the only significant tuning parameter for simulation analysis. Figure \ref{fig:lqr} shows the simulation results obtained on different courses by utilizing this controller.
Other parameters that influenced the controller were:
\begin{itemize}
\item The max iteration for the iterative solution of DARE.
\item The sampling time for the discrete-time model.
\item The tolerance for the iterative solution.
\end{itemize}

While handling curves in the path tracking problem the LQR controller had a decent performance in smooth paths like on the circular course at normal velocities but handling paths with quick changes in curvature and higher velocities got difficult as at boundary points the optimal control action shoots up beyond saturation and possibly due to non-linear dynamics dominating at that time. That is, sudden changes in the path curvature brought about a perturbation that delayed the convergence of the vehicle further. In addition to his, it was observed that the controller well handled constant curvature paths at all velocities.

On paths with a change in concavity i.e., sinusoidal course, the controller didn't track the path with minimal error but at least presented robustness with varying velocity. With the incorporation of a velocity term within the state matrix of this controller, it performed robustly with varying velocity for most of the courses tested. Overall, this controller, in its vanilla form performs well for low varying velocities on different courses and therefore needs some modifications as discussed in the literature for improving its performance on high speed and complex courses.

\subsubsection{Model Predictive Controller}
The Model Predictive Controller uses a predictive method where an internal model predicts the future states of the system and based on the prediction it generates an optimal control action that minimizes a quadratic objective function while satisfying certain constraints. This method helps us to generate optimal control for the system with constraints. For reference tracking, the method is very suitable, considering the robustness and performance of the controller. We use an interior point optimizer library for the optimization process in our case.

Parameters for MPC:
\begin{itemize}
\item Objective function weights for respective error terms.
\item Constraints for the control and state variables.
\item Predicted model for the system.
\item Prediction horizon.
\end{itemize}

The basic strategy of tuning the MPC controller depends on the optimizer used and the way it works. First, we need to set the constraints for the state and control variables, and then we need to change the weights for the objective functions. Similar to the LQR controller, we can set the other weights to be zero initially and give weight to just the cross-track error or randomly initialize the weights for other parameters. Once we know that the weights give an optimal control action, then we can work on fine-tuning. Figure \ref{fig:mpc} shows the performance of this controller using the stated tuning strategy.

When running MPC on paths with abrupt changes in the radius of curvature, we saw the benefit of having a controller that predicts the future changes and optimizes the control actions according to the changes. The controller outputs actions that can ensure continuous and smooth tracking performance. However, on increasing the prediction horizon, the performance is more affected due to more considerable computation time as well as a more significant effect of the future changes to current control action. An overall observation while using MPC controller was its robustness as the performance did not vary a lot with a change in paths as well as velocities. Being computationally heavy is the only demerit that was observed for the controller.

\section{Experiments on Vehicle Platform}
\label{sec:results}
\label{results}
After having implemented the path trackers on the simulation world platform, we tested the same programs on the dedicated passenger car used as a real-world vehicle. Use of ROS came handy as it needed only a few changes to migrate the program from the simulation world and to make them work for an actual passenger car.

\subsection{Longitudinal Control}

\begin{figure}[!tbh]
\centering
\includegraphics[width=70mm]{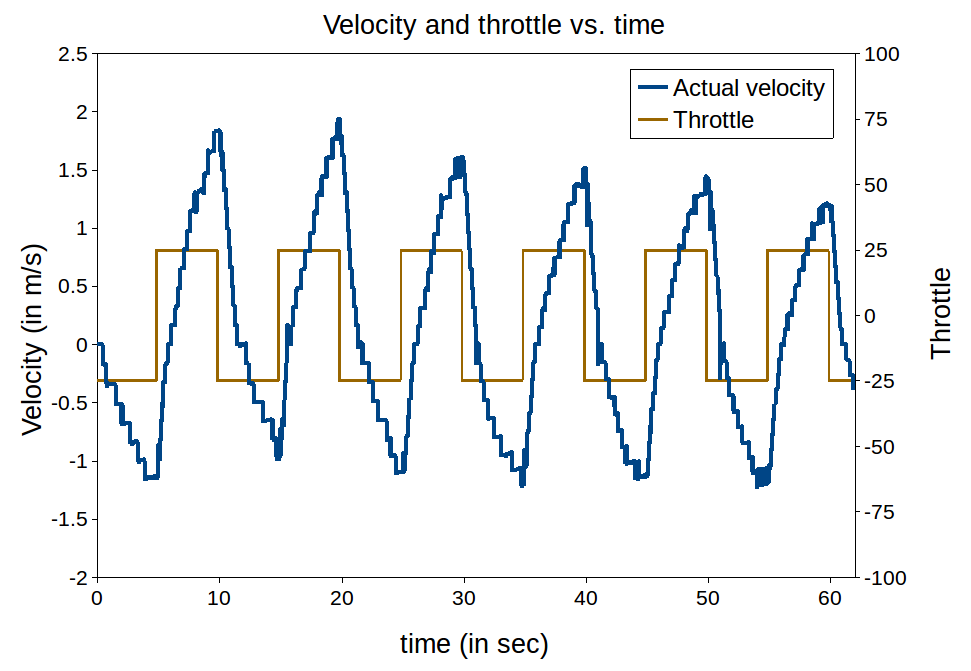} 
\caption{Velocity response of vehicle to square wave throttle input}
\label{fig:vel_response_actual}
\end{figure}

Similar to plant modeling for simulation world, we collected the response data from the instrumented vehicle for tuning the velocity control. Figure \ref{fig:vel_response_actual} shows the response data on square wave throttle input for Mahindra e2o vehicle.

\begin{figure*}[!tbh]
\centering
\begin{subfigure}[Low rise time]{
  \includegraphics[width=72mm]{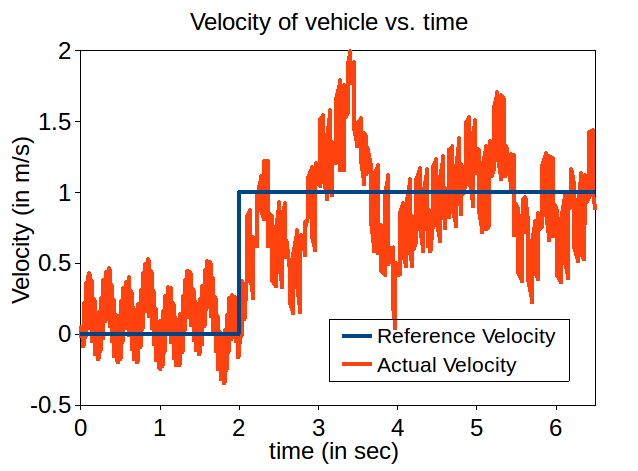}
  \label{fig:step_response_low_rise} }
\end{subfigure}
\begin{subfigure}[Conservative rise time]{
  \includegraphics[width=72mm]{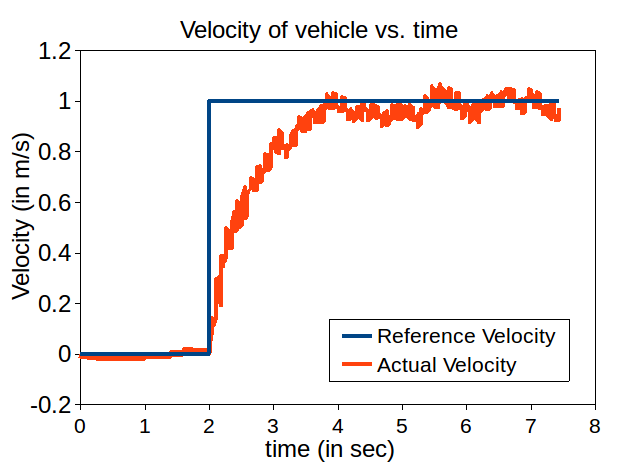}
  \label{fig:step_response_consv_rise}}
\end{subfigure}
\caption{Unit Step Response}
\end{figure*}

The data obtained was very different than observed on the simulation platform. The acceleration and deceleration of this vehicle have a different rate and also non-linear with the throttle. This made tuning the PID controller manually much more difficult and estimating gains using plant modeling was indeed a convenient way out. MATLAB's control toolbox gives the freedom to select an appropriate rise time while estimating the gain value. We first considered a responsive controller by setting the rise time to a low value. Figure \ref{fig:step_response_low_rise} shows the step response for vehicle's velocity control using the estimated gain values with low rise time.

As observed, the response is significantly oscillating, which was not appropriate for path tracking purpose. With this aggressiveness, the rider comfort is also eliminated, leave alone the steady-state error. Hence, we considered using a conservative rise time while tuning the controller for better steady-state error and rider comforts at the price of rise time. Figure \ref{fig:step_response_consv_rise} shows the step response for vehicle's velocity control using the estimated gain values with conservative rise time.

\begin{figure*}[!tbh]
\centering
\begin{subfigure}[Simple PID]{
  \includegraphics[width=72mm]{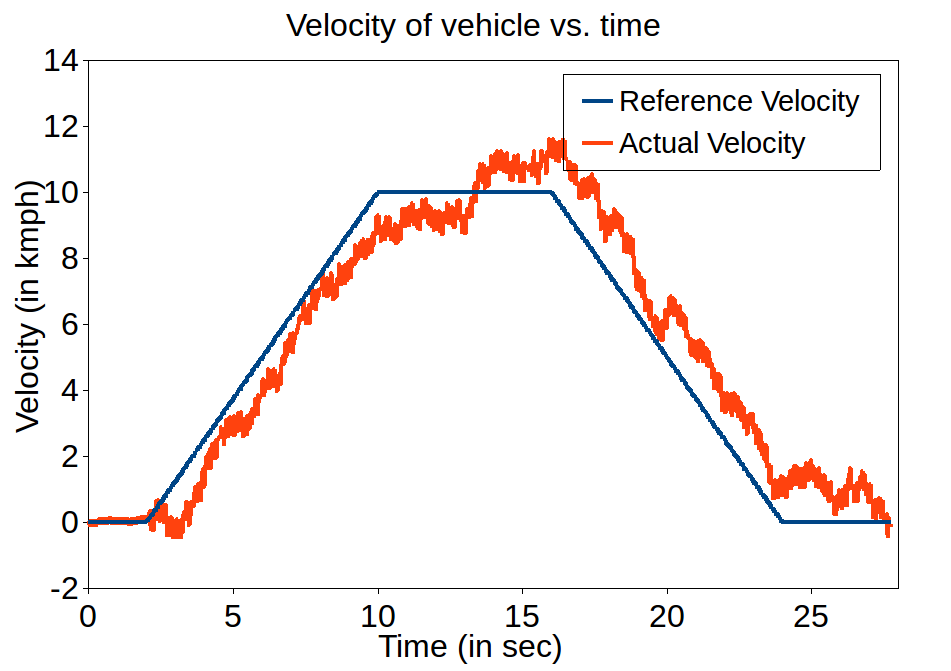}
  \label{fig:trapezium_simple_actual} }
\end{subfigure}
\begin{subfigure}[Adaptive PID]{
  \includegraphics[width=72mm]{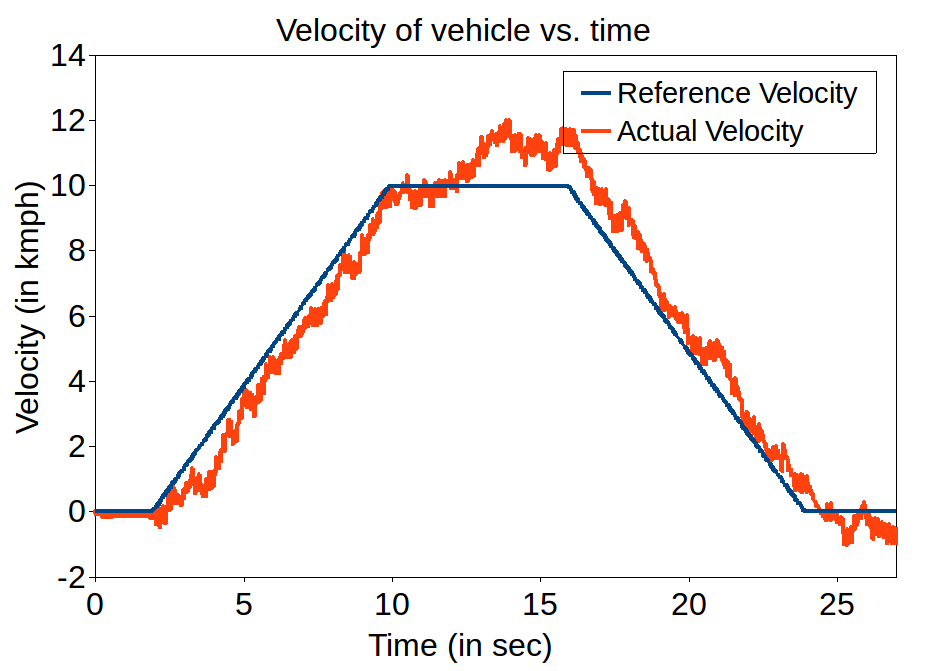}
  \label{fig:trapezium_adaptive_actual}}
\end{subfigure}
\caption{Performance of PID controller on Mahindra e2o}
\end{figure*}

For the path tracking purpose on this vehicle platform, we restricted our experiments to vehicle velocity of 10 kmph as per the availability of field-testing space. We first tested the simple PID controller with the newly estimated gain values for its performance on this velocity. Figure \ref{fig:trapezium_simple_actual} shows the performance of simple PID controller for a trapezium reference profile.

% \begin{figure*}[!tbh]
% \centering
% \begin{minipage}{.5\textwidth}
%   \centering
%   \includegraphics[width=72mm]{results/actual_car/longi_control/trapezium_simple_pid.png}
%   \captionof{figure}{Simple PID}
%   \label{fig:trapezium_simple_actual}
% \end{minipage}%
% \begin{minipage}{.5\textwidth}
%   \centering
%   \includegraphics[width=72mm]{results/actual_car/longi_control/trapezium_adaptive.png}
%   \captionof{figure}}
%   \label{fig:trapezium_adaptive_actual}
% \end{minipage}
% \caption{Performance of PID controller on Mahindra e2o}
% \end{figure*}

As observed, the actual velocity did reach the maximum velocity set point of 10 kmph using the steadily increasing reference input. This way of reaching the set point of 10 Kmph or higher velocity was found to be smoother and comfortable than directly giving 10 Kmph as a step reference. It is also noticed that the actual velocity has two features, one being high-frequency oscillations and the other being low-frequency oscillations. The high-frequency oscillations are attributed due to the presence of noise during actual velocity measurement and estimation. While the low-frequency oscillation is attributed due to the bumpy surface on our field-testing space. This uneven surface perturbated the actual velocity during the experimentations and therefore can be expected to be eliminated on road surfaces in an urban driving scenario. It should be noted that these perturbations lowered the overall performance of the controller for tracking the velocity reference profile.

Further, the vehicle was tested with the presented adaptive PID controller with the same velocity reference profile. Like discussed in simulation analysis of velocity controllers, we used the same gain values along with a nominal gamma. Figure \ref{fig:trapezium_adaptive_actual} shows the performance of adaptive PID controller for the previous velocity reference profile. This controller performed better as it can be observed that the actual velocity profile is closer to the reference profile. Also, the low-frequency oscillations are seen to be dampened with the presence of an adaptive nature. Overall, the adaptive PID controller did perform better; however, setting a nominal gamma value doesn't guarantee its performance under various other conditions that remained untested.

\subsection{Lateral Control}

\begin{figure*}[!tbh]
\centering
\begin{subfigure}[Straight course]{
   \includegraphics[width=35mm]{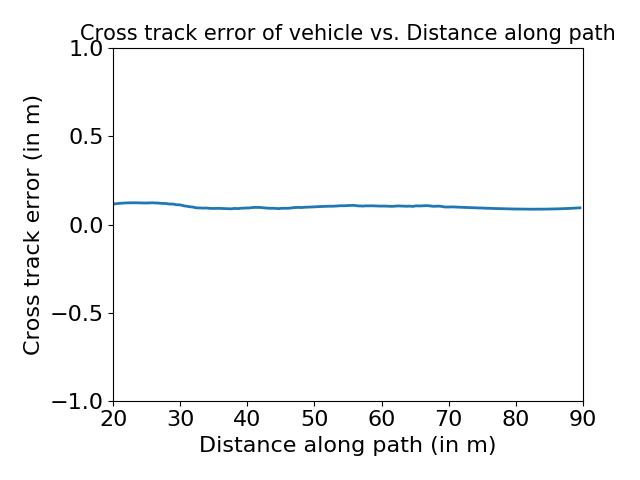}
   %\caption{Conventional feature based monocular visual odometry}
   } 
\end{subfigure}
\begin{subfigure}[Lane change course]{
   \includegraphics[width=35mm]{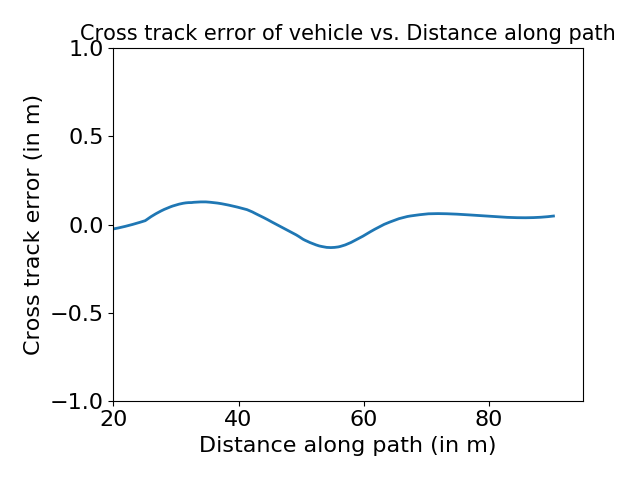}
   %\caption{Deep learning based end-to-end framework}
   }
\end{subfigure}
\begin{subfigure}[Circular course]{
   \includegraphics[width=35mm]{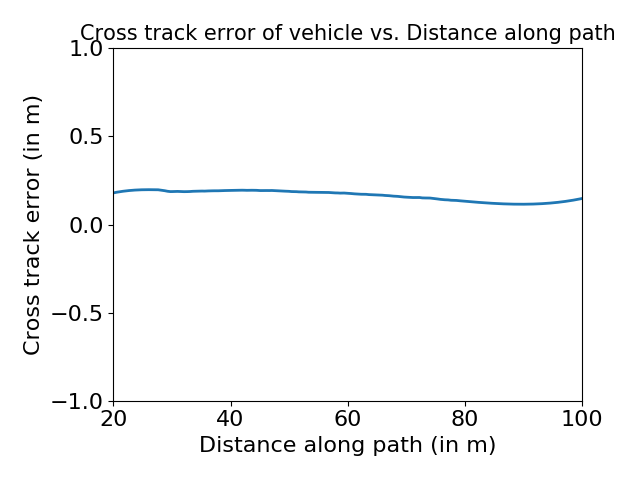}
   %\caption{Deep learning based end-to-end framework}
   }
\end{subfigure}
\begin{subfigure}[Sinusoidal course]{
   \includegraphics[width=35mm]{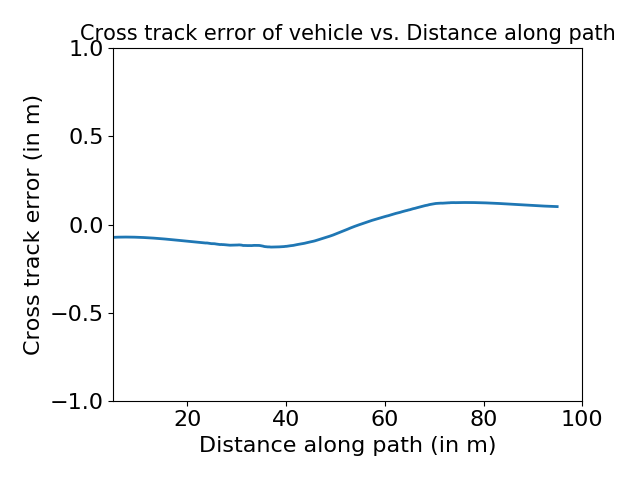}
   %\caption{Deep learning based end-to-end framework}
   }
\end{subfigure}
\caption{Cross track error for pure pursuit tracker on different courses}
\label{fig:pure_pursuit_car}
\end{figure*}

After having the velocity controller tuned, the vehicle was tested with lateral control. A sample performance is shown using the Pure pursuit method with a maximum vehicle speed of 10 Kmph. Figure \ref{fig:pure_pursuit_car} shows the performance of pure pursuit path tracker on different courses in terms of cross-track error.

%figures here

\begin{figure*}[!tbh]
\centering
\begin{subfigure}[Straight course]{
   \includegraphics[width=35mm]{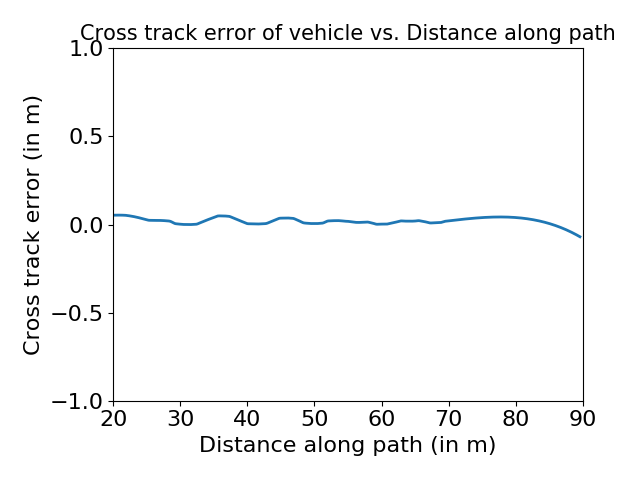}
   %\caption{Conventional feature based monocular visual odometry}
   } 
\end{subfigure}
\begin{subfigure}[Lane change course]{
   \includegraphics[width=35mm]{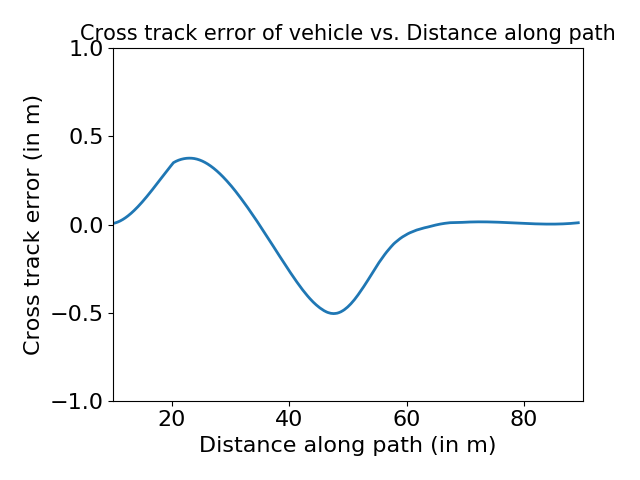}
   %\caption{Deep learning based end-to-end framework}
   }
\end{subfigure}
\begin{subfigure}[Circular course]{
   \includegraphics[width=35mm]{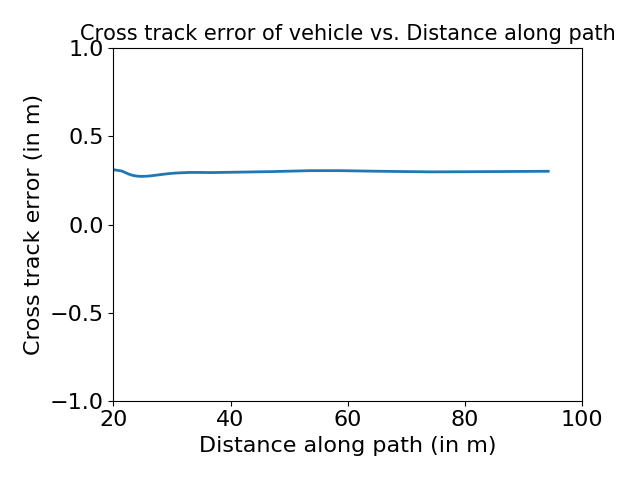}
   %\caption{Deep learning based end-to-end framework}
   }
\end{subfigure}
\begin{subfigure}[Sinusoidal course]{
   \includegraphics[width=35mm]{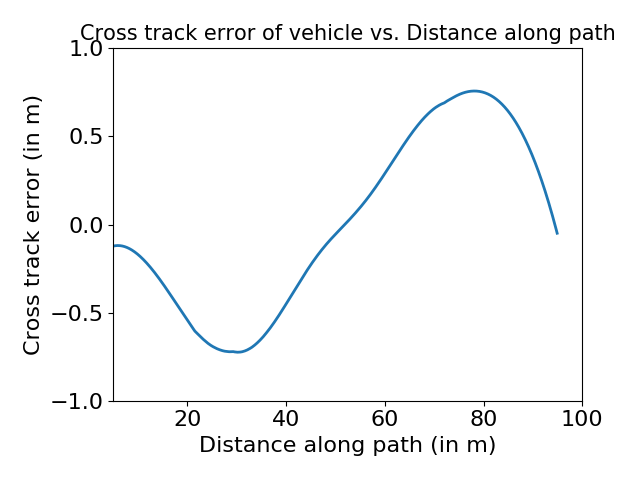}
   %\caption{Deep learning based end-to-end framework}
   }
\end{subfigure}
\caption{Cross track error for PID tracker on different courses}
\label{fig:PID_car}
\end{figure*}

As observed, even at such a low speed, the controller failed to reduce the cross-track error to zero. One of the primary reason is that the steering column fails to deliver the small input steering angle accurately, which is required for path tracker to reduce the cross-track error to zero. The tire-ground interface provides enough friction that the execution of small steering angles by the steering column fails that lets the error accumulate. This kind of observation was possible only after conducting experiments on a real-world platform, which, on the other hand, was absent in the simulation world.  

From experience of tuning the parameters for pure pursuit on simulation, it was possible to relate the performance of the instrumented vehicle's tracking capability with the tuning parameter $k$. Figure \ref{fig:pure_pursuit_car} shows the tuned performance that we were able to achieve. It is possible to obtain better performance if more time can be spent on tuning the controller; however, for the scope of this work, it was only necessary to show the tuning (implementation) study. 

Unlike the Pure pursuit tuning process, the PID controller was not observed to be very intuitive to be tuned. Figure \ref{fig:PID_car} shows the performance of a tuned controller that we were able to achieve. This scenario is quite similar to the difficulties faced while tuning the PID controller for simulation. Though in theory, it is known the importance of gains $K_p$, $K_i$ \& $K_d$ as part of the tuning process; however, they were hardly found to be mimicked on our simulation and real-world data. Moreover, its tuning was found to be much more involving than pure pursuit that was tuned with little effort. Hence, the PID steering law cannot be recommended for path tracking purpose in general.

\section{Conclusions}
\label{sec:conc}
This work presents some methods in the path tracking domain for resolving the complex problem of accurate vehicle navigation. We first developed the experimental setup required for simulation world performance evaluation and later for real-world platform performance evaluation. Development of a simulation world platform for path tracking implementation eased the process of developing an individual tracker algorithm and allowed quick debugging of the complete workflow before it was implemented on the passenger car. Use of ROS as a middleware framework for both simulation world vehicle and real-world vehicle ensured a smooth transition of the codebase to our workstation for real-world experiments. Simulation world platform also allowed fast tuning of control algorithms, observations from which were used to tune the same control algorithm on passenger car efficiently. The platform for the development of real-world scenario was chosen as Mahindra e2o vehicle that is an all-electric CAN controlled vehicle. This vehicle was equipped with various sensors like INS, wheel encoders, and also with a workstation as a controller's computation unit. This instrumented passenger car platform allowed robust experimental validation in different operating conditions, some of which cannot be accurately modeled on a simulation platform. Experiments were performed on this vehicle with courses and velocity profile similar to the ones mentioned in the simulation world subsection. This helped in observing and quantifying the deviation of a real-world physical system from a simulated world one.

Since the path tracking problem comprises longitudinal control and lateral control, we developed them independently. First, the longitudinal velocity control was developed as it is the pre-requisite to accurate performance of lateral control. Using PID as a velocity controller was useful as it performed quite optimally on simulation world data. Though the manually tuned controller wasn't on par with industry grade performance, after having the plant modeled, a significant increase in performance was observed. Plant modeling was most effective using unit step response that was modified to square wave input for consensus calculation. Sinusoidal wave input also did model plant well; however, the steering angle calculation using that model comprised of high-frequency oscillations. This won't be effective on a real-world platform as there exists a degree of latency between calculation and performing the computed throttle. Hence, the plant modeled using sinusoidal wave input was ruled out. We verified the claim that adaptive PID improved performance over the simple PID method as it utilized learning gain values at every time step. Adaptive PID method also showed some resilience to change in vehicle mass that will be a frequently encountered scenario in a real-world application.
 
The lateral control methods implemented in our work so far comprised of a geometric model and linearized dynamic vehicle model. We rederived known methods like Pure pursuit, Stanley, PID, LQR, and MPC for our use case and proved its convergence on MATLAB. Further, we used a computer simulator to demonstrate the working of our implementations of different path tracking methods and also produced some plots for evaluating the performance for each of these methods. All of the controllers were tested on predetermined paths like a straight line, circular course, lane shift course, and sinusoidal course. Rather than having them tested on a random driving course, they were validated against specific geometry of paths to provide insights to their relative advantages and disadvantages in an organized manner. A very detailed model can be cumbersome to obtain and use; thus, the tracking methods described in this study made use of vehicle models that approximate vehicle motion. The simplifications, linearization, and small steering angle assumption of these models varied, which also led to the existence of a minimal cross-track error. From our simulation study, we concluded that up to medium level vehicle speed with an urban driving scenario involving simpler paths, geometric methods of pure pursuit and Stanley could be utilized. PID steering method is not recommended in a real-world application specifically due to high sensitivity to gain values. LQR in its vanilla form gave descent performance on simulation world data, but at times can go haywire if the calculations are not restricted to some limits. This problem is solved in MPC, and therefore it performed well under all scenarios even in its simplest form of implementation. This method is highly recommended for real-world application, the only difficulty being the estimation of vehicle-specific parameters like the LQR method.

Implementation of path trackers on an instrumented vehicle became easy after prototyping the controllers in the simulation world. Standardized development on ROS made it possible to directly utilize the programs developed for the simulation world for the instrumented vehicle with some minor changes. Experience of tuning the controller from a simulation vehicle also made it easy to tune the vehicle platform. For longitudinal velocity control, inaccuracies arose due to latency between the calculation of throttle value and its actual achievement. Moreover, the test arena available for experimentation was an unleveled ground that usually is not encountered in an urban driving scenario. Overall, the velocity controller was able to set the actual velocity to reference velocity as required for the appropriate performance of the lateral control. Implementation of pure pursuit and PID controller both were exactly the same as used for gazebo vehicle. Tuning of pure pursuit was as simple as in the simulation world, and a decent performance was obtained that can be utilized for an urban driving scenario. Tuning of PID controller was tedious and time taking because of its unintuitive nature while steering towards path direction. This controller, therefore, is not recommended towards steering control, however, can be used for velocity control. Problems like reaction time of executing steering angle precisely and inability to execute small steering angle changes lead to deviation of real-world performance from simulation world performance. Also, some of the unmodelled dynamics like steering column inertia and tire-road interface friction lead to performance degradation. These factors tend to affect more to the performance of controllers at higher vehicle speed that was not experimented within our study.\\
The number and types of methods discussed in this study yet are quite limited when compared to the existence of massive literature on path trackers. We wish to evaluate some more methods to bring a comparison based performance evaluation on a large scale. Also, the methods implemented in this study were limited to vehicle velocity of 10 Kmph, 20 Kmph, and 35 Kmph. We wish to conduct experiments on much higher speeds like 45 Kmph + to get a significant amount of differences between these methods. Also, we wish to evaluate the path tracking capabilities on other courses like an eight-figure shaped course, U-turn, and long driving scenario for thorough analysis. Clearly, it was observed that MPC has a lot of scope to be the ultimate path tracker algorithm for autonomous vehicles as per the simulation data. However, the biggest challenge posed while implementing MPC for real-time applications was the enormous computational cost associated with optimization. For non-linear models, linearization of the model poses as another additional challenge. Reducing these computational costs is an active topic in ongoing research. Availability of an accurate model is another major challenge. The proposed solutions to this challenge include using Iterative or Learning MPC, which correctly learns the system model and/or modifies the cost function.\\

*Conflict of interest-none declared

%\newpage 

%\bibliographystyle{dcu}
\bibliographystyle{elsarticle-num}
\addcontentsline{toc}{section}{References}
\bibliography{references}
\end{document}